%% file: 0_main.tex
\documentclass{josis_mdf}

\usepackage[utf8]{inputenc} 
\usepackage[T1]{fontenc}    
\usepackage{hyperref}       
\usepackage{url}            
\usepackage{booktabs}       
\usepackage{amsfonts}       
\usepackage{nicefrac}       
\usepackage{microtype}      
\usepackage{lipsum}
\usepackage{fancyhdr}       
\usepackage{graphicx}       
\graphicspath{{figures/}}     

\usepackage[numbers,square,sort&compress]{natbib}
\usepackage{amsmath}
\usepackage{array}
\usepackage{multirow}
\usepackage[table]{xcolor}
\usepackage{caption}
\usepackage{subcaption}
\usepackage{soul}
\usepackage{makecell}
\usepackage{float}
\usepackage{authblk}
\usepackage{enumitem}

\usepackage[capitalize,noabbrev]{cleveref}

\begin{document}

\title{Development of deep biological ages aware of morbidity and mortality based on unsupervised and semi-supervised deep learning approaches 

}

\author[1]{Seong-Eun Moon\thanks{Equal contribution.}}
\author[2]{Ji Won Yoon\textsuperscript{*}}
\author[1]{Shinyoung Joo}
\author[3]{Yoohyung Kim}
\author[4]{Jae Hyun Bae}
\author[5]{Seokho Yoon}
\author[1]{Haanju Yoo\thanks{Corresponding authors.}}
\author[2,6,7]{Young Min Cho\textsuperscript{\textdagger}}
\affil[1]{NAVER Cloud AI Lab}
\affil[2]{Department of Internal Medicine, Seoul National University Hospital Healthcare System Gangnam Center}
\affil[3]{Department of Internal Medicine, Seoul National University Hospital}
\affil[4]{Department of Internal Medicine, Korea University Anam Hospital \protect\\ Korea University College of Medicine}
\affil[5]{DaNaA Data}
\affil[6]{Department of Internal Medicine, Seoul National University College of Medicine}
\affil[7]{Institute on Aging, Seoul National University}

\affil[1]{\texttt \{seongeun.moon, sy.joon, haanju.yoo\}@navercorp.com}
\affil[2, 3]{\texttt \{jwyoonmd, yhkimmd, ymchomd\}@snu.ac.kr}
\affil[4]{\texttt fermatah@gmail.com}
\affil[5]{\texttt seokho.yoon@danai.co.kr}

\maketitle

\begin{abstract}

\paragraph{Background}
While deep learning technology, which has the capability of obtaining latent representations based on large-scale data, can be a potential solution for the discovery of a novel aging biomarker, existing deep learning methods for biological age estimation usually depend on chronological ages and lack of consideration of mortality and morbidity that are the most significant outcomes of aging. 
\paragraph{Methods}
This paper proposes a novel deep learning model to learn latent representations of biological aging in regard to subjects' morbidity and mortality. The model utilizes health check-up data in addition to morbidity and mortality information to learn the complex relationships between aging and measured clinical attributes. 
\paragraph{Findings}
The proposed model is evaluated on a large dataset of general populations compared with KDM and other learning-based models. Results demonstrate that biological ages obtained by the proposed model have superior discriminability of subjects' morbidity and mortality. 
\end{abstract}

\section{Introduction}
\label{sec:intro}
\input{1_intro}

\section{Related works}
\label{sec:rel_works}
\input{2_related_works}

\section{Proposed method}
\label{sec:model}
\input{3_proposed_method}

\section{Dataset and preprocessing}
\label{sec:data}
\input{4_data}

\section{Results}
\label{sec:exp}
\input{5_experiment}

\section{Conclusion}
\label{sec:conclusion}
\input{7_conclusion}


\bibliographystyle{unsrtnat}  
\bibliography{references}  

\clearpage
\appendix
\input{appendix}

\end{document}

%% file: 1_intro.tex
Along with the massive attention of the public on aging, many studies have tried to quantitatively measure the level of aging.
Chronological age is a traditional indicator of aging, however as demonstrated in previous research \cite{jylhava2017biological, jia2017common}, the biological aging of our body varies depending on genetic and environmental factors. 
This encouraged many researchers to find a novel biomarker of aging in the belief that an appropriate intervention on the biomarker will improve the quality of our life.
The alternative indicator of aging is often called a biological age in contrast with the chronological age. 

Although the biological age based on DNA methylation or telomere length has been popularly investigated in recent decades \cite{horvath2018dna, vaiserman2021telomere, rutledge2022measuring}, these methods require costly examinations such as DNA sequencing, which makes it difficult to obtain large-scale data for developing a generalized biological age and apply the developed biological age to real-world.
Accordingly, many researchers also utilized data obtained from conventional health tests such as blood tests and anthropometric measurements \cite{levine2013modeling, rahman2019centroid}. 
Moreover, approaches for extracting aging-related biomarkers from raw data have been proposed.
For instance, the principal component analysis (PCA) was applied to the results of common examinations including blood and urine tests in \cite{levine2013modeling}.
Deep learning-based approaches, which are expected to extract deep representations of aging, have been introduced recently for biological age estimation \cite{bae2021comparison, pyrkov2018extracting}.

Still, most of the previous research highly depends on chronological age. 
The biological ages are obtained from the statistical analysis of chronological age-related information or the model trained to predict chronological ages. 
This is fundamentally because of the lack of true indicator for aging but also because it was assumed that most (healthy) population has the biological age same as their chronological age. 
However, the assumption does not stand for some cases.
For example, subjects without disease in their old ages should be considered healthier than their chronological ages, while it is common to not have any illness at young ages.
Moreover, the aging pattern of the young generation may differ from that of the old population due to better nutrition and developed healthcare.

Therefore, to establish a generalized model for biological aging, training of the model needs not be strictly oriented by chronological ages. 
Furthermore, diverse populations in terms of their ages and morbidity should be regarded for biological age modeling so that the model can understand age-dependent characteristics of the morbidity.
It is naturally expected that the diversity of age-related morbidity will hinder the performance of chronological age prediction, raising the need for a dedicated evaluation metric for biological ages.

In this paper, we propose a deep learning-based approach for biological age estimation that regards age-related morbidity and mortality for biological modeling without relying on chronological ages.
Our model captures the subjects' health states based on the morbidity and mortality information and draws the gap between biological and chronological ages.
The predicted gap is added to the chronological ages to obtain biological ages.
Here, the proposed biological age is learned in an unsupervised manner, which relieves the potential error propagated from noisy supervised targets (i.e., chronological ages). 
At the same time, our model is trained based on 1) chronological age prediction, 2) negative pairs of morbidity, 3) correlation with mortality, and 4) semantic consistency of the gap to extract meaningful information for biological age. 

Our contribution can be summarized as follows:
\begin{itemize}
    \item A large-scale cohort was brought to biological age modeling including morbidity and mortality information. 
    \item We propose an autoencoder-based deep learning model to estimate the gap between biological and chronological ages from health checkup data. 
    The unsupervised structure enables the predicted biological age to avoid a strong tie with chronological ages that may be noise-prone. 
    \item Our model learns the morbidity and mortality information for the biological age by embedding clinical common senses into training losses. 
    In particular, we made negative pairs for the data with diseases and trained the model to capture the differences that stem from diseases by contrasting elements of the negative pairs. 
    Also, the model is trained to maximize the negative correlation between the number-of-day-to-death and predicted gap using mortality data.
    \item A variety of study populations was defined depending on the age-related morbidity, i.e., \textit{super normal}, \textit{normal}, \textit{average}, and \textit{whole} groups.
    We conducted exhaustive experiments to evaluate the representativeness of the groups for biological age modeling.
\end{itemize}


%% file: 2_related_works.tex
\paragraph{Traditional methods:}
Traditional methods for biological age estimation include multiple linear regression (MLR), principal component analysis-based method (which we will simply call PCA for convenience), and Klemera and Doubal’s method (KDM).
MLR is the initial approach used for the BA estimation \cite{hollingsworth1965correlations,dubina1983biological}, in which the biological age is estimated as a weighted sum of biomarkers related to aging. While MLR is simple to implement and easy to understand, it has several weaknesses. For instance, MLR models tend to regress towards the mean of target values that induces distortions of biological ages at the regression edge and discontinuity in the aging rate \cite{jia2017common}.

\citeauthor{nakamura1988assessment} proposed using principal component analysis for biological age estimation \cite{nakamura1988assessment}, which has been used in numerous studies since then \cite{nakamura1989biological,  park2009developing, nakamura2007method,zhang2014select, nakamura2008sex}. 
In the method, the biomarkers most relevant to chronological age and independent of each other, i.e., the principal components, are chosen to generate biological age scores.
PCA overcomes the problems of MLR, but its estimation results are relatively less interpretable than MLR models because the principal components may be compounded from biomarkers. Moreover, since PCA is a linear model, it cannot learn the non-linear characteristics of the variables. Despite the theoretical improvement of PCA, \citeauthor{levine2013modeling}'s study showed no significant performance difference between MLR and PCA \cite{levine2013modeling}.

\citeauthor{klemera2006anew} came up with a mathematical model, called KDM, which they claimed to be optimal for biological age estimation \cite{klemera2006anew}. KDM regards that the level of aging is represented by biomarkers that systematically change depending on the chronological age.
Under the presumption, the biological age is obtained as a weighted sum of distances from the age-related average for each biomarker.
Several studies experimentally demonstrated that KDM is the best method to estimate the mortality-related biological age \cite{levine2013modeling,jia2017common, cho2010empirical}, however, recent models have started to outperform KDM \cite{rahman2019centroid,kim2022risk}. KDM relies on an assumption that may not be satisfied in the real world; it assumes a close to linear relationship between biomarkers and biological age. Moreover, its dependency on the chronological ages leaves the uncertainty on true targets of the biological age. 

\paragraph{Recent methods:}
Recently, deep learning technology is used for the biological age estimation task by training a deep neural network that predicts subjects' chronological ages. Notably, in \cite{bae2021comparison} and \cite{putin2016deep}, machine learning methods such as XGBoost and random forest were compared to deep neural networks with multiple fully connected layers. \citeauthor{pyrkov2018extracting} employed 1-dimensional CNN and performed a similar in comparison with conventional machine learning methods \cite{pyrkov2018extracting}. 
Unsurprisingly, deep learning models outperformed the other methods in terms of the accuracy of the chronological age prediction. However, these works focused on building a model that predicts the chronological age of healthy individuals and assumed that the predicted age represents the exact health state. However, it is well known that deep learning models, in general, cannot handle out-of-distribution samples properly \cite{pooch2020can,zhang2021deep,berend2020cats}. Thus, the biological age estimation of such a model may be unstable for the subject with a disease. We show that using the data including various morbidity conditions enables a more robust measurement of biological age.

\citeauthor{rahman2019centroid} proposed an age cluster-based algorithm for biological age modeling \cite{rahman2019centroid}. 
Initially, the subjects for training are clustered depending on their chronological ages, and then the mean and standard deviation are calculated for 16 features from the National Health and Human Nutrition Examination Surveys (NHANES) for each cluster.
When a subject for the test is given, the neighbor clusters are chosen by comparing the distance between the centroid of clusters and the feature of the test subject.
The biological age is obtained as a weighted average of the chronological ages of those neighbor clusters. This method gave comparable results to a deep neural network in terms of mortality prediction. However, this clustering-based approach still depends on chronological ages and can only be used when sufficient data for each age group are available. Moreover, the predicted biological age may be unstable if there is a missing observation of features.

Furthermore, there has been an approach to embedding morbidity information into biological ages in \cite{kim2022risk}.
The latent representations of EMR data were extracted from the autoencoder, which is trained based on the KL divergence between the reconstructed data and the risk scores that evaluate health risks. Then, biological ages are calculated via a formula resembling the PCA method. The learning direction based on the risk scores helps the model to learn the morbidity condition of the given data. This unsupervised model showed outperforming performance for the prediction of disease incidence compared with previous methods. Moreover, the group with a larger gap between biological and chronological ages yields higher risk scores.

%% file: 3_proposed_method.tex
\subsection{Goal}

Our goal is to provide a more reliable estimate of biological age by encoding the health status of subjects using the morbidity and mortality information. Our model estimates the gap between biological and chronological ages using latent features corresponding to the $[GAP]$ special token. In addition, chronological age is predicted based on the features of $[CA]$ token to obtain supplementary information of subjects' health states.
We utilize the morbidity of subjects for biological age estimation based on the fact that it is common for subjects with diseases to have a higher biological age than healthy subjects. We substitute values within the normal range for biomarkers related to the diseases to generate contrastive pairs of the data from subjects with the diseases. 
The biological age calculated from the corrected normal data is expected to be lower than that obtained from the original data with diseases. We also consider the mortality by training the proposed model to yield biological ages that are negatively correlated with the number of days to death. By incorporating the biological age-related rules into the loss function of the proposed model, we demonstrate that the proposed model can estimate the biological age of the general population including the subjects with diseases in contrast to previous works that targeted only healthy subjects.

\subsection{Model architecture}

\begin{figure}[htp]
\centering
\includegraphics[width=\textwidth]{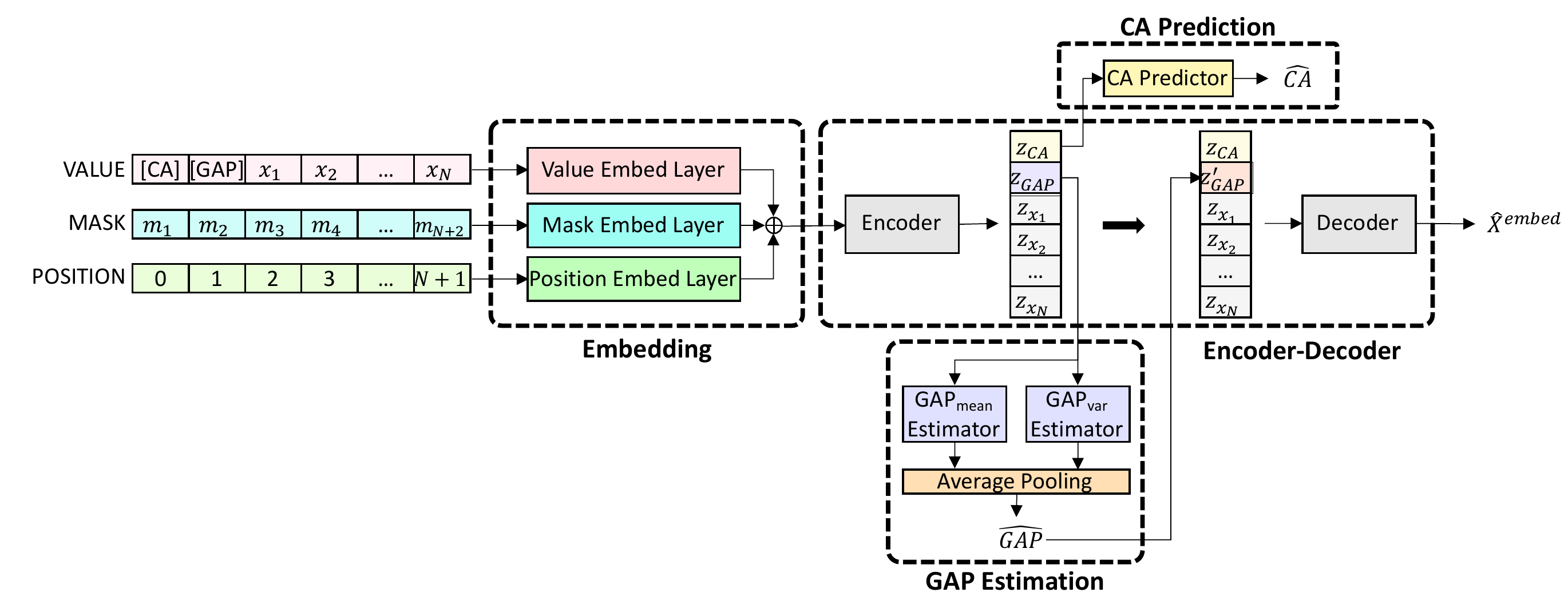}
\caption{\label{fig:model-arch} An overview of the proposed model. } 
\end{figure}

The proposed model architecture is illustrated in \cref{fig:model-arch}. The proposed model is largely divided into four parts: Embedding, Encoder-decoder, CA prediction, and GAP estimation, and all the parts are trained jointly. The Encoder-decoder module learns to extract latent representations that describe the health state of subjects based on the reconstruction task. The CA prediction module is trained to minimize the chronological age prediction loss, which also guides the encoder to put CA-related information into the latent features for $[CA]$ token. The GAP estimation module learns to estimate the gap between biological and chronological ages by reflecting the rules on morbidity and mortality. Details of each module are as follows.

\paragraph{Embedding}
Input data $\mathbf{x}=[x_1, x_2, ..., x_N]$ goes through the embedding process first. 
We attach two special tokens, $[CA]$ and $[GAP]$, to the beginning of input data for chronological age prediction and biological age estimation, respectively.
Health check-up data usually contains missing variables, therefore, we assume that the missing values were filled using an appropriate imputation strategy. 
The features, which are originally measured or imputed afterward, are discriminated using the imputation mask $\mathbf{m}=[m_1, ..., m_{N+2}]$.
The value of the mask is zero for the original variables and one otherwise. 
The position information of input data is described in an absolute manner, i.e., positions of the two special tokens and $N$ features are encoded using integers $\mathbf{p} = [0, ..., N+1]$.

While the imputation masks and positions are embedded using typical embedding layers, $e_m$ and $e_p$, the conventional embedding approach is hard to be adopted for input features because most of the feature values are continuous. 
Therefore, we adopt a fully connected layer $e_f$ for the feature value embedding. 
The embedded feature values, imputation masks, and position information are integrated using a summation operation. 
That is,
\begin{equation}
    \mathbf{X}^{embed} = e_f(\mathbf{x}) + e_m(\mathbf{m}) + e_p(\mathbf{p}).
    \label{eq:embed}
\end{equation}

\paragraph{Encoder-decoder}
As the proposed model presumes no property for the encoder and decoder structures, any kind of learning network can be employed for the Encoder-decoder module as long as they have enough computational capability. 
We select the Performer network \cite{choromanski2021rethinking}, an efficient Transformer with linear self-attention, for the encoder and decoder. 
Note that the usage of special tokens to aggregate the relevant information may vary depending on the network structure. 
For instance, it is more natural to attach the special tokens at the end of the feature vector for recurrent neural networks. 

The encoder $E$ receives $\mathbf{X}^{embed}$ and extracts latent features $\mathbf{Z}=[\mathbf{z}_{CA}, \mathbf{z}_{GAP}, \mathbf{z}_{x_1}, ..., \mathbf{z}_{x_N}]$.
The decoder $D$ restores the embedded input data $\hat{\mathbf{X}}^{embed}$ using the latent features $\mathbf{Z}$.
The training loss for reconstruction of $K$ samples is calculated using mean squared error (MSE), i.e.,
\begin{equation}
    \mathcal{L}_{recon} = \frac{1}{K(N+2)} \sum_{i=1}^K \sum_{j=1}^{N+2} \left(
    \hat{\mathbf{X}}^{embed}_{i, j} - \mathbf{X}^{embed}_{i, j}
    \right)^2,
    \label{eq:recon_loss}
\end{equation}
where $\mathbf{X}^{embed}_{i, j}$ indicates the embedded description for $j$-th feature of $i$-th sample. 
 
\paragraph{CA prediction}
Given latent features for the special token $[CA]$, we predict chronological ages using two fully connected layers with nonlinear activation functions in the CA prediction module. 
The performance of chronological age prediction is evaluated in two ways, MSE and R-squared score.
When the CA prediction network produces $\hat{y}$ for true chronological age $y$, the training loss is defined as:
\begin{equation}
    \mathcal{L}_{CA} = \frac{1}{K} \sum_{i=1}^K (\hat{y}_i - y_i)^2 + \lambda_{CA_{R2}} \left( 1 - \frac{\sum_{i=1}^K (\hat{y}_i - y_i)^2 }{ \sum_{i=1}^K (y_i - \bar{y})^2} \right).
    \label{eq:ca_loss}
\end{equation}
$\lambda_{CA_{R2}}$ is designated to balance the MSE and R-squared score values, and $\bar{y}$ refers to the average of $y$. 
This loss function is designed to enhance the chronological age prediction performance and at the same time, encourage the encoder to extract information related to the chronological age and aggregate the extracted information into the latent features for $[CA]$ token.

\paragraph{GAP estimation}
We parameterize $\mathbf{z}_{GAP}$, the latent features for $[GAP]$ token, as a probability distribution.
Then, gap estimates are sampled from its distribution using the reparameterization trick \cite{kingma2014autoencoding}. 
We use the maximum mean discrepancies (MMD) loss \cite{gretton2012kernel} to drive the estimated gap values $\mathbf{g}$ to follow the normal distribution $\mathcal{N} (0, 1)$. 
\begin{equation}
    \mathcal{L}_{dist} = \text{MMD} \left( p(\mathbf{g} | \mathbf{z}_{GAP}), \mathcal{N}(0, 1) \right)
    \label{eq:dist_loss}
\end{equation}

In addition to the MMD loss, we introduce the consistency loss to force $z_{GAP}$ to be independent to $z_{CA}$.
We make a positive pair for $\mathbf{g}$ by randomly shuffling the $z_{CA}$ across a mini-batch.
The latent features including the shuffled $z_{CA}$ is notated as $\mathbf{Z}^{'}$.
Then, we minimize the MSE between gap values obtained from the original $\mathbf{Z}$ and $E(D(\mathbf{Z}^{'})):=\mathbf{Z}^*$.
Therefore, the consistency loss can be written as:
\begin{equation}
    \mathcal{L}_{consist} = \frac{1}{K} \sum_{i=1}^K \left( f_{GAP_{mean}} (\mathbf{z}_{GAP}) - f_{GAP_{mean}} (\mathbf{z}^*_{GAP}) \right)^2,
    \label{eq:consist_loss}
\end{equation}
where $f_{GAP_{mean}}$ indicates the network for mean value inference of the gap. 

To guide our model to reflect the morbidity of subjects, we use a contrastive loss that ensures the marginal difference between gap estimates obtained from the data of subjects with diseases and data with substituted biomarkers by values within the normal range. 
For example, when a subject is considered to have DM based on her/his blood sugar level, we make a pair of data that consists of the original health check-up results and the data whose blood sugar level was corrected to the normal range. 
The contrastive loss for morbidity can be defined as follows:
\begin{equation}
    \mathcal{L}_{contrast} = \max \left(0, \frac{1}{K} \sum_{i=1}^K \left( \gamma - \mathbf{g}_{org} - \mathbf{g}_{corr} \right) \right),
    \label{eq:contrast_loss}
\end{equation}
where $\gamma$ modulates the margin of gap estimates. 

Moreover, we include the mortality loss that forces the estimated gap values to have a negative correlation with the number of days to death.
The correlation is evaluated using the Pearson correlation coefficient (PCC; see Section \ref{sec:data-feature-analysis} and Equation \ref{eq:pcc} for details of PCC). 
\begin{equation}
    \mathcal{L}_{mort} = 1 + \text{PCC}_{\mathbf{g}, days\_to\_death}
    \label{eq:mort}
\end{equation}

As a result, the total loss function of the proposed model can be summarized as:

\begin{equation}
    \mathcal{L} = \lambda_1  \mathcal{L}_{recon} + \lambda_2  \mathcal{L}_{CA} + \lambda_3  \mathcal{L}_{dist} + \lambda_4  \mathcal{L}_{consist} + \lambda_5  \mathcal{L}_{contrast} + \lambda_6  \mathcal{L}_{mort},
    \label{eq:total}
\end{equation}
where $\lambda_1, ..., \lambda_6$ are the weight values to regulate the influence of losses of different types.

%% file: 4_data.tex
\newcommand{\sfeature}{\textit{base}}
\newcommand{\mfeature}{\textit{morbidity-related}}
\newcommand{\lfeature}{\textit{entire}}

\subsection{Study population}

\begin{table}
    \centering
    \caption{Criteria to determine morbidity states of DM, HBP, DLP, and MS. Abbreviations for features are as follows: FBS= fasting blood sugar, SBP=systolic blood pressure, DBP=diastolic blood pressure, LDLC=low-density lipoprotein cholesterol, TG=triglyceride, HDLC=high-density lipoprotein cholesterol, WC=waist circumference.}
    \begin{tabular}{c m{.37\textwidth} m{.5\textwidth}}
        \toprule \midrule
         & \multicolumn{1}{c}{Normal} & \multicolumn{1}{c}{Illness} \\ \midrule
        
        DM & 
        \shortstack[l]{$\bullet$\quad FBS $<$ 100 mg/dL AND \\ $\bullet$\quad HbA1C < 5.7 \%} &
        \shortstack[l]{$\bullet$\quad FBS $\geq$ 126 mg/dL OR \\ $\bullet$\quad HbA1c $\geq$ 6.5 \% OR \\ $\bullet$\quad taking DM medicine} \\ \cmidrule(lr){1-3}
        
        HBP & 
        \shortstack[l]{$\bullet$\quad SBP < 120 mmHg AND \\ $\bullet$\quad DBP < 80 mmHg} & 
        \shortstack[l]{$\bullet$\quad SBP $\leq$ 140 mmHg OR \\ $\bullet$\quad DBP $\geq$ 90 OR \\ $\bullet$\quad taking HBP medicine} \\ \cmidrule(lr){1-3}
        
        DLP &
        \shortstack[l]{$\bullet$\quad LDLC < 100 mg/dL AND \\ $\bullet$\quad TG < 150 mg/dL AND \\ $\bullet$\quad HDLC $\geq$ 60 mg/dL} & 
        \shortstack[l]{$\bullet$\quad LDLC $\geq$ 160 mg/dL OR \\ $\bullet$\quad TG $\geq$ 200 mg/dL OR \\ $\bullet$\quad HDLC < 40 mg/dL OR \\ $\bullet$\quad taking DLP medicine} \\ \cmidrule(lr){1-3}
        
        MS & \multicolumn{1}{c}{-} &
        \shortstack[l]{In the case three or more of the following \\ conditions are satisfied \\
        $\bullet$\quad WC $\geq$ male: 90 cm, female: 85 cm \\ $\bullet$\quad TG $\geq$ 150 mg/dL \\ $\bullet$\quad HDLC < male: 40 mg/dL, female: 50 mg/dL \\ $\bullet$\quad SBP $\geq$ 135 mmHg AND DBP $\geq$ 85 mmHg \\ $\bullet$\quad FBS $\geq$ 100 mg/dL} \\
        \midrule \bottomrule
    \end{tabular}
    \label{tab:morbidity-criteria}
\end{table}

We employed health check-up data of 151,283 subjects acquired from October 2013 to 2020 for analysis.
The data contains a total of 88 attributes including demographic information, anthropometry records, blood test results, urine test results, and so on. 
In addition, the clinical history of the subjects collected through a questionnaire was used for analyzing the morbidity states. 
Specifically, diagnostic history of diabetes mellitus (DM), high blood pressure (HBP), dyslipidemia (DLP), cancer, cardiovascular disease (CVD), and cardiovascular accident (CVA) and medication status for DM, HBP, and DLP were considered.
Furthermore, we utilized mortality information of the subjects specified by death dates and corresponding causes of death to regard the relationship between predicted biological ages and mortality. 

\begin{figure}
    \centering
    \includegraphics[width=0.53\textwidth]{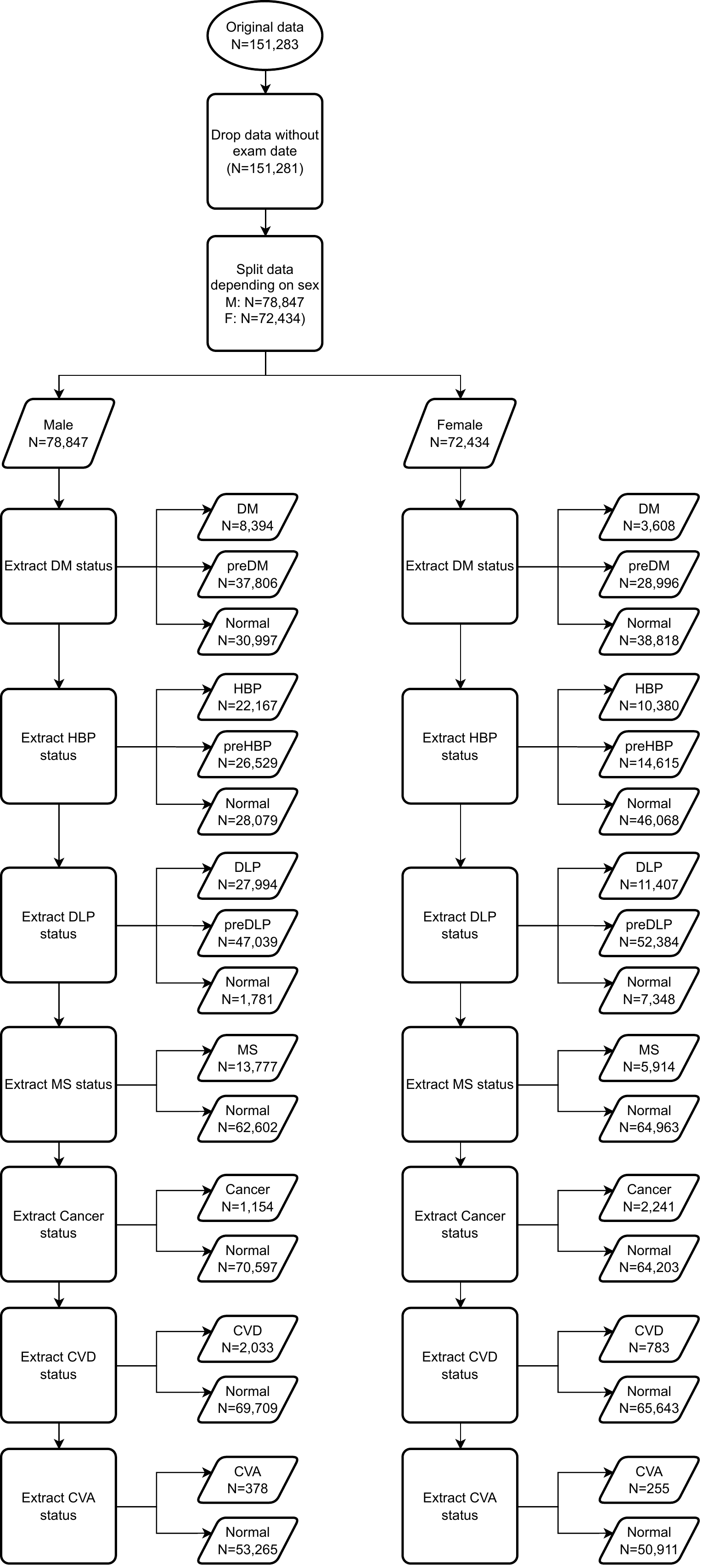}
    \caption{A flowchart for data processing including the disease labeling.}
    \label{fig:data_proc}
\end{figure}

While the subjects' morbidity for cancer, CVD, and CVA was determined based on the diagnostic history, the morbidity states of DM, HBP, DLP, and MS were defined based on health check-up and questionnaire results as described in Table \ref{tab:morbidity-criteria}. 
The subjects who do not satisfy the MS condition were regarded as normal for MS morbidity. 
Moreover, the subjects excluded by both normal and illness conditions of DM, HBP, and DLP were regarded as in a pre-stage of the corresponding disease. 
An overall flow to extract morbidity states and the number of subjects assigned to each state is shown in Figure \ref{fig:data_proc}.


\begin{table}
    \centering
    \caption{Statistics of the data in terms of its total count, average, and standard deviation.}
    \begin{tabular}{c c c c c c c}
        \toprule \midrule
        \multirow{2}{*}{Attribute} & \multicolumn{3}{c}{Male} & \multicolumn{3}{c}{Female} \\ \cmidrule(lr){2-4} \cmidrule(lr){5-7}
        & Count & Mean & Std. & Count & Mean & Std. \\ \midrule
        Age & 78,847 & 46.41 & 12.07 & 72,434 & 45.67 & 12.47 \\
        Time-to-death [day] & 2,240 & 2,794 & 1,676 & 1,106 & 2,665 & 1,689 \\
        BMI [kg/m$^2$] & 76,431 & 24.62 & 11.27 & 70,762 & 21.92 & 3.31 \\
        HbA1c [\%] & 73,470 & 5.73 & 0.74 & 68,649 & 5.61 & 0.55 \\
        \midrule \bottomrule
    \end{tabular}
    \label{tab:data}
\end{table}

\subsection{Representative population groups for biological age estimation}
\label{sec:data-population}

We examined four types of study populations depending on subjects' morbidity to evaluate the representativeness of populations. 
We first extracted the subjects without diseases, which corresponds to the \textit{super normal} group that has been frequently employed in previous studies \cite{bae2021comparison, an2022individualized}.
The \textit{normal} group additionally includes subjects who are in the pre-disease stage of DM, HBP, and DLP. 
Eight features related to the diseases, i.e., Hemoglobin A1c (DM), fasting blood sugar (DM, MS), systolic blood pressure (HBP, MS), diastolic blood pressure (HBP, MS), low-density lipoprotein cholesterol (DLP), triglyceride (DLP, MS), high-density lipoprotein cholesterol (DLP, MS), and waist circumference (MS), were considered to define the \textit{average} group.
The subjects whose disease-related features $X^D = \left\{ x^D_1, ..., x^D_8 \right\} $ are in the range of $[\mu_{x^D_i} - 2\sigma_{x^D_i}, \mu_{x^D_i} + 2\sigma_{x^D_i}]$, where $\mu_{x^D_i}$ and $\sigma_{x^D_i}$ indicate the mean and standard deviation of feature $x^D_i$, are included in the \textit{average} group.
Finally, the \textit{whole} group refers to the entire subject in the dataset. 

For each type of population, the subjects were randomly split for the training, validation, and test according to a ratio of 0.70, 0.15, and 0.15, respectively. 
The data division was repeated three times with different random seeds.
We report the average results of the three repeated experiments as the final results through Section \ref{sec:exp}.


\subsection{Selection of feature sets}
\label{sec:data-feature}

\subsubsection{Correlation analysis for chronological ages}
\label{sec:data-feature-analysis}

\begin{table}
    \centering
    \caption{Results of the correlation analysis between observed features and chronological ages for the entire subjects (All), female subjects (F), and male subjects (M).}
    \begin{tabular}{c m{.43\textwidth} m{.43\textwidth}}
        \toprule \midrule
            & \textit{Whole} & \textit{Super normal} \\ \midrule
        All & 
            BUN, eGFR (CKD-EPI), eGFR (MDRD), FEV1, FEV1-FVC ratio, FVC, Hemoglobin A1c, Visceral fat area  &
            Albumin, eGFR (CKD-EPI), eGFR (MDRD), FEV1, FEV1-FVC ratio, FVC percentile \\ \midrule
        F & 
            Abdominal fatness, eGFR (CKD-EPI), eGFR (MDRD), FEV1, FEV1-FVC ratio, FVC, Height, Systolic blood pressure, Visceral fat area, Waist circumference &
            Albumin, Calcium, eGFR (CKD-EPI), eGFR (MDRD), FEV1, FEV1-FVC ratio, FVC, Height, Helicobacter pylori antibodies, Mean corpuscular volume, Red blood cell count \\ \midrule
        M & 
            Albumin, eGFR (CKD-EPI), eGFR (MDRD), FEV1, FEV1-FVC ratio, FVC, Height, Hemoglobin A1c, Intracellular fluid, Mineral, Protein, Red blood cell count &
            eGFR (CKD-EPI), eGFR (MDRD), FEV1, FEV1 percentile, FEV1-FVC ratio, FVC, Visceral fat area \\ \midrule
        \bottomrule
    \end{tabular}
    \label{tab:feature_correlation_analysis}
\end{table}

To determine the biomarkers that are closely related to aging, we analyzed the relationship between features and chronological ages using the Pearson correlation coefficient (PCC), Spearman rank-order correlation coefficient (SROCC), and mutual information (MI) for the \textit{whole} and \textit{super normal} groups.
The PCC of two variables $X_1=\{x_{1,1}, ..., x_{1,n}\}$ and $X_2=\{x_{2,1}, ..., x_{2,n}\}$ is defined as:
\begin{equation}
    \text{PCC}_{X_1, X_2} = \frac { \text{cov}(X_1, X_2) } {\sigma_{X_1} \sigma_{X_2}},
    \label{eq:pcc}
\end{equation}
where $\text{cov}$ indicates the covariance. 
The value of PCC ranges between -1 and 1, which indicates the perfectly negative and positive linear relationships, respectively.
If there is no relationship between the two variables, the PCC results in zero.

The SROCC measures the PCC of ranked variables to evaluate the monotonicity rather than the linear relationship between variables. 
That is, the SROCC is calculated as:
\begin{equation}
    \text{SROCC}_{X_1, X_2} = \frac { \text{cov}(R_{X_1}, R_{X_2}) } {\sigma_{R_{X_1}} \sigma_{R_{X_2}}},
    \label{eq:srocc}
\end{equation}
where $R_{X_1}$ and $R_{X_2}$ refer to the ranked variables of $X_1$ and $X_2$, respectively. 
If there is no rank-tie so that the variables have $n$ distinct ranks, the SROCC also can be obtained as $1 - \frac{ 6 \sum_{i=1}^n R_{x_{1, i}} - R_{x_{2, i}}} {n(n^2 - 1)}$. 
As the calculation of SROCC is the same as that of PCC, it is also between -1 and 1 while the outcomes are interpreted in terms of the monotonic relationship instead of the linear relationship. 

The MI evaluates the informational correlation between two variables. 
When $X_1$ and $X_2$ are independent, the MI between them becomes zero. 
A larger value of MI for $X_1$ and $X_2$ indicates that $X_1$ contains a larger amount of information about $X_2$ and vice versa. 
Regarding the probability distributions $P_{X_1}$, $P_{X_2}$, and their joint distribution $P_{X_1 X_2}$, the MI is calculated by
\begin{equation}
    \text{MI}_{X_1, X_2} = \sum_{i=1}^n P_{X_1 X_2}(x_{1,i}, x_{2,i}) \log \frac {P_{X_1 X_2}(x_{1,i}, x_{2,i})} {P_{X_1}(x_{1,i}) P_{X_2}(x_{2,i})}.
    \label{eq:mi}
\end{equation}
The Equation \ref{eq:mi} can be rewritten based on Shannon's entropy $H$ as $H(X_1) - H(X_1|X_2)$ or $H(X_2) - H(X_2|X_1)$, from which we can interpret the value of MI as the reduced amount of uncertainty about $X_1$ ($X_2$) by observing $X_2$ ($X_1$). 

The features whose PCC, SROCC, or MI values with chronological ages are in the top 10\% are shown in Table \ref{tab:feature_correlation_analysis}.
We targeted the features that are generally measured for health check-ups, for which the features with a missing rate larger than 50\% were excluded from the analysis.

It is noticeable that the features related to lung function, i.e., FEV1, FEV1-FVC ratio, and FVC are selected for all cases. 
Comparing the results of \textit{whole} and \textit{super normal} population, the chronic disease-relevant features such as hemoglobin A1c and systolic blood pressure show significance only for the \textit{whole} group. 

As calculations of the FVC percentile and eGFRs include the chronological age, they are not considered as relevant features of the chronological age. 
Moreover, the visceral fat area, abdominal fatness, intracellular fluid, mineral, and protein, which are calculated based on the bioelectrical impedance measurement, are discarded due to the uncertainty of its accuracy. 
The height also shows a significant relationship with chronological age.
However, it probably reflects the enhanced nutrition in the growth period rather than the effect of aging.

\subsubsection{Feature selection}
\label{sec:data-feature-selection}

We selected twelve features to describe various aspects of health states based on the result of correlation analysis and expert knowledge.

\begin{itemize}

\item \textbf{Anemia:} 
Two features, i.e., the red blood cell count (RBC) and mean corpuscular volume (MCV), were chosen to describe the anemia condition, which frequently occurs in old age. 
\item \textbf{Fatness:} 
We used the waist circumference (WC), body mass index (BMI), skeleton muscle mass (SMM), body fat mass (BFM), and fat-free mass (FFM) as indicators of fatness.
Note that we employed FR and MR instead of the calculation-based features. 
\item \textbf{Inflammation:}
We added the high-sensitivity C-reactive protein (hsCRP) to describe the amount of inflammation in the body. 
\item \textbf{Kidney function:}
Instead of the eGFR features, we chose creatinine (Cr) which is the source data for eGFR calculations.
\item \textbf{Lung function:}
The FEV1 and FVC were picked to consider the lung function based on the result of the correlation analysis.
\item \textbf{Metabolism:}
The hemoglobin A1c (HbA1c) was considered to represent the metabolic state of the body.
The metabolic disorder, i.e., DM, also can be captured by using HbA1c.
\item \textbf{Nutrition:}
The albumin (Alb), which shows a significant relationship with chronological age mainly for the \textit{super normal} population, was employed as a nutritional marker of the body. 

\end{itemize}

These features are regarded as minimal for understanding the health status of subjects. 
Therefore, we call this feature set \sfeature. 

In addition to the \sfeature\ feature set, we built the \mfeature\ feature set to consider the morbidity of subjects. 
All disease-related features listed in Section \ref{sec:data-population} were added to the \sfeature\ feature set. 
That is, fasting blood sugar (FBS), systolic blood pressure (SBP), diastolic blood pressure (DBP), low-density lipoprotein cholesterol (LDLC), triglyceride (TG), and high-density lipoprotein cholesterol (HDLC) were involved. 
Therefore, the \mfeature\ feature set contains a total of eighteen features.

Finally, we consider the entire set of biomarkers except for the features, which are calculated based on the chronological age or bioelectrical impedance measurements. 
A total of 88 features are included in the \lfeature\ feature set.
However, the actual number of features used for modeling differs depending on the missing rate. 
The features that are not completely measured for training subjects were rejected from the modeling while partially missing measurements are imputed by the mean value of the training data.

%% file: 5_experiment.tex
\subsection{Implementation}
\label{sec:exp-imp}

We built separate models for female and male subjects to consider the biological differences depending on biological gender. 
Experiments were conducted using PyTorch. 
The input data were converted to a 128-dimensional vector during the embedding process.
We used a 3-layer Performer network with four head attention and 1024 hidden dimensional feedforward layers for the encoder and decoder.
We employed AdamP optimizer \cite{heo2021adamp} to train the proposed model.
We selected a batch size of 2,000 and a learning rate of $1e-3$ for training, which continued for 1000 epochs while the early stopping was applied if there is no performance improvement for the validation data during successive 100 epochs. 

The \textit{KDM}, CA cluster-based (\textit{CAC}) \cite{rahman2019centroid}, and \textit{DNN} \cite{bae2021comparison} approaches were employed for comparison.
We implemented the baseline models according to their original papers.
The hyperparameters of DNN, e.g., the number of hidden layers and hidden nodes, the learning rate, and the number of training epochs, were selected as the same as that of the best model in \cite{bae2021comparison}. 
The features related to chronological age were chosen from \lfeature\ feature set for KDM in accordance with the original paper.
Put simply, we discarded tumor markers, eyesight measurements, categorical variables, and other calculated features from the \lfeature\ set for the KDM approach. 
We also examined KDM using the standard \lfeature\ set, which showed relatively unstable results compared with the dedicated \lfeature\ set.

\subsection{Discriminability for disease states}
\label{sec:exp-morbidity}

\newcommand{\ood}{\cellcolor{gray!10}}
\newcolumntype{o}{>{\columncolor{gray!10}}r}

\begin{table}
    \centering
    \caption{Average BA gap values for overall disease states of male (M) and female (F) subjects. The healthy (Hlty) population refers to the subjects without any disease or risk of disease while the average (Avg) population includes the subjects in a high-risk group of diseases unless they are diagnosed with any disease (Unhlty). Numbers with gray-colored background indicate that the corresponding test result is for data whose distribution differs from that of training data (out-of-distribution cases), and bold numbers mark the case that satisfies 1) a negative gap for healthy subjects, 2) a small absolute gap for average subjects (i.e., $< 1$), and 3) a positive gap for unhealthy subjects. The cases satisfying the relative condition between subjects groups (healthy < average < unhealthy) are underlined.}
    \small
    \setlength\tabcolsep{5pt}
    \def\arraystretch{0.8}
    \begin{tabular}{c c c r r r r r r r r r}
        \toprule \midrule
        
        \multicolumn{2}{c}{\multirow{2}{*}{Population}} & \multirow{2}{*}{Model} & \multicolumn{3}{c}{\lfeature} & \multicolumn{3}{c}{\mfeature} & \multicolumn{3}{c}{\sfeature} \\ \cmidrule(lr){4-12}
        & & & Hlty & Avg & Unhlty & Hlty & Avg & Unhlty &  Hlty & Avg & Unhlty \\ \midrule
        
        \multirow{16}{*}{M} &
        \multirow{4}{*}{\textit{Whole}} & 
            KDM & 3.65 & -0.94 & 0.33 & -0.80 & -5.55 & 2.35 & 5.59 & -2.47 & 0.80 \\
            & & CAC & -1.62 & -2.06 & -3.45 & \underline{-10.74} & \underline{-5.39} & \underline{0.12} & -0.04 & -2.85 & -1.66 \\
            & & DNN & 0.02 & 0.60 & 0.19 & 1.19 & 0.56 & -0.10 & 3.31 & 1.08 & -0.57 \\ \cmidrule(lr){3-12}
            & & proposed & \textbf{\underline{-3.73}} & \textbf{\underline{-0.81}} & \textbf{\underline{0.47}} & \underline{-4.21} & \underline{-1.54} & \underline{0.29} & \textbf{\underline{-2.23}} & \textbf{\underline{-0.75}} & \textbf{\underline{0.49}} \\ \cmidrule(lr){2-12}
            
        & \multirow{4}{*}{\textit{Average}} & 
            KDM & 2.72 & -0.29 & 0.41 & \underline{-4.73} & \underline{-3.38} & \underline{2.63} & 0.40 & -0.84 & 0.34 \\
            & & CAC & \underline{-3.44} & \underline{-1.94} & \underline{-1.71} & \underline{-12.85} & \underline{-4.60} & \underline{0.85} & \underline{-6.35} & \underline{-3.71} & \underline{-1.83} \\
            & & DNN & -0.24 & 0.05 & -0.63 & 1.62 & 1.05 & -0.15 & 2.68 & 0.79 & -0.85 \\ \cmidrule(lr){3-12}
            & & proposed & \underline{-4.28} & \underline{-1.20} & \underline{1.27} & \underline{-5.32} & \underline{-1.93} & \underline{0.82} & \textbf{\underline{-2.51}} & \textbf{\underline{-0.79}} & \textbf{\underline{0.93}} \\ \cmidrule(lr){2-12}
            
        & \multirow{4}{*}{\textit{Normal}} & 
            KDM & \textbf{\underline{-2.90}} & \textbf{\underline{0.04}} & \ood{\textbf{\underline{1.08}}} & \textbf{\underline{-3.93}} & \textbf{\underline{-0.16}} & \ood{\textbf{\underline{5.91}}} & 2.37 & 0.03 & \ood{0.32} \\
            & & CAC & \textbf{\underline{-3.60}} & \textbf{\underline{-0.07}} & \ood{\textbf{\underline{0.65}}} & \textbf{\underline{-9.21}} & \textbf{\underline{-0.08}} & \ood{\textbf{\underline{4.56}}} & \underline{-1.94} & \underline{-0.89} & \ood{\underline{-0.22}} \\
            & & DNN & -0.24 & -0.21 & \ood{-1.89} & 1.57 & 0.14 & \ood{-2.20} & 3.73 & 1.71 & \ood{-1.31} \\ \cmidrule(lr){3-12}
            & & proposed & \textbf{\underline{-2.13}} & \textbf{\underline{-0.06}} & \ood{\textbf{\underline{1.34}}} & \underline{-0.71} & \underline{1.70} & \ood{\underline{4.10}} & \underline{-0.67} & \underline{-0.16} & \ood{\underline{-0.03}} \\ \cmidrule(lr){2-12} 
            
        & \multirow{4}{*}{\textit{Super normal}} & 
            KDM & \textbf{\underline{-0.81}} & \ood{\textbf{\underline{0.22}}} & \ood{\textbf{\underline{0.36}}} & \underline{-0.10} & \ood{\underline{4.46}} & \ood{\underline{9.79}} & 0.85 & \ood{-2.65} & \ood{-1.58} \\
            & & CAC & -0.96 & \ood{-3.73} & \ood{-8.28} & 1.64 & \ood{-1.35} & \ood{-5.69} & 0.80 & \ood{-1.75} & \ood{-5.97} \\
            & & DNN & -0.73 & \ood{-0.17} & \ood{-6.31} & 0.26 & \ood{-4.95} & \ood{-16.86} & -0.44 & \ood{-2.25} & \ood{-6.02} \\ \cmidrule(lr){3-12}
            & & proposed & 0.54 & \ood{-0.05} & \ood{0.07} & \underline{1.87} & \ood{\underline{3.45}} & \ood{\underline{5.89}} & \underline{0.59} & \ood{\underline{2.21}} & \ood{\underline{4.34}} \\ \midrule 
            
        \multirow{16}{*}{F} &
        \multirow{4}{*}{\textit{Whole}} & 
            KDM & \textbf{\underline{-1.31}} & \textbf{\underline{-0.59}} & \textbf{\underline{0.54}} & \underline{-22.68} & \underline{-10.00} & \underline{8.51} & \underline{-22.25} & \underline{-10.44} & \underline{9.50} \\
            & & CAC & -4.15 & -4.79 & -4.30 & \underline{-7.78} & \underline{-5.04} & \underline{0.19} & -3.14 & -3.58 & -1.28 \\
            & & DNN & 0.31 & 0.40 & -0.37 & 0.52 & 0.57 & -0.02 & 1.76 & 0.49 & -1.06 \\ \cmidrule(lr){3-12}
            & & proposed & \textbf{\underline{-3.29}} & \textbf{\underline{-0.99}} & \textbf{\underline{1.19}} & \underline{-3.11} & \underline{-1.11} & \underline{0.69} & \textbf{\underline{-0.59}} & \textbf{\underline{-0.31}} & \textbf{\underline{0.66}} \\ \cmidrule(lr){2-12}
            
        & \multirow{4}{*}{\textit{Average}} & 
            KDM & \textbf{\underline{-2.05}} & \textbf{\underline{-0.57}} & \textbf{\underline{1.67}} & \underline{-9.43} & \underline{-2.51} & \underline{7.76} & \underline{-7.62} & \underline{-2.20} & \underline{6.68} \\
            & & CAC & \underline{-6.03} & \underline{-3.74} & \underline{1.23} & \underline{-7.98} & \underline{-3.22} & \underline{2.42} & \underline{-4.67} & \underline{-2.80} & \underline{0.37} \\
            & & DNN & 0.85 & 1.09 & 0.08 & 0.08 & 0.01 & -1.39 & 3.22 & 1.91 & -1.74 \\ \cmidrule(lr){3-12}
            & & proposed & \textbf{\underline{-3.47}} & \textbf{\underline{-0.79}} & \textbf{\underline{2.17}} & \textbf{\underline{-4.04}} & \textbf{\underline{-0.62}} & \textbf{\underline{2.81}} & \textbf{\underline{-1.70}} & \textbf{\underline{-0.54}} & \textbf{\underline{2.64}} \\ \cmidrule(lr){2-12}
            
        & \multirow{4}{*}{\textit{Normal}} & 
            KDM & \textbf{\underline{-8.60}} & \textbf{\underline{0.17}} & \ood{\textbf{\underline{10.62}}} & \textbf{\underline{-12.29}} & \textbf{\underline{-0.21}} & \ood{\textbf{\underline{15.75}}} & \textbf{\underline{-9.70}} & \textbf{\underline{-0.01}} & \ood{\textbf{\underline{13.17}}} \\
            & & CAC & \underline{-6.01} & \underline{-1.75} & \ood{\underline{4.05}} & \underline{-9.80} & \underline{-1.44} & \ood{\underline{5.57}} & \underline{-4.47} & \underline{-2.24} & \ood{\underline{0.08}} \\
            & & DNN & 1.44 & 1.20 & \ood{-1.78} & -0.11 & 0.60 & \ood{-2.62} & 1.31 & 0.44 & \ood{-4.47} \\ \cmidrule(lr){3-12}
            & & proposed & \textbf{\underline{-1.56}} & \textbf{\underline{0.26}} & \ood{\textbf{\underline{3.27}}} & 0.31 & 0.22 & \ood{0.28} & \textbf{\underline{-0.66}} & \textbf{\underline{0.19}} & \ood{\textbf{\underline{0.91}}} \\ \cmidrule(lr){2-12}
            
        & \multirow{4}{*}{\textit{Super normal}} & 
            KDM & \underline{-0.16} & \ood{\underline{6.40}} & \ood{\underline{11.30}} & \underline{-0.79} & \ood{\underline{12.84}} & \ood{\underline{28.03}} & \underline{-0.24} & \ood{\underline{7.97}} & \ood{\underline{18.83}} \\
            & & CAC & -0.28 & \ood{1.48} & \ood{-1.77} & 1.10 & \ood{5.64} & \ood{0.95} & 1.17 & \ood{1.76} & \ood{-2.25} \\
            & & DNN & -1.82 & \ood{-3.80} & \ood{-11.19} & -0.33 & \ood{-2.95} & \ood{-13.09} & -0.10 & \ood{-4.04} & \ood{-13.27} \\ \cmidrule(lr){3-12}
            & & proposed & \underline{0.05} & \ood{\underline{1.19}} & \ood{\underline{1.89}} & \underline{2.08} & \ood{\underline{6.02}} & \ood{\underline{8.20}} & \textbf{\underline{-0.22}} & \ood{\textbf{\underline{0.03}}} & \ood{\textbf{\underline{0.69}}} \\ 
            \midrule \bottomrule
    \end{tabular}
    \label{tab:overall-morbidity}
\end{table}

Table \ref{tab:overall-morbidity} describes the gap between biological and chronological ages depending on the overall health status of subjects, where the results depending on the type of diseases are available in Appendix \ref{appsec:morbidity_gap}.
Negative values of gap indicate that subjects are healthier than people who have the same chronological age on average, and positive gap values imply the contrary case.
If the biological ages are well estimated, the healthy subjects should have negative gap values while that for the average subjects might be closer to zero.
More importantly, agreeable biological ages are required to point the illness of subjects up, which can be represented as positive gap values.

\begin{figure}
    \centering
    \begin{subfigure}[t]{.4\textwidth}
        \includegraphics[width=\textwidth]{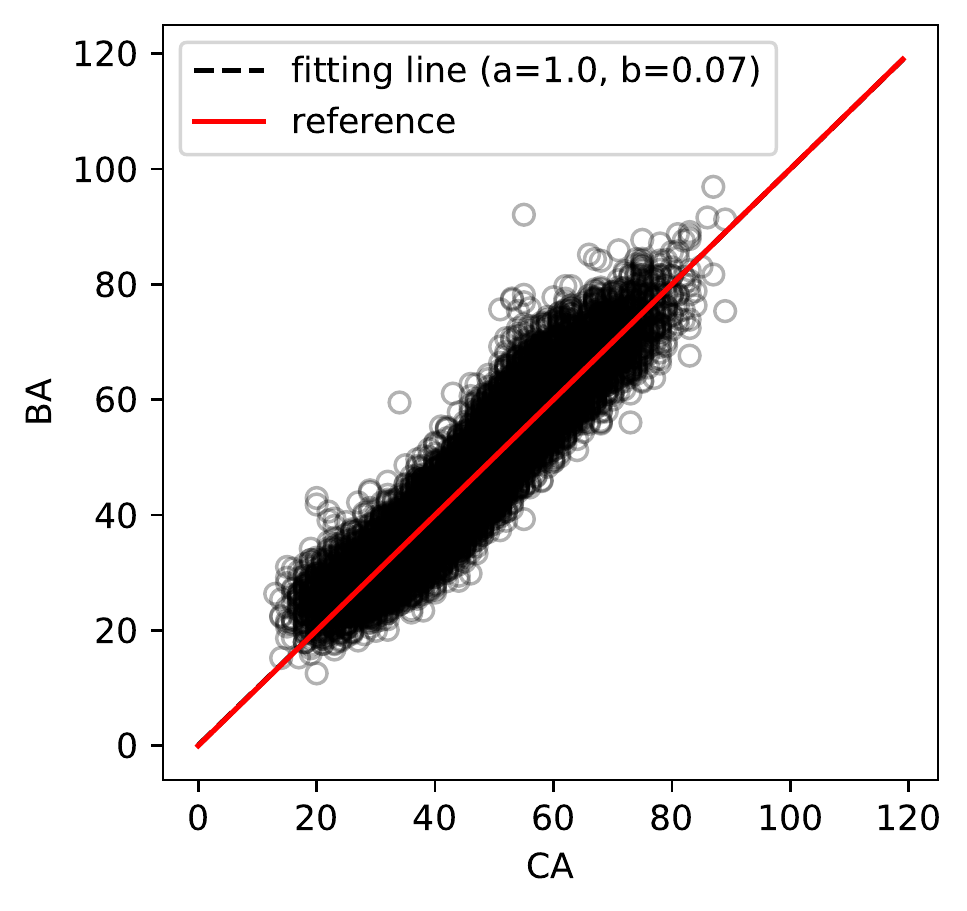}
        \caption{KDM}
    \end{subfigure}
    \begin{subfigure}[t]{.4\textwidth}
        \includegraphics[width=\textwidth]{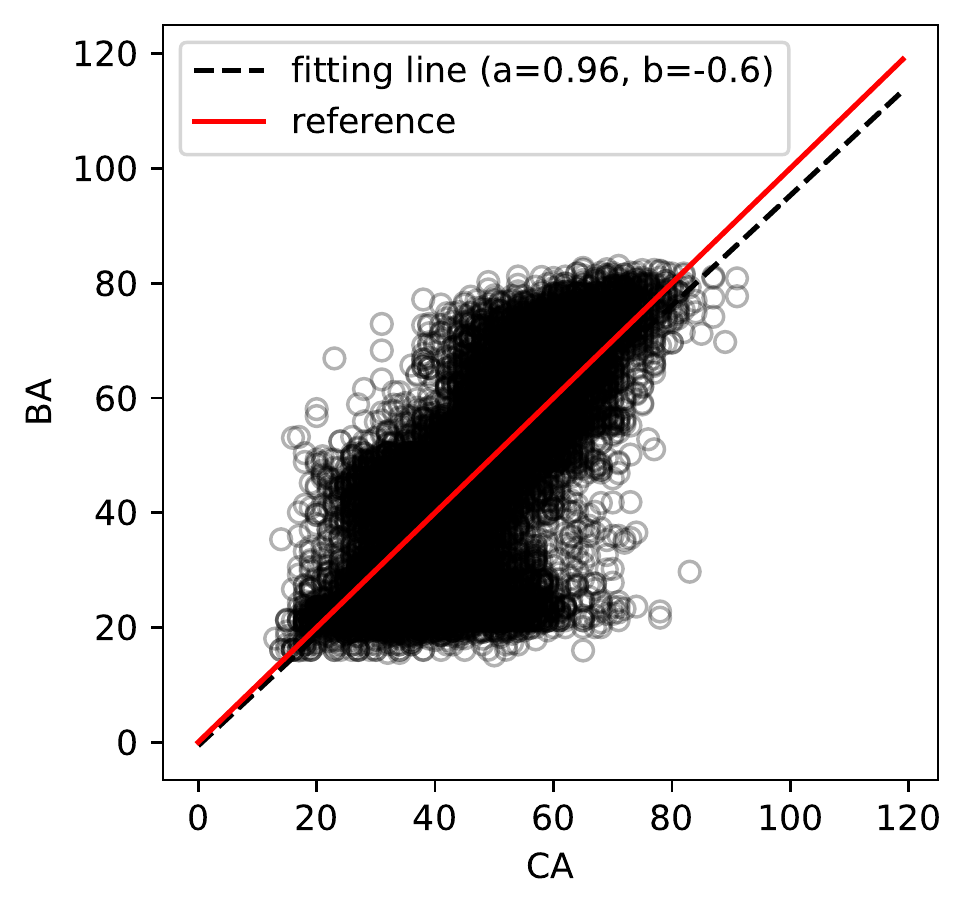}
        \caption{CAC}
    \end{subfigure}
    \begin{subfigure}[t]{.4\textwidth}
        \includegraphics[width=\textwidth]{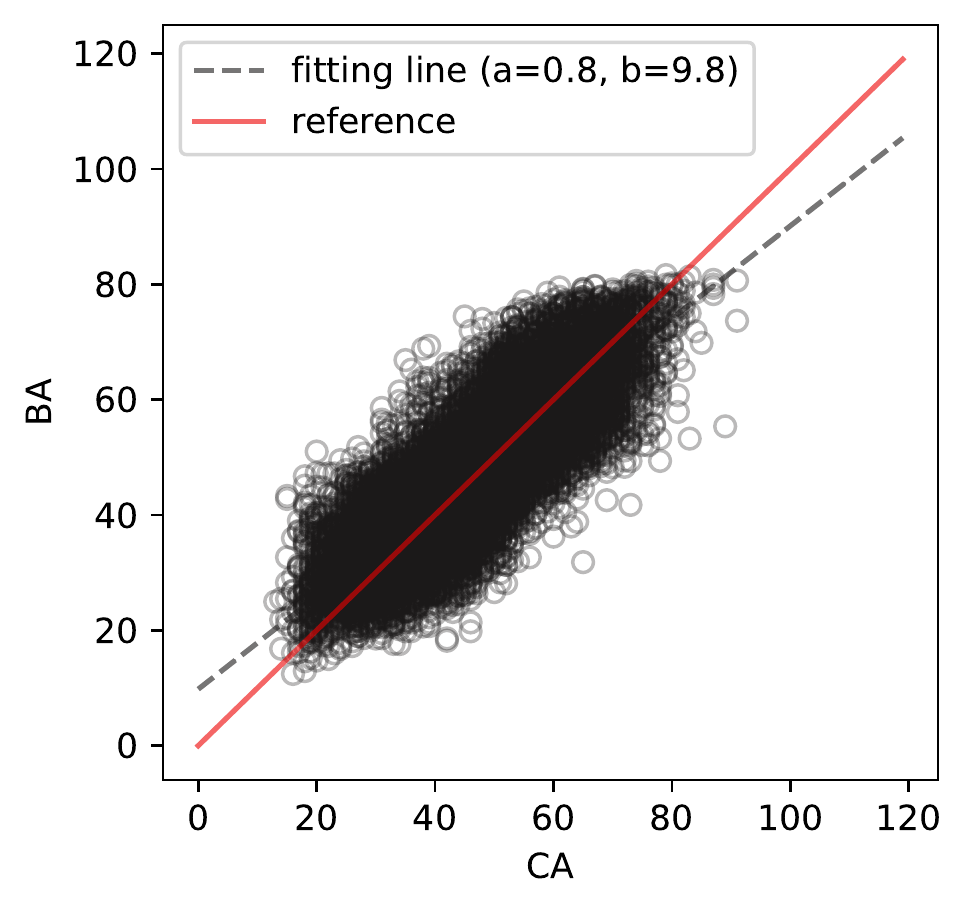}
        \caption{DNN}
    \end{subfigure}
    \begin{subfigure}[t]{.4\textwidth}
        \includegraphics[width=\textwidth]{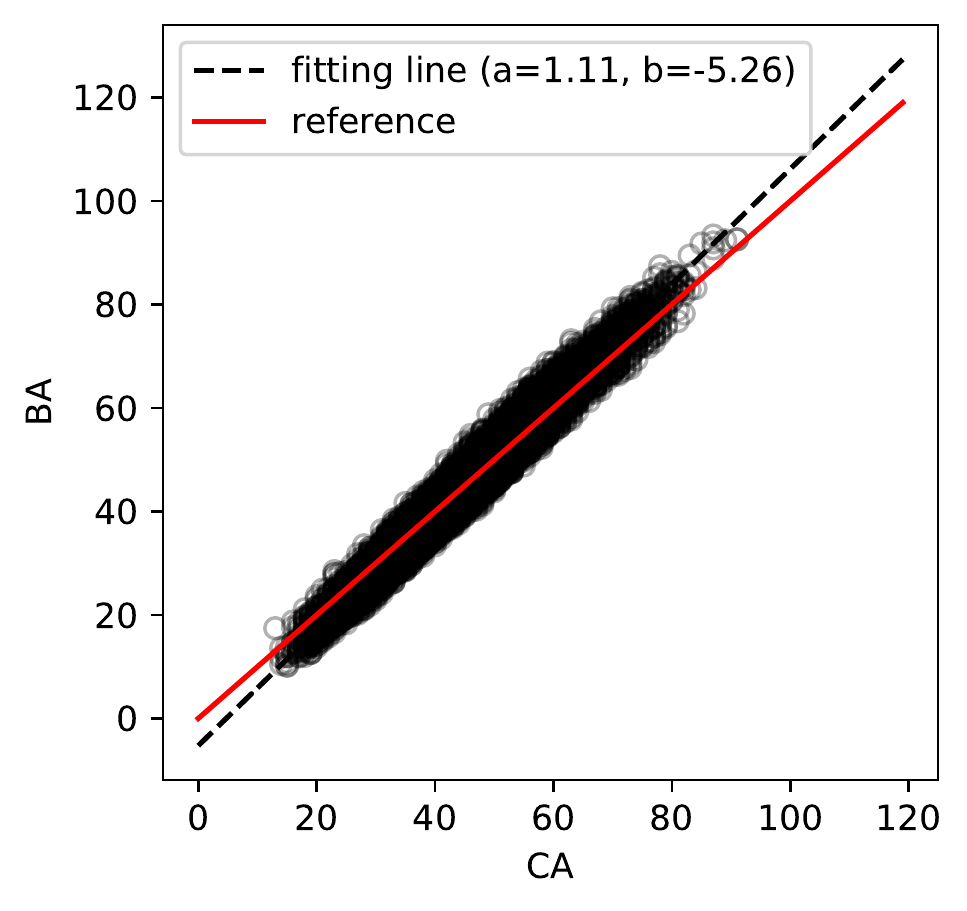}
        \caption{proposed}
    \end{subfigure}
    \caption{Estimated biological ages (BAs) of the baselines and proposed model toward chronological ages (CAs). The test results of all three runs are shown here for female-\textit{average}-\lfeature\ case. The dashed black lines are fitted to the CA and obtained BA as $BA=a\times CA + b$, and the red lines correspond to the $BA=CA$ case.}
    \label{fig:scatter-average-f}
\end{figure}

\begin{figure}
    \centering
    \begin{subfigure}[t]{0.8\textwidth}
        \includegraphics[width=\textwidth]{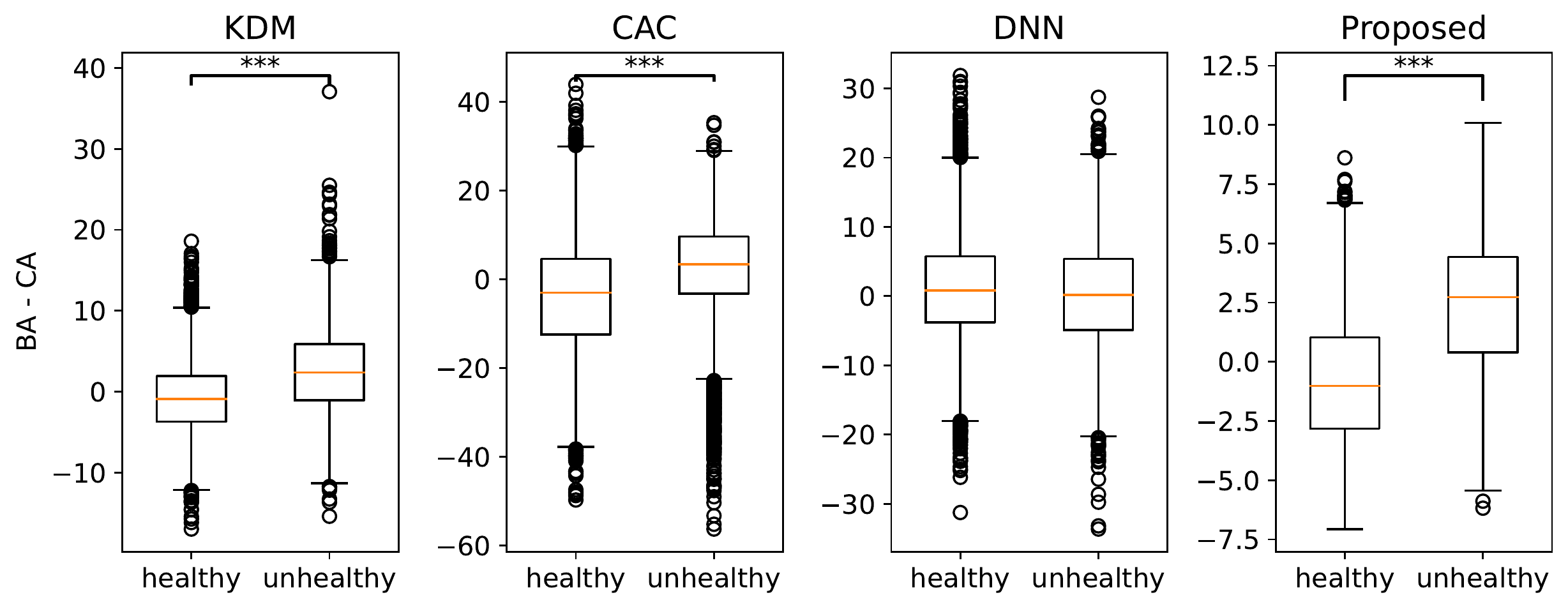}
        \caption{female-\textit{average}-\lfeature}
    \end{subfigure}
    \begin{subfigure}[t]{.8\textwidth}
        \includegraphics[width=\textwidth]{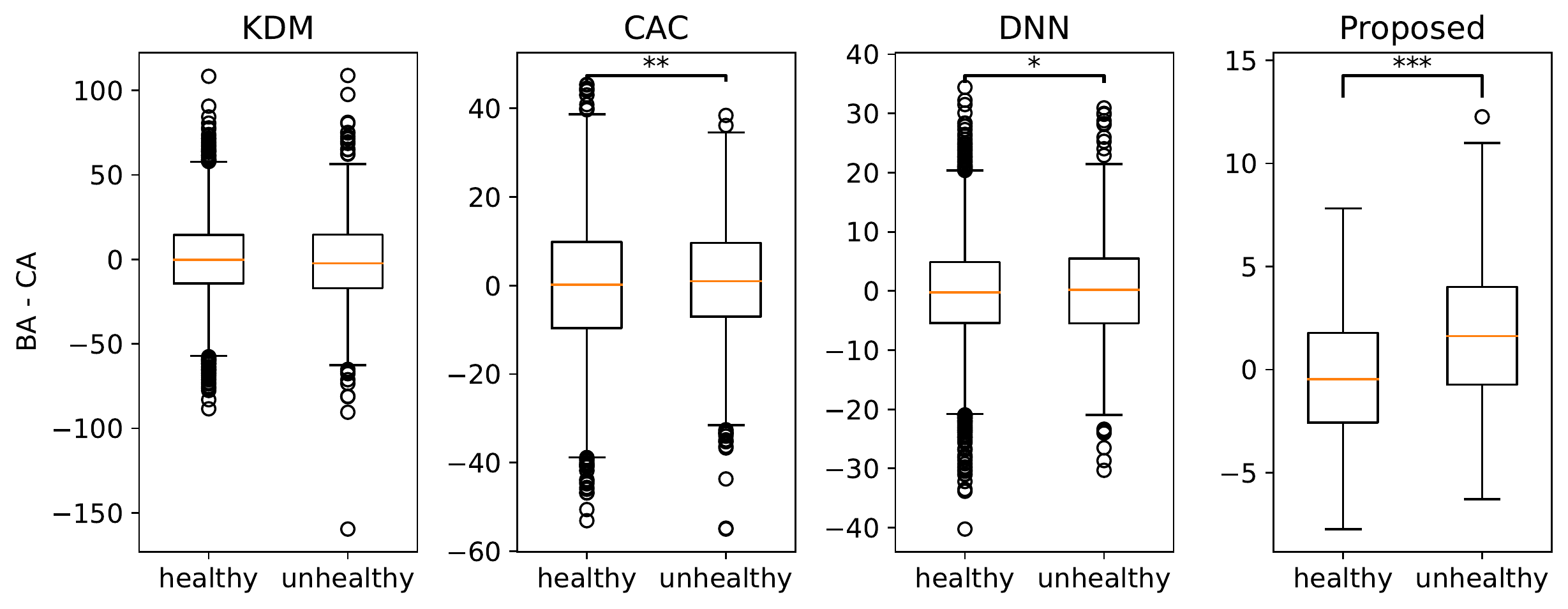}
        \caption{male-\textit{normal}-\lfeature}
    \end{subfigure}
    \caption{Boxplots of the gap between biological and chronological ages depending on the health status of subjects (healthy vs. unhealthy). The statistical significance of gap differences is evaluated by t-test for two independent samples ($^*$: $p$-value $< 0.05$, $^{**}$: $p$-value $< 0.01$, $^{***}$: $p$-value $< 0.001$).}
    \label{fig:all-morbidity-boxplot}
\end{figure}

The results in Table \ref{tab:overall-morbidity} demonstrate that the proposed model robustly estimates the health states of subjects.
Exceptional cases are observed only when the model was trained using \textit{normal} and \textit{super normal} populations, where the model had no chance to explore the morbidity information due to the inherent limitation of the training population. 
In contrast, baseline models fail to reflect the difference in health states in most cases.
KDM and CAC seem to succeed in a few cases, for instance, when the training population is female and \textit{average}, and the \lfeature\ feature set is employed.
However, as we can see from Figures \ref{fig:scatter-average-f} and \ref{fig:all-morbidity-boxplot}, the baseline models resulted in biological ages with overly high variability even for the in-distribution test data.
Particularly, DNN fails to capture the morbidity information of subjects due to its substantial dependency on chronological ages.
Besides resulting in smaller gap values for out-of-distribution data that is not convincing, DNN tends to underestimate the morbidity.
For example, DNN yields the average gap value of 0.08 for unhealthy subjects in female-\textit{average}-\lfeature\ case whereas the healthy and average subjects have average gap values of 0.85 and 1.09, respectively.  

Another interesting observation from Figure \ref{fig:scatter-average-f} is that the proposed model tends to estimate the biological ages of the old population as slightly larger than that of the young generation, i.e., $a > 1$. 
In contrast, CAC and DNN have smaller slope values (0.96 and 0.80, respectively), which is illogical considering that the young population has been exposed to improved nutrition and healthcare services.
One of the possible explanations is that as CAC and DNN were trained to predict the chronological ages of subjects without considering the imbalanced distribution of chronological ages, the models tend to predict the biological age as close to the mean chronological age (45.6 in the training data for Figure \ref{fig:scatter-average-f}) to achieve the learning objective. 


Comparing results of the proposed model obtained with different training populations, we can note that the model trained by using \textit{whole} and \textit{average} population has a better capability of discriminating the health states of subjects\footnote{For instance, although the relationship between healthy, average, and unhealthy subjects is well calibrated, the average gap value for healthy subjects is positive in male-\textit{super normal}-\lfeature, male-\textit{super normal}-\mfeature, male-\textit{super normal}-\sfeature, female-\textit{normal}-\mfeature, female-\textit{super normal}-\lfeature, and female-\textit{super normal}-\mfeature\ cases.}, which is also shown as elaborated differences between gap values in Table \ref{tab:overall-morbidity}.
This is an expected outcome because the morbidity contrastive loss is not adoptable when the model is trained by using \textit{normal} and \textit{super normal} populations because the subjects with the disease are not included in the population groups.
Moreover, it can be observed that CAC with no guidance for considering morbidity demonstrates improved discriminability when a larger training population is utilized.
Therefore, as we disclosed in Section \ref{sec:intro}, the variety of training populations is crucial to obtain reliable biological ages. 


\begin{figure}[t]
    \centering
    \begin{subfigure}[t]{.45\textwidth}
        \includegraphics[width=\textwidth]{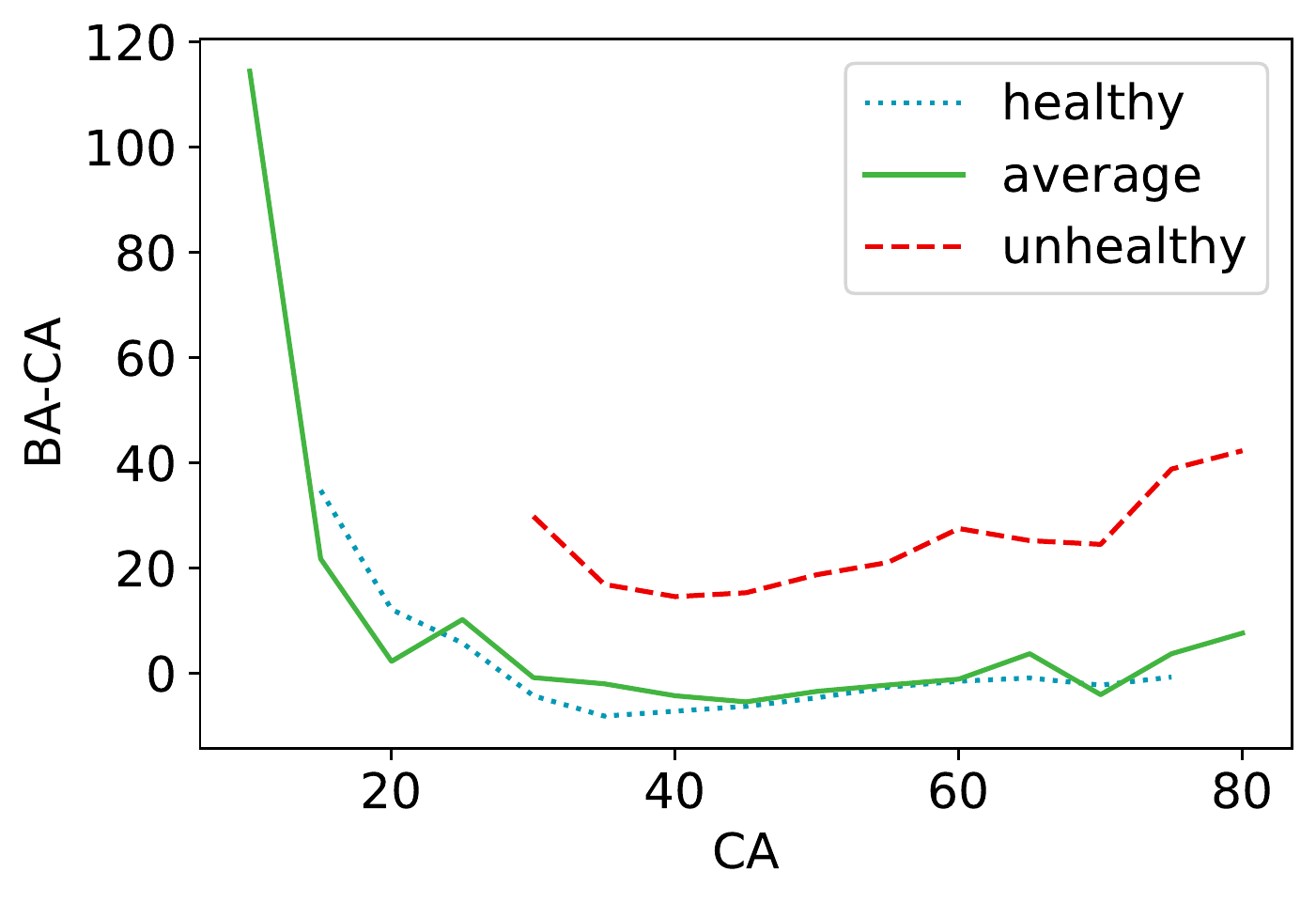}
        \caption{KDM}
    \end{subfigure}
    \begin{subfigure}[t]{.45\textwidth}
        \includegraphics[width=\textwidth]{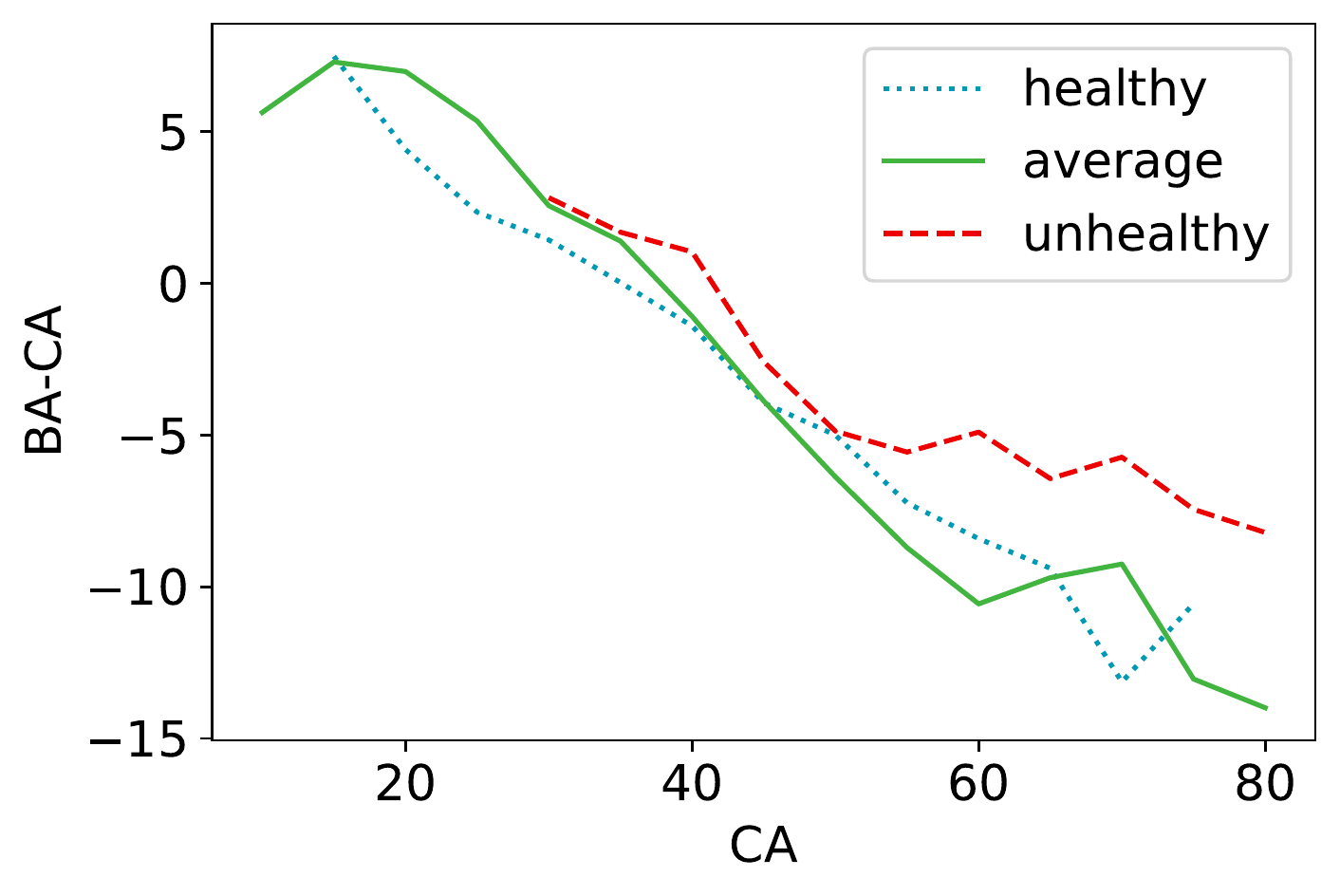}
        \caption{CAC}
    \end{subfigure}
    \begin{subfigure}[t]{.45\textwidth}
        \includegraphics[width=\textwidth]{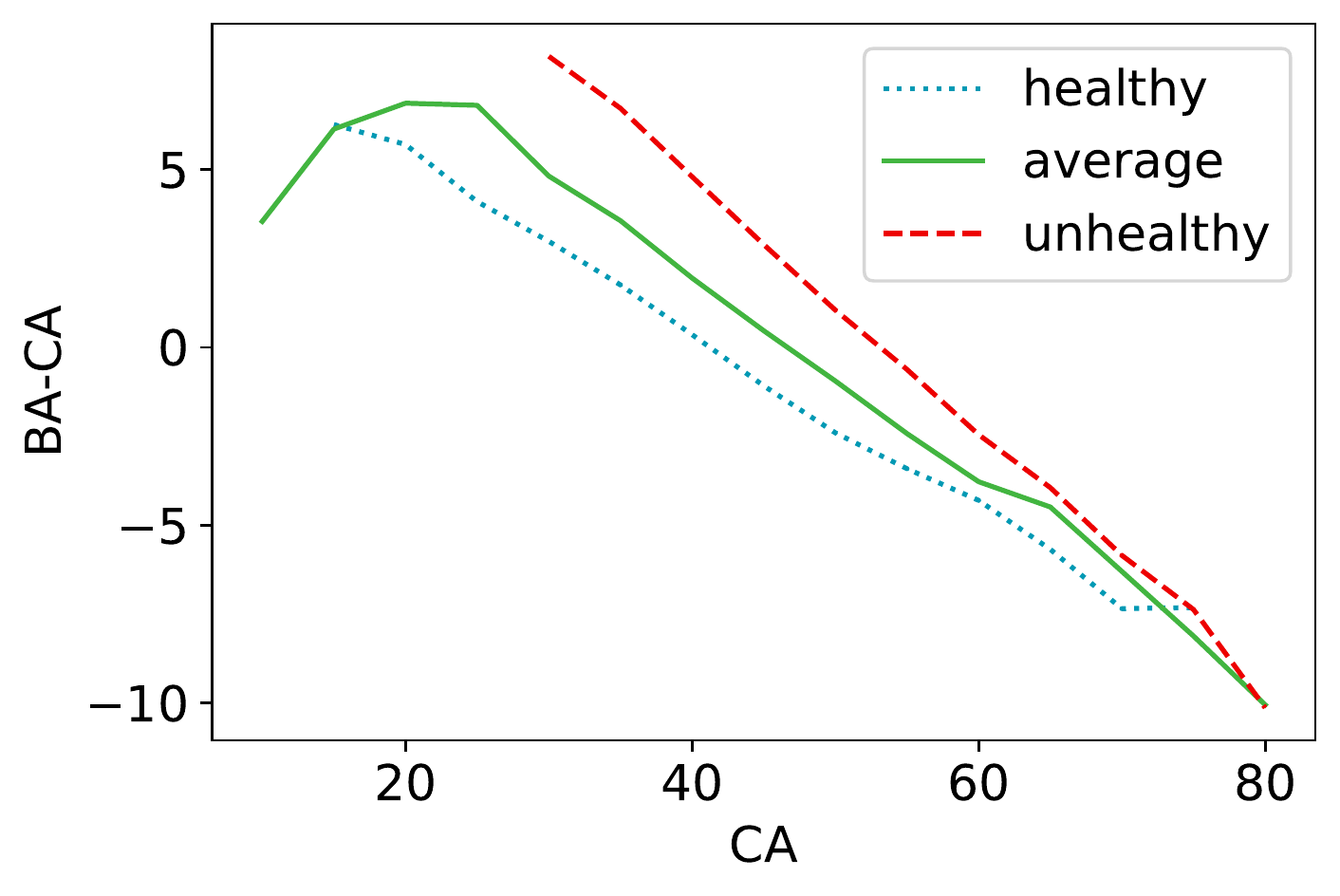}
        \caption{DNN}
    \end{subfigure}
    \begin{subfigure}[t]{.45\textwidth}
        \includegraphics[width=\textwidth]{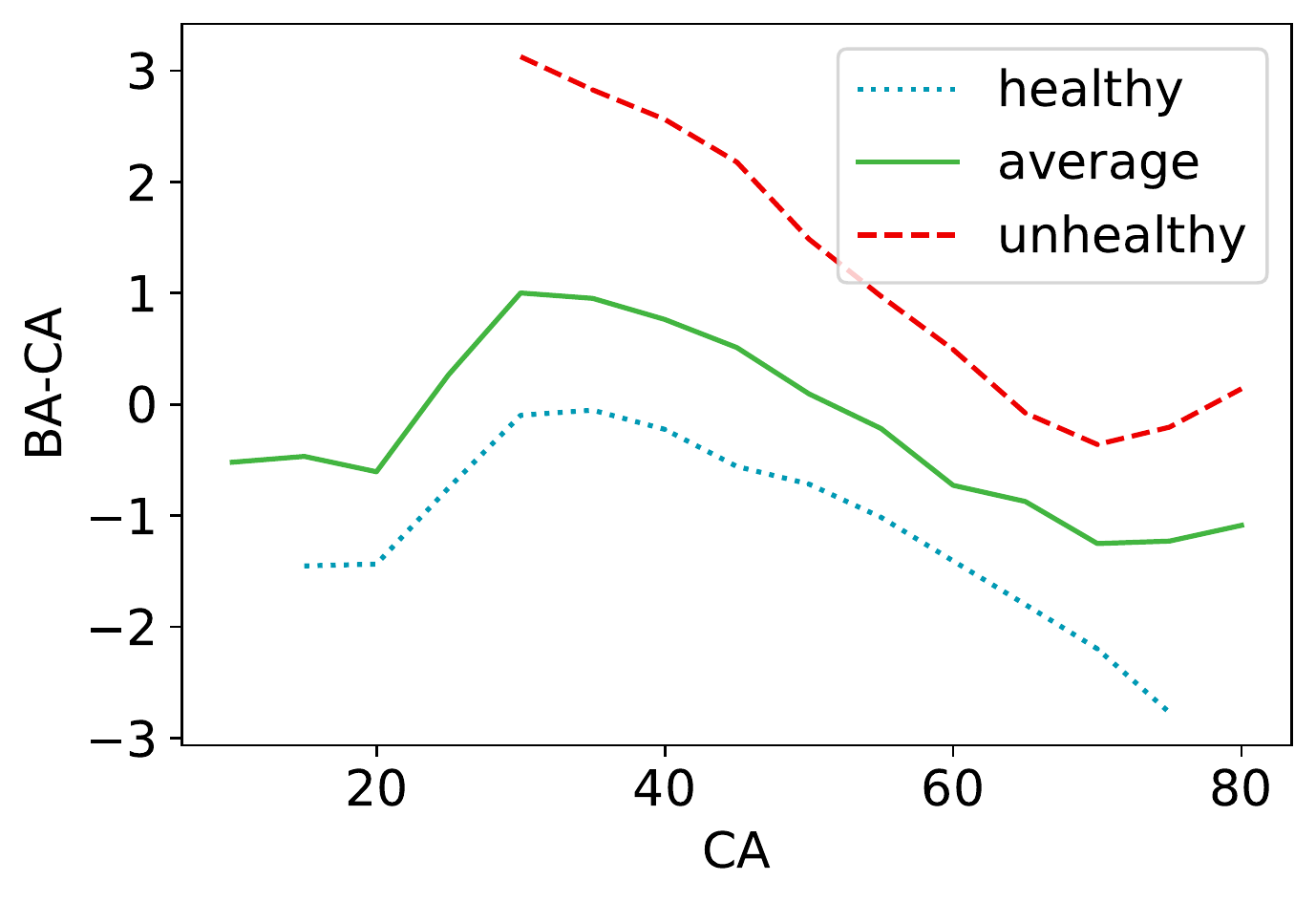}
        \caption{proposed}
    \end{subfigure}
    \caption{The gap between biological and chronological ages depending on the morbidity of \textbf{DM} in male-\textit{whole}-\lfeature\ case. The gap values are averaged for 5-year intervals, that is, the average gap value of subjects whose chronological ages are in a range of 40-44 is shown at $CA=40$. Green dotted lines indicate the results of healthy subjects, and red dashed lines describe the results of unhealthy subjects. The results of average subjects are illustrated as yellow solid lines. The colored areas cover from 0.25 to 0.75 quantiles.}
    \label{fig:all-l-morbidity-dm}
\end{figure}

\begin{figure}
    \centering
    \begin{subfigure}[t]{.45\textwidth}
        \includegraphics[width=\textwidth]{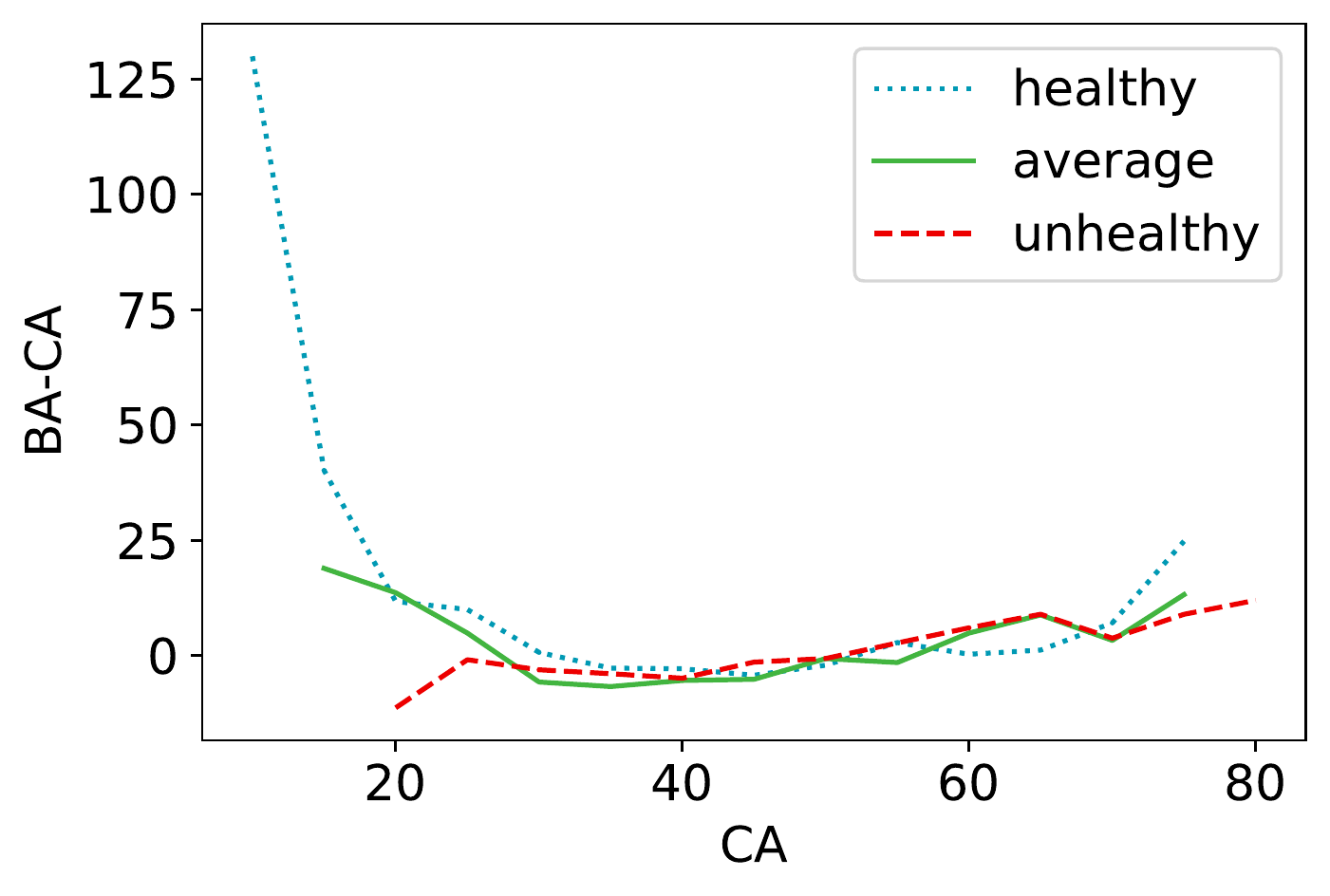}
        \caption{KDM}
    \end{subfigure}
    \begin{subfigure}[t]{.45\textwidth}
        \includegraphics[width=\textwidth]{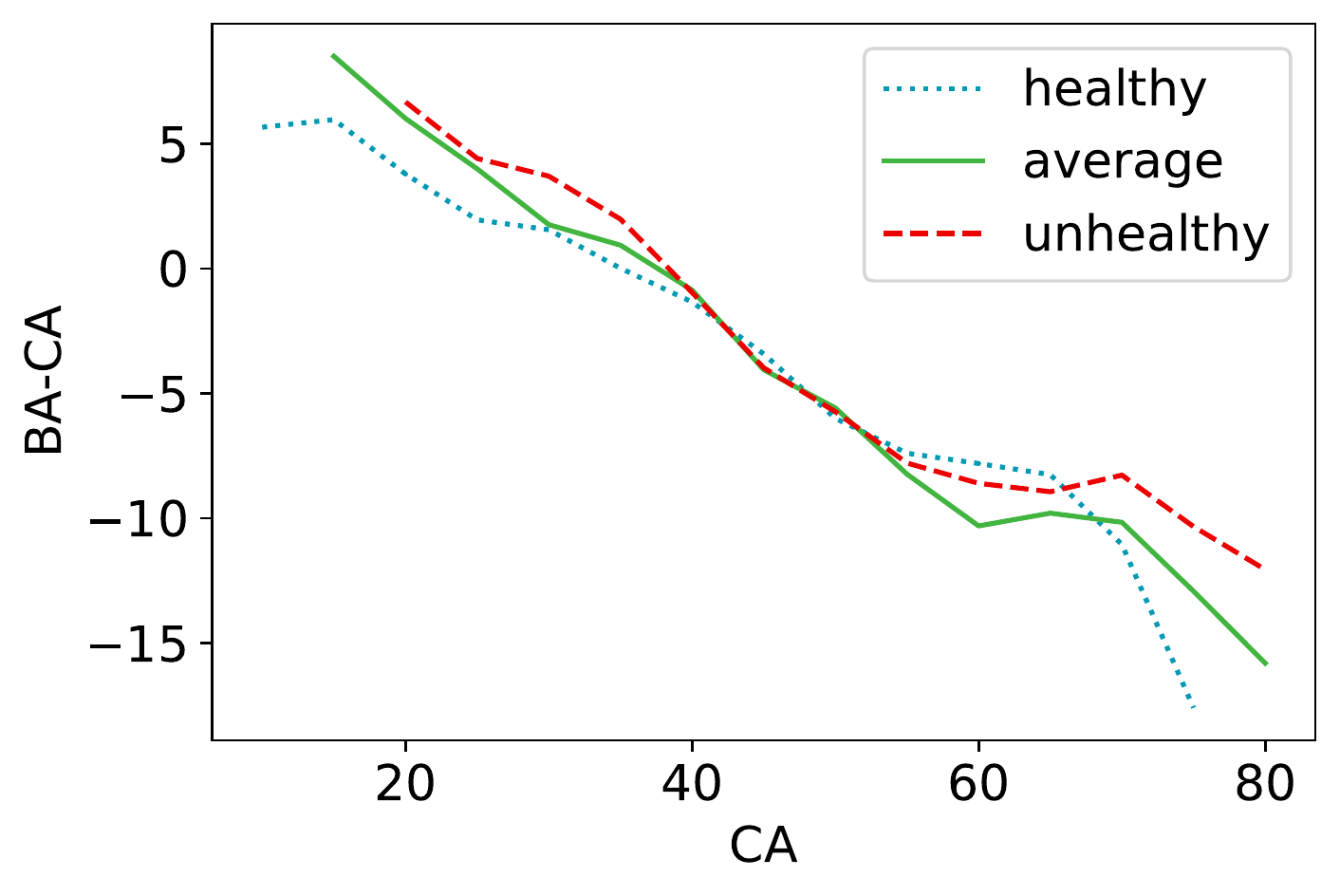}
        \caption{CAC}
    \end{subfigure}
    \begin{subfigure}[t]{.45\textwidth}
        \includegraphics[width=\textwidth]{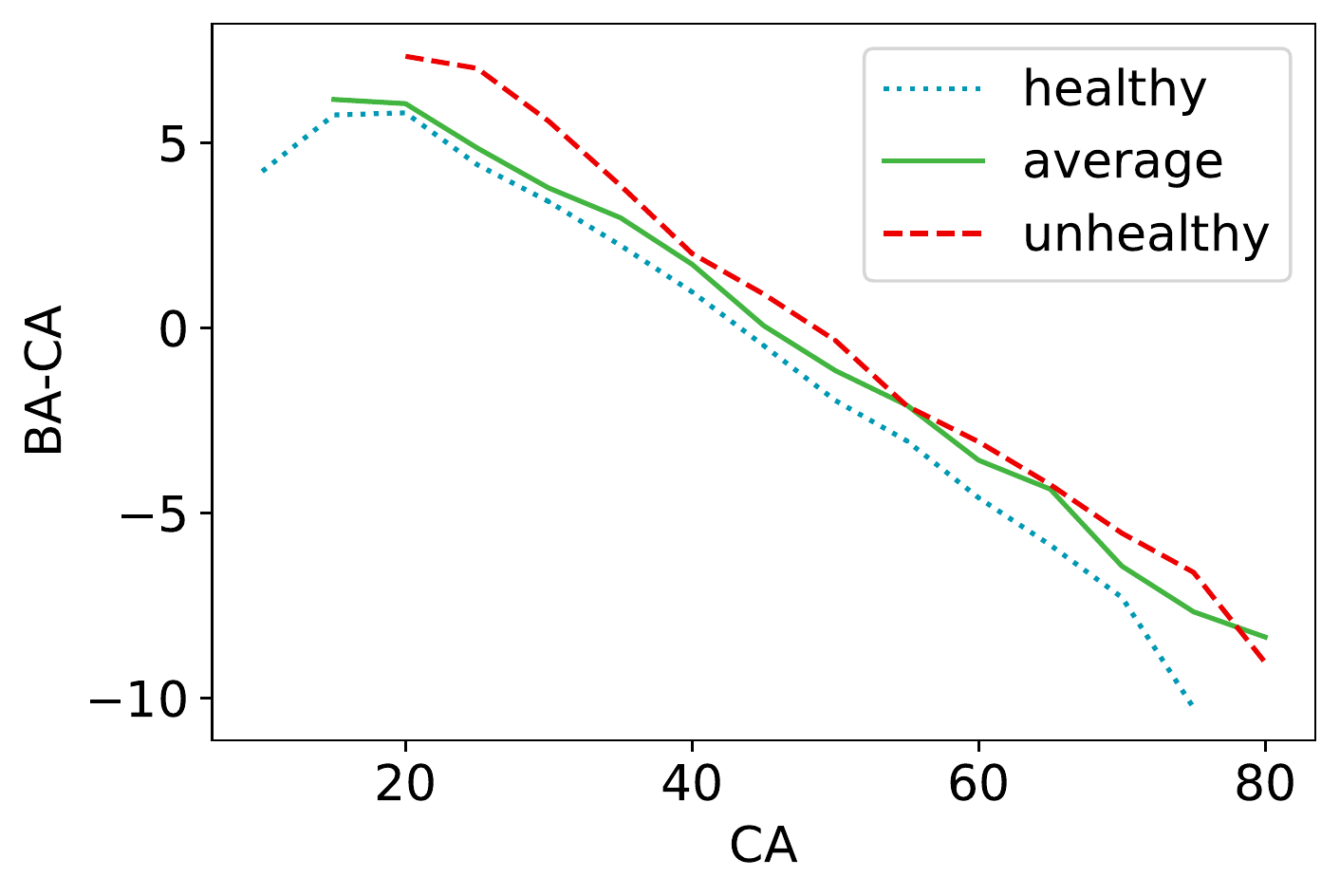}
        \caption{DNN}
    \end{subfigure}
    \begin{subfigure}[t]{.45\textwidth}
        \includegraphics[width=\textwidth]{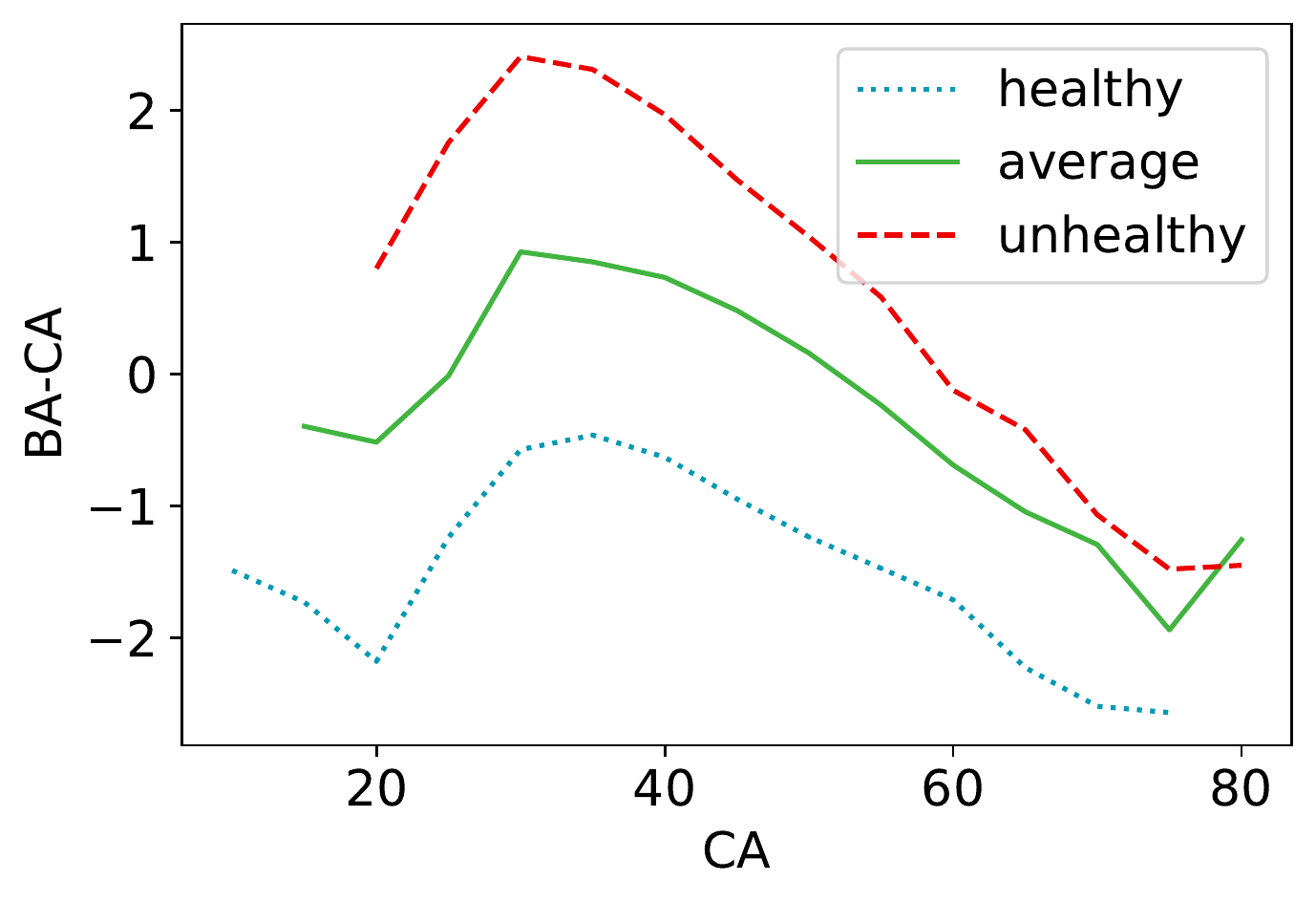}
        \caption{proposed}
    \end{subfigure}
    \caption{The gap between biological and chronological ages depending on the morbidity of \textbf{HBP} in male-\textit{whole}-\lfeature\ case. }
    \label{fig:all-l-morbidity-hbp}
\end{figure}

\begin{figure}
    \centering
    \begin{subfigure}[t]{.45\textwidth}
        \includegraphics[width=\textwidth]{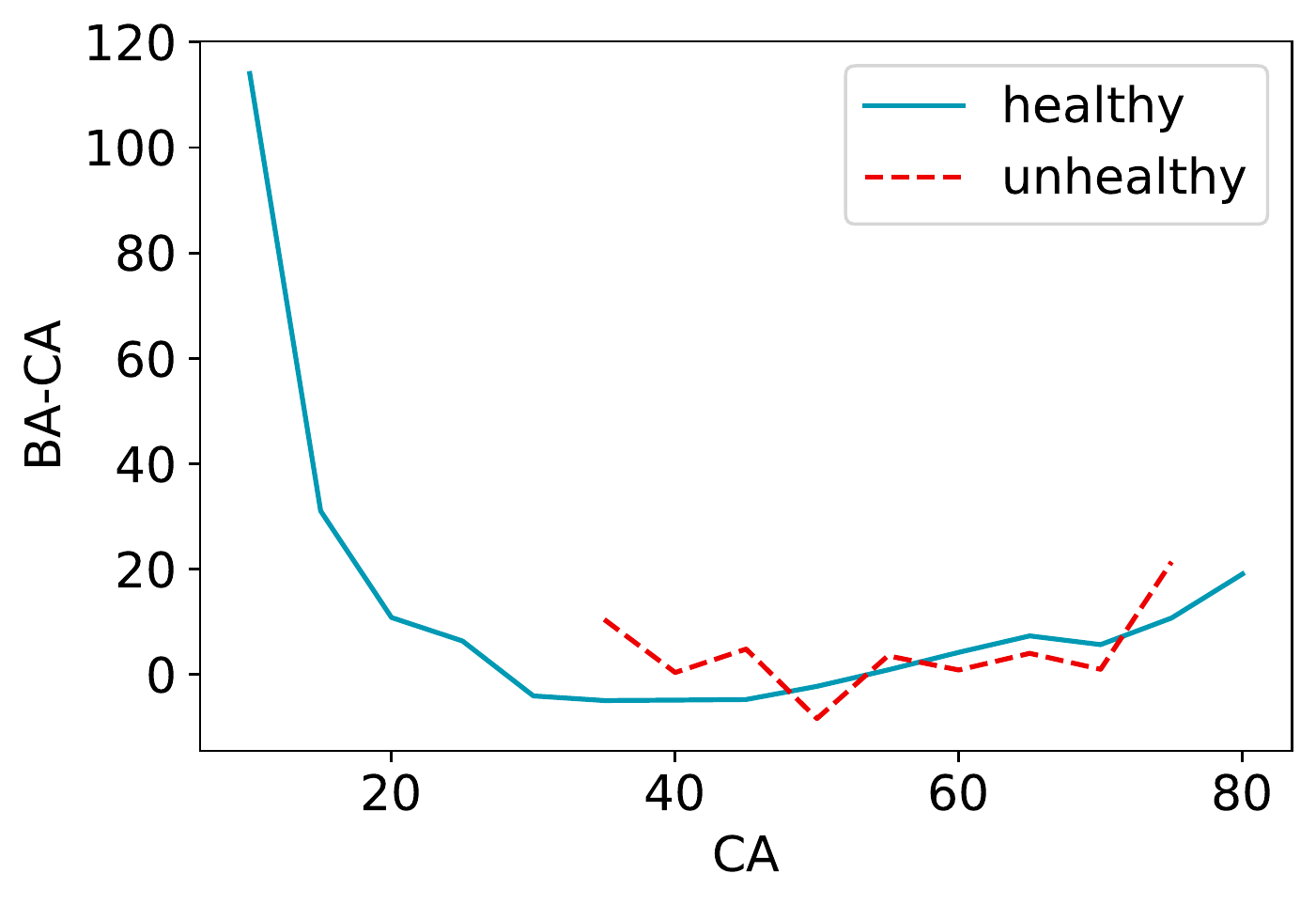}
        \caption{KDM}
    \end{subfigure}
    \begin{subfigure}[t]{.45\textwidth}
        \includegraphics[width=\textwidth]{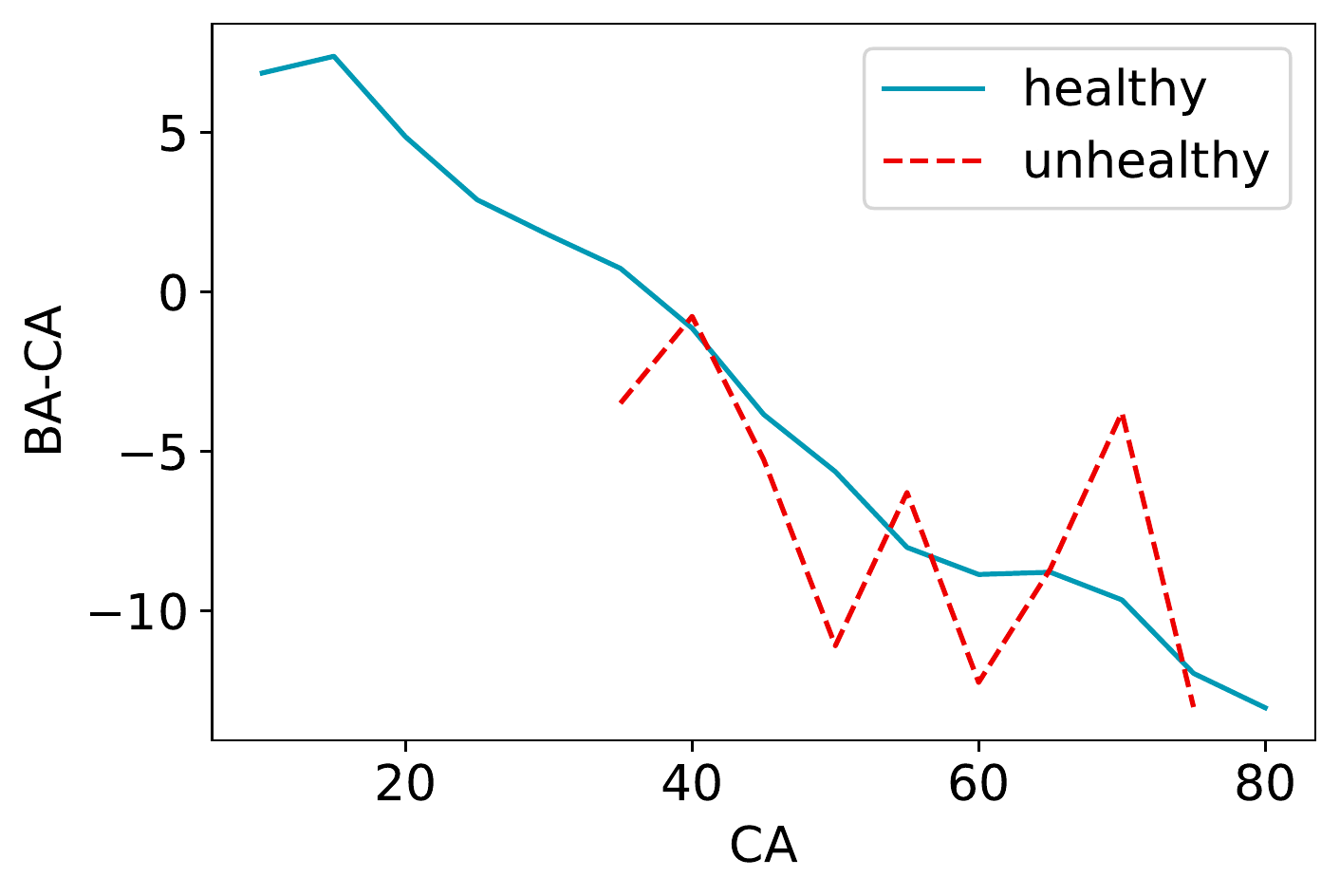}
        \caption{CAC}
    \end{subfigure}
    \begin{subfigure}[t]{.45\textwidth}
        \includegraphics[width=\textwidth]{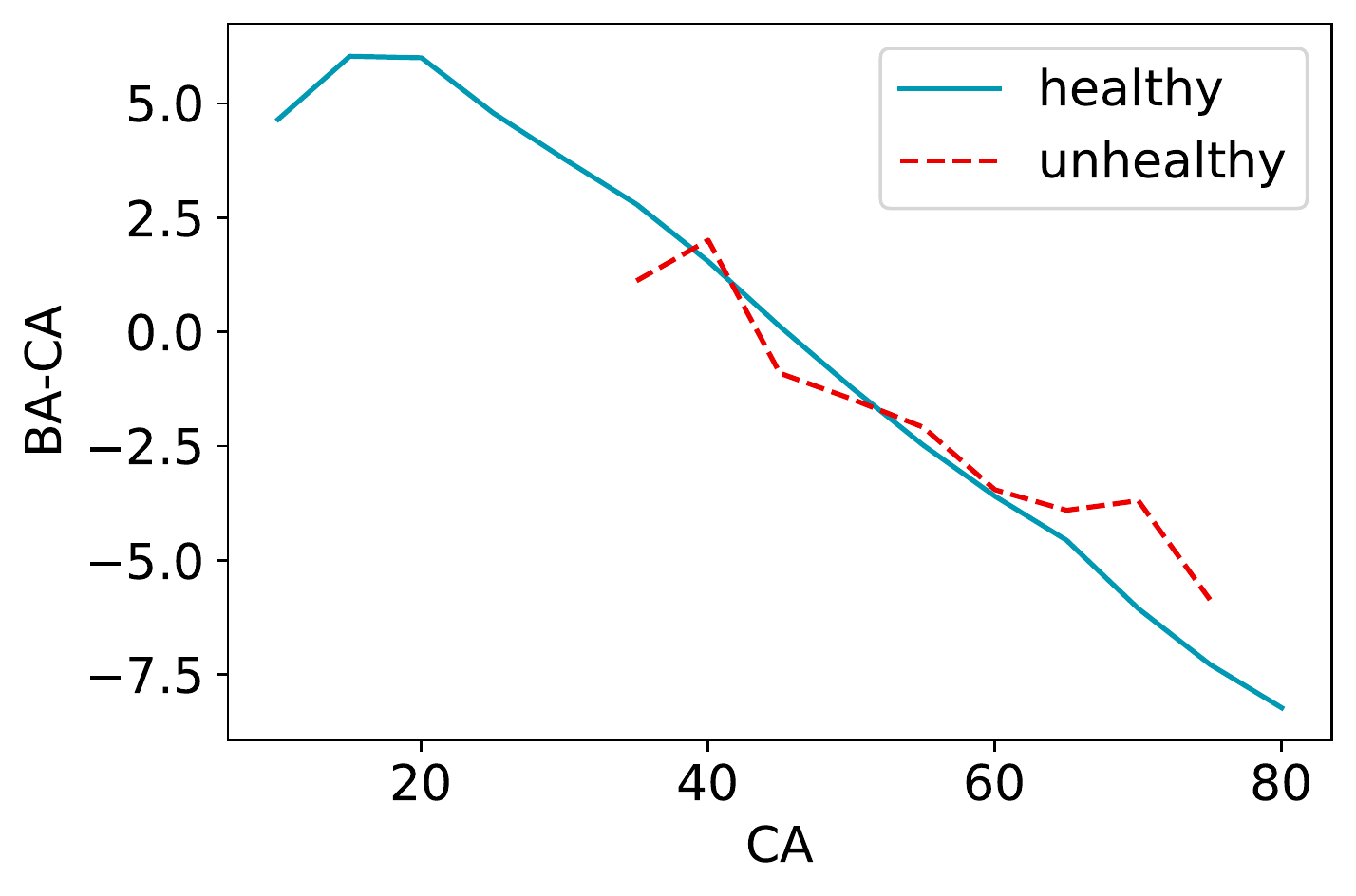}
        \caption{DNN}
    \end{subfigure}
    \begin{subfigure}[t]{.45\textwidth}
        \includegraphics[width=\textwidth]{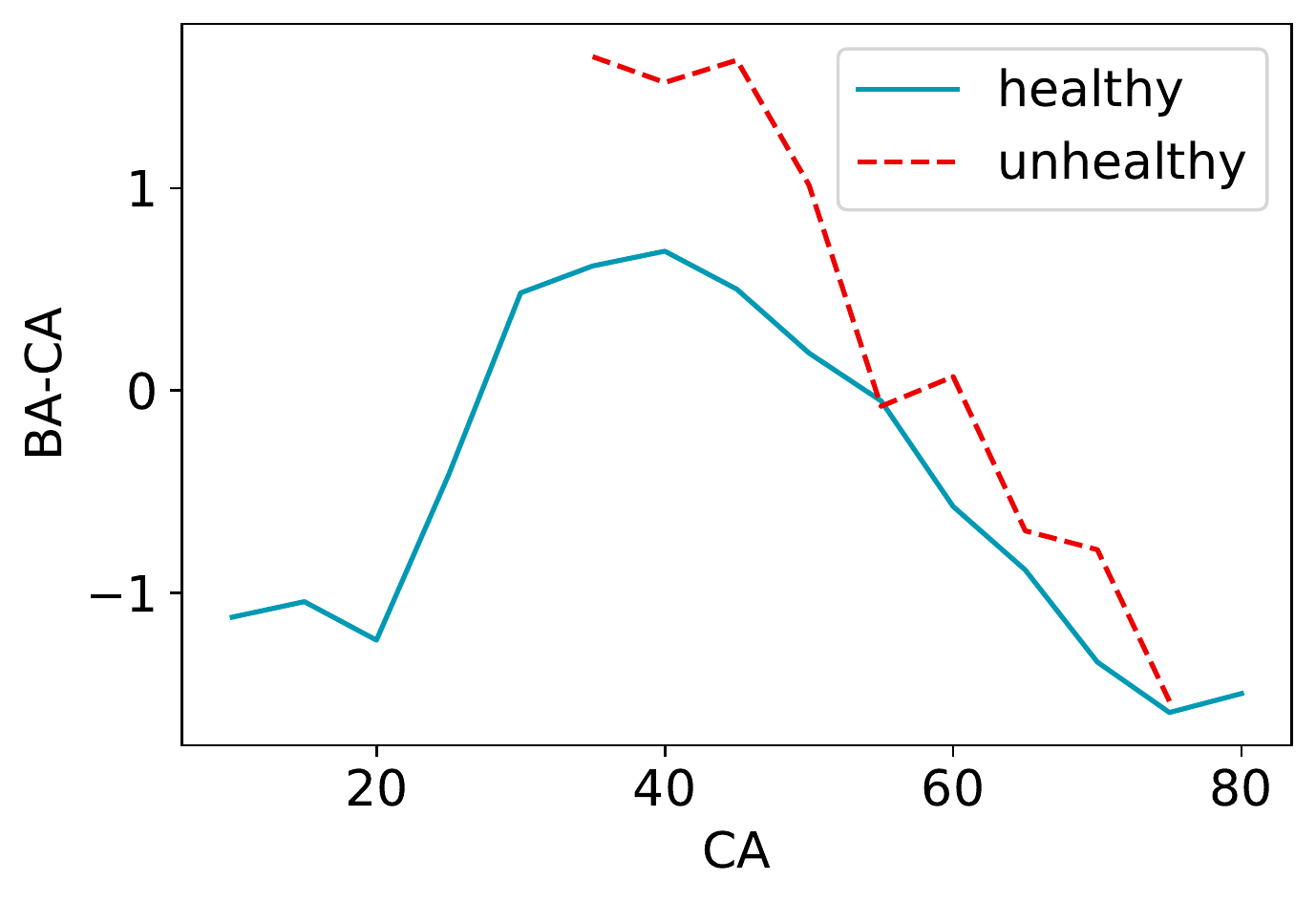}
        \caption{proposed}
    \end{subfigure}
    \caption{The gap between biological and chronological ages depending on the morbidity of \textbf{cancer} in male-\textit{whole}-\lfeature\ case. Green lines indicate the results of healthy subjects, and red dashed lines describe the results of unhealthy subjects.}
    \label{fig:all-l-morbidity-ca}
\end{figure}

The obtained biological ages are evaluated based on subjects' morbidity of DM, HBP, and cancer for the male-\textit{whole}-\lfeature\ case in Figures \ref{fig:all-l-morbidity-dm}, \ref{fig:all-l-morbidity-hbp}, and \ref{fig:all-l-morbidity-ca}, respectively.
Note that the information on cancer (including CVD and CVA in Appendix \ref{appsec:morbidity_age_gap}) was not given to the proposed model during training, however, is employed here to demonstrate that the model considers general health states. 
The proposed model consistently shows considerable discriminability for the morbidity of subjects regardless of the type of disease. 
The baseline models fail to distinguish different states of diseases, which is apparent from their large variance of gap values. 
We also can discover other evidence of the unstable biological estimation of CAC and DNN from Figures \ref{fig:all-l-morbidity-dm}, \ref{fig:all-l-morbidity-hbp}, and \ref{fig:all-l-morbidity-ca}, i.e., their gap values apparently decrease as the chronological age increases.
The results of female subjects and the other diseases, DLP, MS, CVD, and CVA, can be found in Appendix \ref{appsec:morbidity_age_gap}.

\subsection{Relationship with mortality}
\label{sec:exp-mortality}

\begin{table}
    \centering
    \caption{Linear regression results for time-to-death using chronological and biological ages for \textit{whole} population. 
    The relationship with mortality is evaluated using the slope of the linear regression line, R-squared score ($R^2$) between the regression results and true time-to-death, and Pearson correlation coefficient (PCC) between the ages and time-to-death.
    Bold slope values indicate the most statistically significant case, i.e., the obtained slope value is considerably smaller than zero. 
    The $R^2$ and PCC results with bold numbers correspond to the best cases. 
    The slope and PCC are calculated using gap values for the proposed (gap) model. }
    \small
    \setlength\tabcolsep{5pt}
    \def\arraystretch{0.8}
    \begin{tabular}{c c c r r r r r r}
        \toprule \midrule
        
        \multirow{2}{*}{Population} & \multirow{2}{*}{Feature set} & \multirow{2}{*}{Model} & \multicolumn{2}{c}{Slope} & \multicolumn{2}{c}{$R^2$} & \multicolumn{2}{c}{PCC} \\ \cmidrule(lr){4-9}
        & & & CA & BA & CA & BA & CA & BA \\ \midrule
        
        \multirow{15}{*}{M} & \multirow{5}{*}{\lfeature} &
        KDM (BA) & \multirow{5}{*}{-4.683}
        & \textbf{-6.092}$^{***}$ & \multirow{5}{*}{0.001} & 
        0.048 & \multirow{5}{*}{-0.032} & -0.219 \\
        & & CAC (BA) & & \textbf{-13.487}$^{***}$ & & 0.030 & & -0.172 \\
        & & DNN (BA) & & -3.732$^{\quad\;}$ & & 0.001 & & -0.024 \\
        & & proposed (BA) & & \textbf{-19.938}$^{***}$ & & 0.020 & & -0.141 \\ \cmidrule(lr){2-9}
        
        & \multirow{5}{*}{\mfeature} &
        KDM & \multirow{5}{*}{-1.601} 
        & \textbf{-5.922}$^{***}$ & \multirow{5}{*}{0.000} & 0.017 &
        \multirow{5}{*}{-0.012} & -0.130 \\
        & & CAC & & -8.886$^{**\; }$ & & 0.009 & & -0.093\\
        & & DNN & & -14.674$^{**\; }$ & & 0.009 & & -0.097 \\
        & & proposed (BA) & & \textbf{-13.812}$^{***}$ & & 0.011 & &  -0.105 \\ 
        & & proposed (gap) & & \textbf{-139.756}$^{***}$ & & \textbf{0.101} & & \textbf{-0.317} \\ \cmidrule(lr){2-9}
        
        & \multirow{5}{*}{\sfeature} &
        KDM & \multirow{5}{*}{6.959$^{\;}$} 
        & \textbf{-6.498}$^{***}$ & \multirow{5}{*}{0.002} & 0.018 &  \multirow{5}{*}{0.048} & -0.133 \\
        & & CAC & & -2.863$^{\quad\; }$ & & 0.001 & & -0.031 \\
        & & DNN & & -6.953$^{\quad\; }$ & & 0.002 & & -0.046 \\
        & & proposed (BA) & & -5.946$^{\quad\;}$ & & 0.002 & & -0.045 \\ 
        & & proposed (gap) & & \textbf{-112.617}$^{***}$ & & \textbf{0.080} & & \textbf{-0.273} \\ \cmidrule(lr){1-9}

        \multirow{15}{*}{F} & \multirow{5}{*}{\lfeature} &
        KDM & \multirow{5}{*}{3.794} 
        & 4.867$^{\quad\;}$ & \multirow{5}{*}{0.001} & 0.001 & \multirow{5}{*}{0.029} & 0.037 \\
        & & CAC & & -3.779$^{\quad\; }$ & & 0.002 & & -0.046 \\
        & & DNN & & 2.654$^{\quad\; }$ & & 0.000 & & 0.017 \\
        & & proposed (BA) & & -3.823$^{\quad\;}$ & & 0.001 & & -0.032 \\ 
        & & proposed (gap) & & \textbf{-167.651}$^{***}$ & & \textbf{0.086} & & \textbf{-0.274} \\ \cmidrule(lr){2-9}
        
        & \multirow{5}{*}{\mfeature} &
        KDM & \multirow{5}{*}{13.708} 
        & 0.789$^{\quad\;}$ & \multirow{5}{*}{0.011} & 0.003 & \multirow{5}{*}{0.106} & 0.058 \\
        & & CAC & & 6.991$^{\quad\;}$ & & 0.005 & & 0.071 \\
        & & DNN & & 23.668$^{\quad\;}$ & & 0.026 & & 0.160 \\
        & & proposed (BA) & & 6.530$^{\quad\;}$ & & 0.003 & & 0.054 \\ 
        & & proposed (gap) & & \textbf{-152.662}$^{***}$ & & \textbf{0.074} & & \textbf{-0.221} \\ \cmidrule(lr){2-9}
        
        & \multirow{5}{*}{\sfeature} &
        KDM & \multirow{5}{*}{11.272} 
        & 4.196$^{\quad\;}$ & \multirow{5}{*}{0.007} & 0.028 & \multirow{5}{*}{0.082} & 0.167 \\
        & & CAC & & 11.086$^{\quad\;}$ & & 0.015 & & 0.122 \\
        & & DNN & & 23.056$^{\quad\;}$ & & 0.023 & & 0.153 \\
        & & proposed (BA) & & 2.601$^{\quad\;}$ & & 0.000 & & 0.019 \\ 
        & & proposed (gap) & & \textbf{-95.941}$^{***}$ & & \textbf{0.049} & & \textbf{-0.210} \\
        \midrule \bottomrule
    \multicolumn{9}{l}{\footnotesize{$^{*}:\ p$-value$ < 0.05,\ ^{**}:\ p$-value$ < 0.01,\ ^{***}:\ p$-value$ < 0.001$}} \\
    \end{tabular}
    \label{tab:whole-mortality}
\end{table}

It is expected that biological ages have predictive power for subjects' mortality; that is, a larger biological age should indicate a shorter time-to-death.
We compare the results of linear regression to the time-to-death using chronological age and estimated biological ages in Table \ref{tab:whole-mortality}. 
The biological age obtained from the proposed model are examined in two ways, i.e., the univariate linear regression using BA ($= \text{CA} + \text{gap}$) and multivariate linear regression using CA and gap.
The proposed model, particularly when the gap values are employed for evaluation, exhibits the best performance in all cases whereas the baseline models gain only marginal improvement by adopting the multivariate strategy (results can be found in Appendix \ref{appsec:gap_models}.) 
Among the baseline methods, KDM, which is known to be effective for mortality prediction, outperforms the others overall. 
This implies the effectiveness of the gap-based approach compared to the direct modeling of biological ages.

Unlike the previous results for morbidity where nothing shows superiority over the others among three different feature sets (\lfeature, \mfeature, and \sfeature), models trained with \lfeature\ and \mfeature\ sets demonstrate a better ability to predict mortality in male and female subjects, respectively.
This probably indicates that the gain by using additional information is meaningful despite its sparsity. 

\begin{figure}
    \centering
    \begin{subfigure}[t]{.45\textwidth}
        \includegraphics[width=\textwidth]{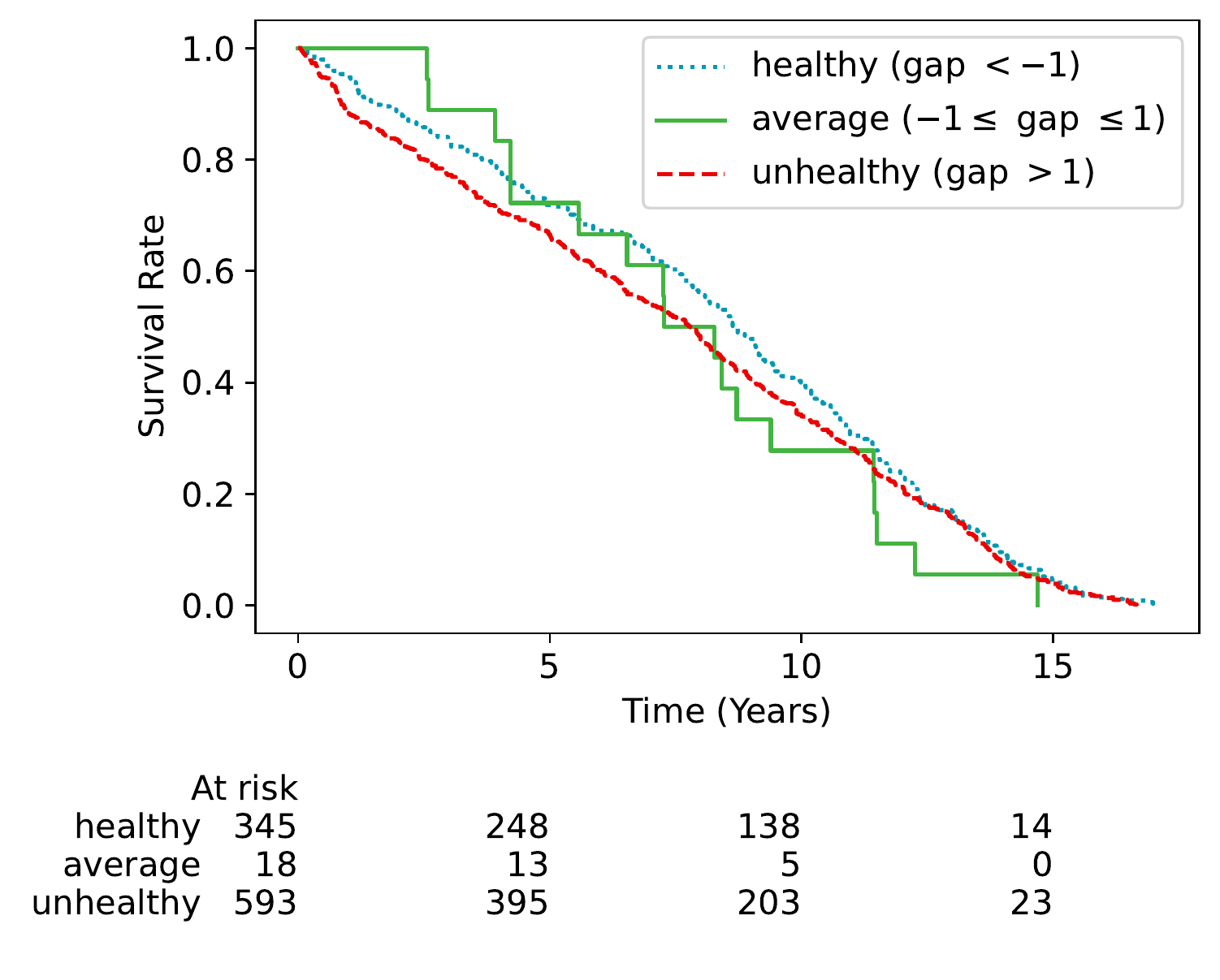}
        \caption{KDM ($p$-value=0.076)}
    \end{subfigure}
    \begin{subfigure}[t]{.45\textwidth}
        \includegraphics[width=\textwidth]{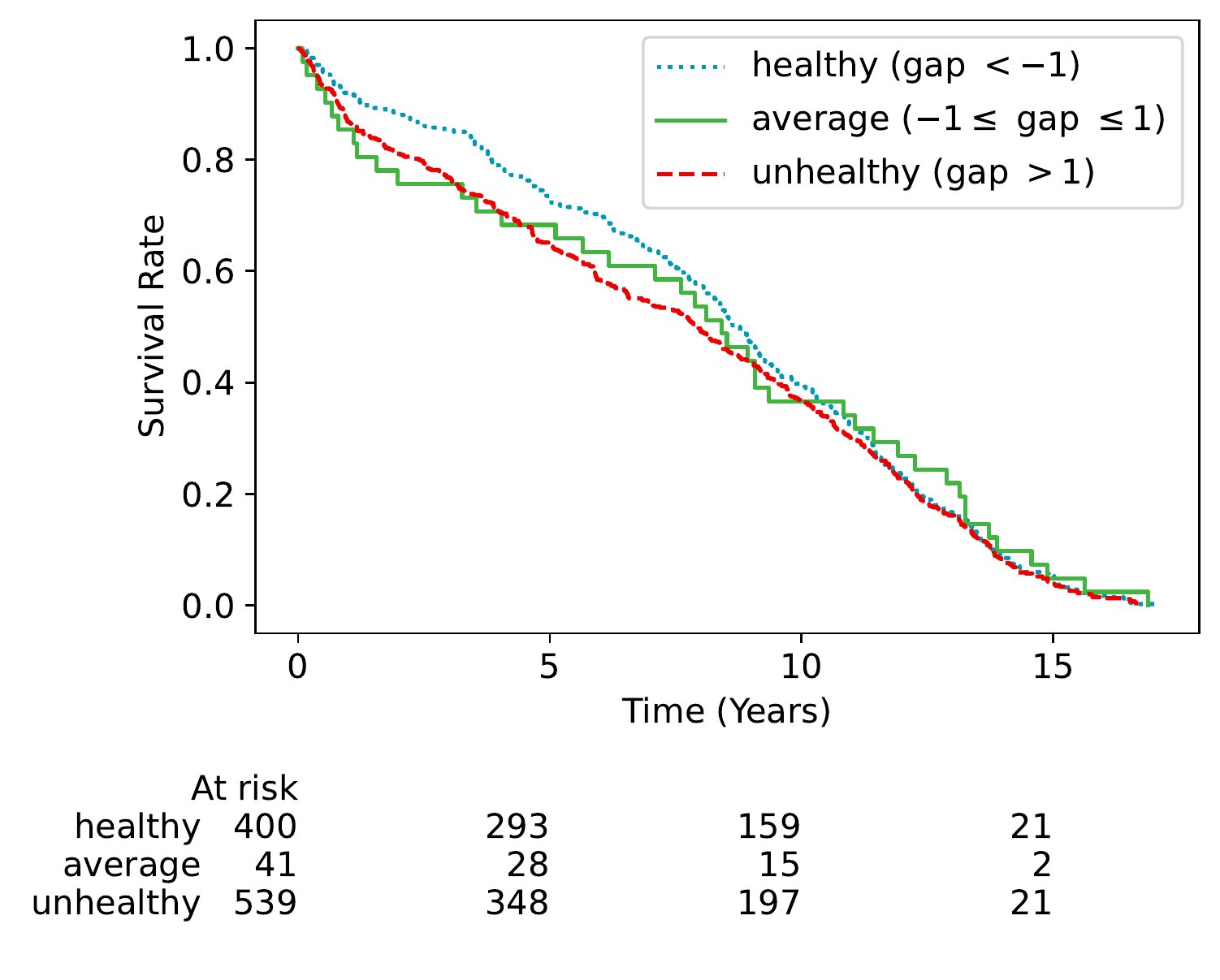}
        \caption{CAC ($p$-value=0.207)}
    \end{subfigure}
    \begin{subfigure}[t]{.45\textwidth}
        \includegraphics[width=\textwidth]{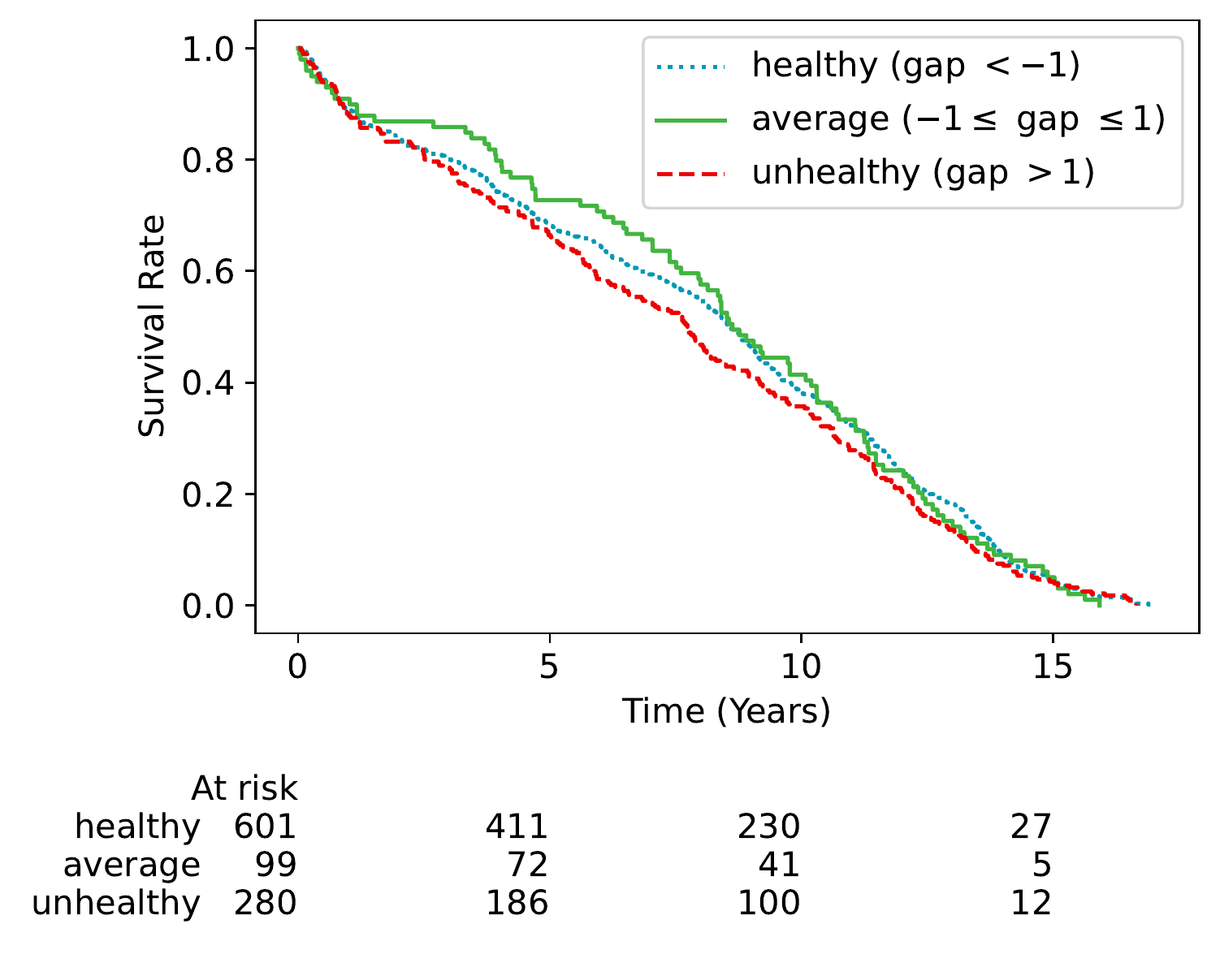}
        \caption{DNN ($p$-value=0.138)}
    \end{subfigure}
    \begin{subfigure}[t]{.45\textwidth}
        \includegraphics[width=\textwidth]{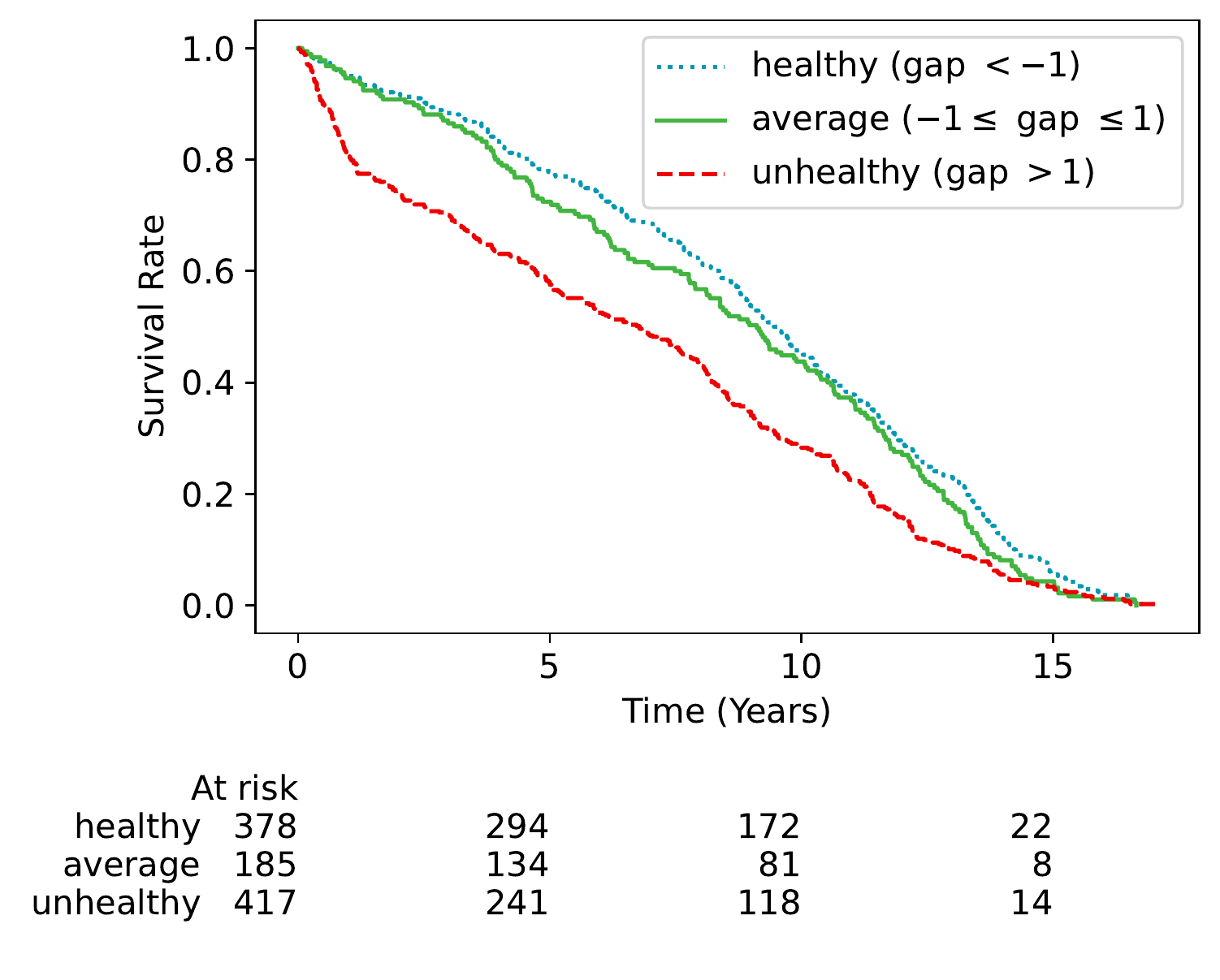}
        \caption{proposed ($p$-value<0.001)}
    \end{subfigure}
    \caption{Results of the survival analysis using the gap between biological and chronological ages for male-\textit{whole}-\lfeature\ case. The $p$-values in parentheses indicate the significance of the discrepancy between the Kaplan-Meier curves of healthy and unhealthy subjects obtained by log-rank tests. }
    \label{fig:kmf}
\end{figure}

To further validate the relationship between the obtained biological ages and mortality, we conduct a survival analysis based on the male-\textit{whole}-\lfeature\ case.
As we adopted three repeated runs with different random seeds for a single experiment setting, the test data with mortality information are extracted from the three runs that results in a total of 956 samples (334,306, and 316 samples, respectively) for the analysis.
Figure \ref{fig:kmf} describes the Kaplan-Meier curves of healthy, average, and unhealthy subjects groups divided by the gap values between biological and chronological ages. 
Specifically, subjects whose gap values are smaller than -1 (larger than 1) are regarded as healthy (unhealthy), while the remaining subjects are included in the average group. 
We also evaluate the difference in Kaplan-Meier curves for healthy and unhealthy subjects using log-rank tests.

The analysis shows that the proposed model efficiently discriminates subjects whose future aspect of mortality differs. 
It is also noteworthy that the survival rate of unhealthy subjects steeply drops in the early stage, which infers that our biological age model is even more powerful for the short-term prediction of mortality. 
In contrast, the biological ages obtained from the baseline models fail to distinguish the subjects with different expectations of mortality. 

\begin{figure}
    \centering
    \begin{subfigure}[t]{.45\textwidth}
        \includegraphics[width=\textwidth]{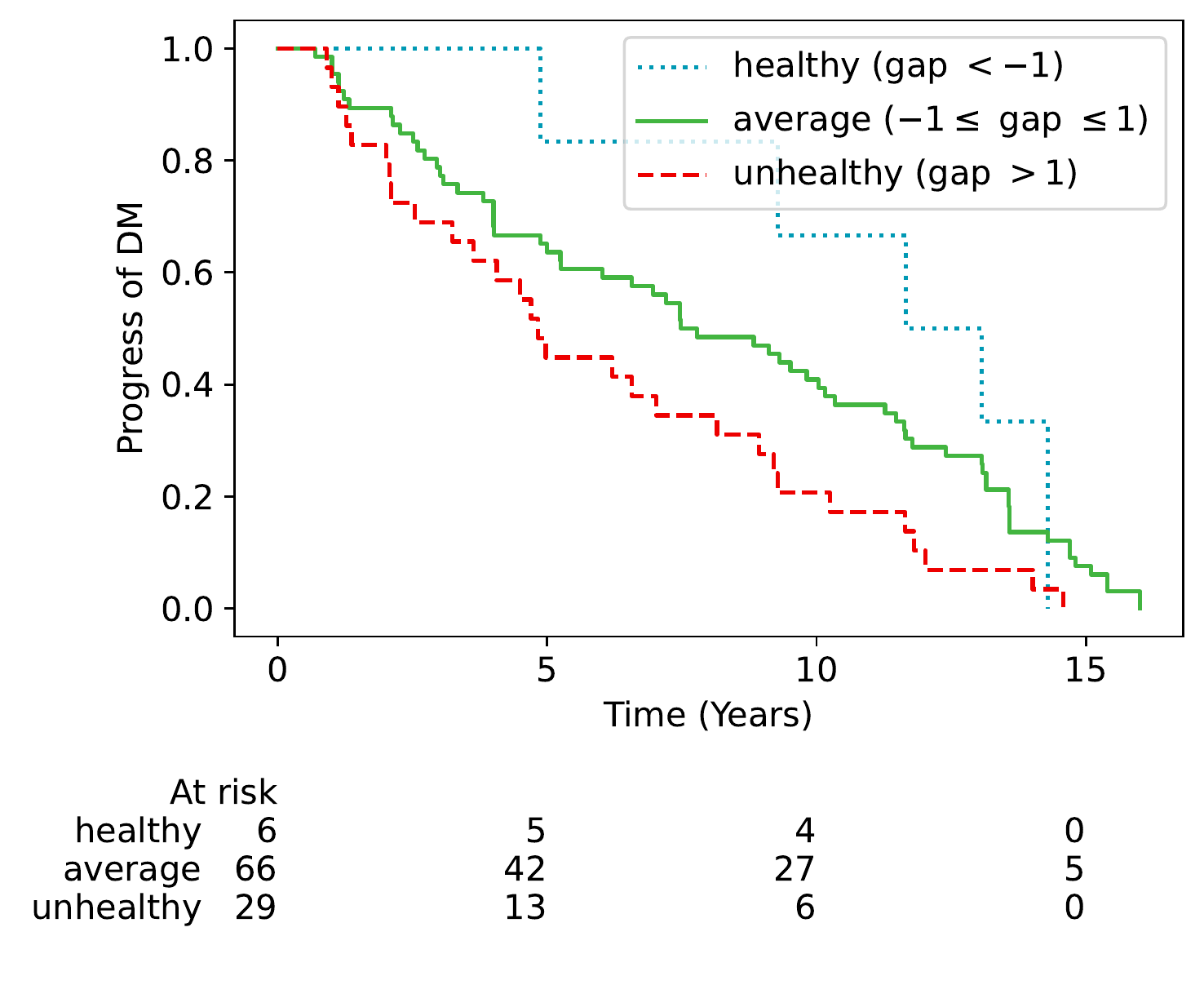}
        \caption{KDM ($p$-value=0.048)}
    \end{subfigure}
    \begin{subfigure}[t]{.45\textwidth}
        \includegraphics[width=\textwidth]{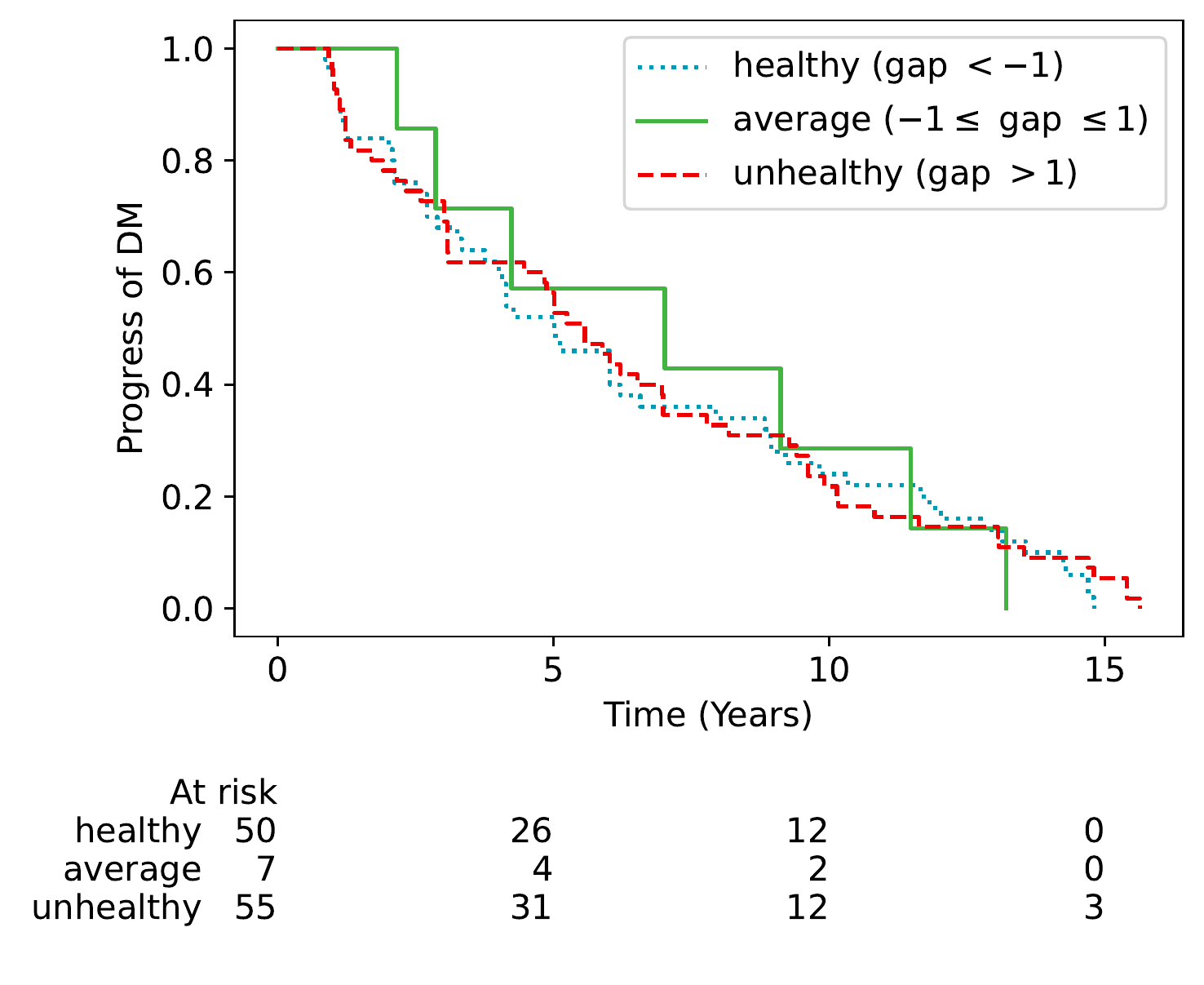}
        \caption{CAC ($p$-value=0.687)}
    \end{subfigure}
    \begin{subfigure}[t]{.45\textwidth}
        \includegraphics[width=\textwidth]{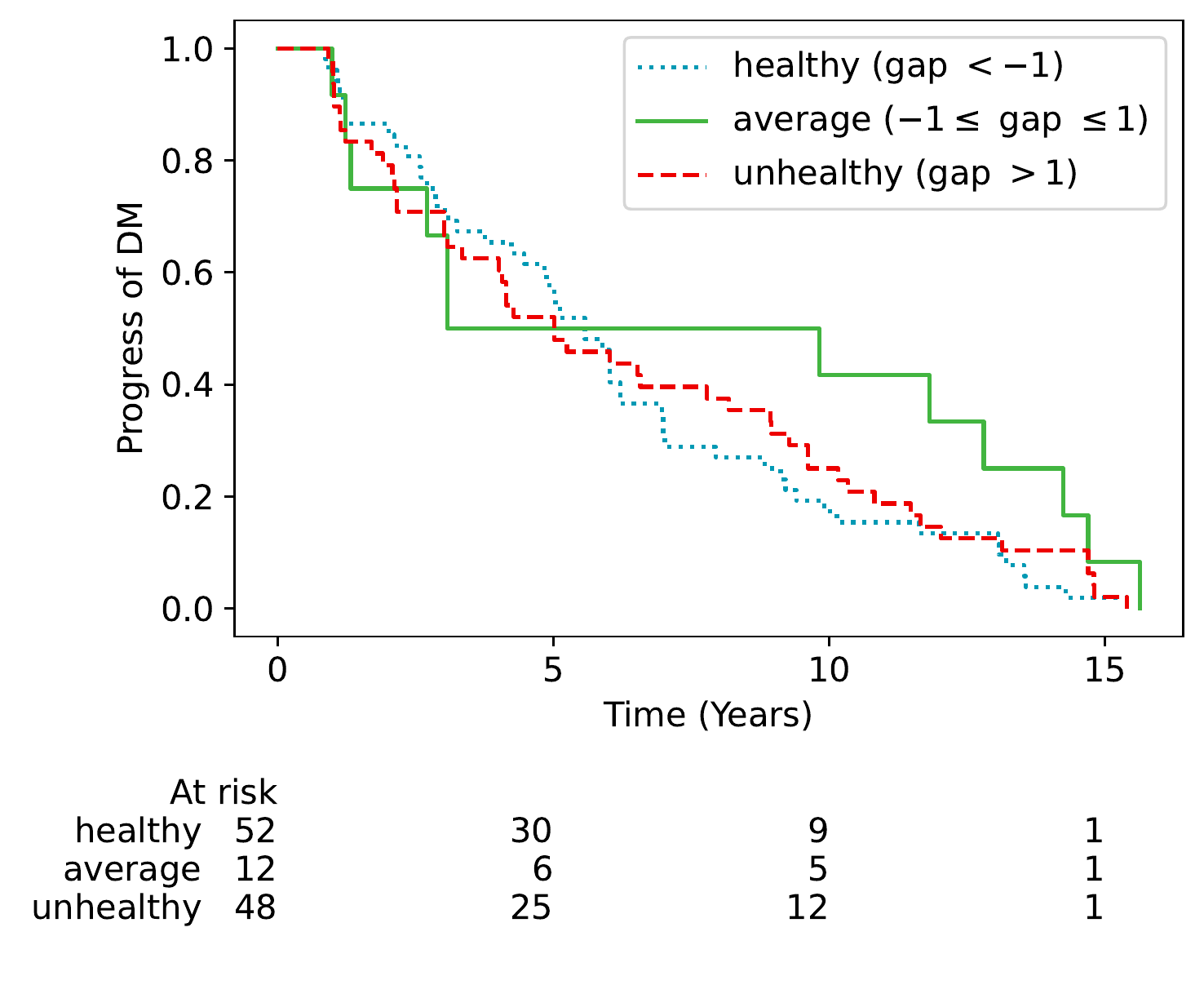}
        \caption{DNN ($p$-value=0.603)}
    \end{subfigure}
    \begin{subfigure}[t]{.45\textwidth}
        \includegraphics[width=\textwidth]{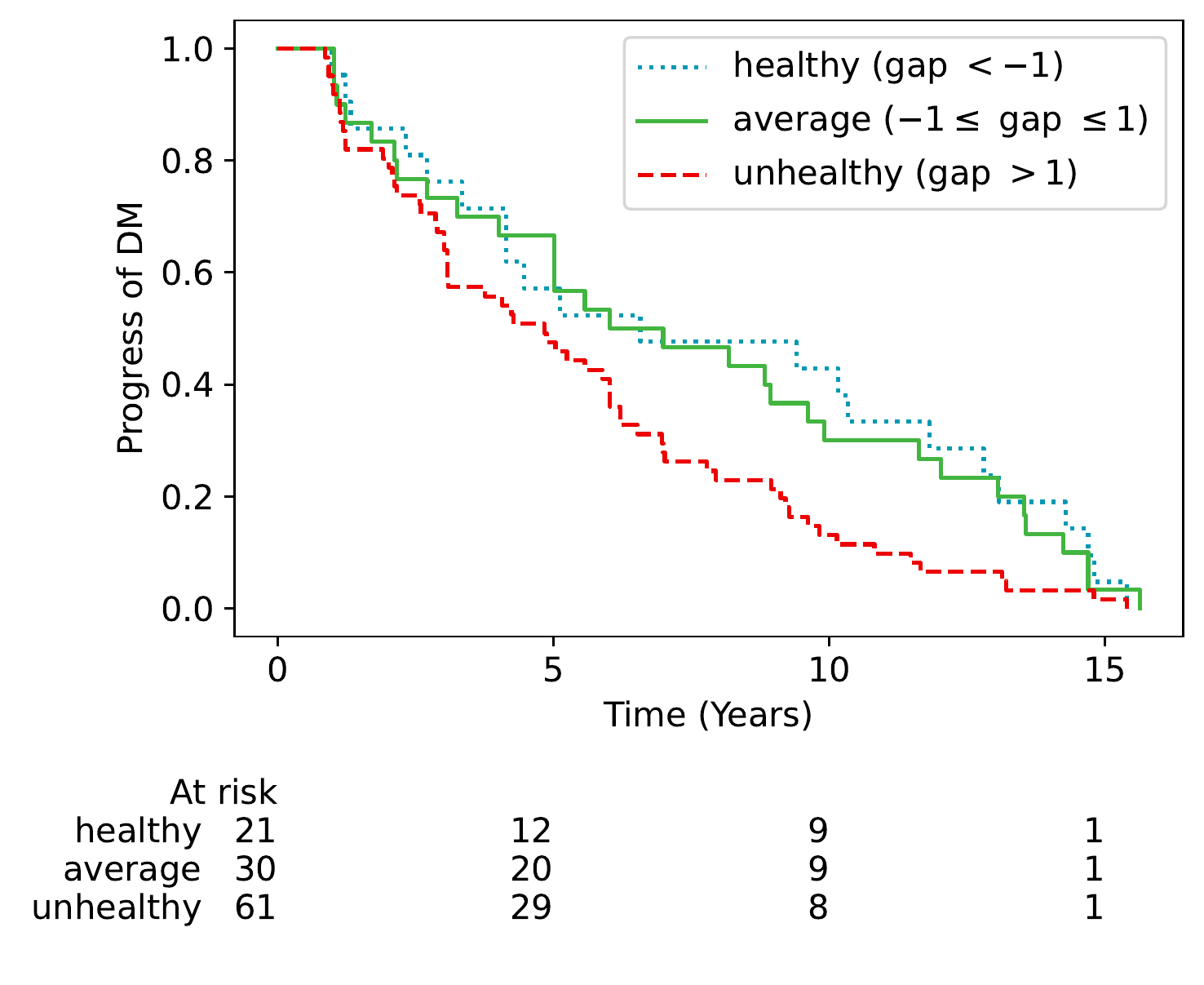}
        \caption{proposed ($p$-value=0.032)} 
    \end{subfigure}
    \caption{Results of the DM onset analysis using the gap between biological and chronological ages for female-\textit{whole}-\lfeature\ case. The $p$-values in parentheses indicate the significance of the discrepancy between the Kaplan-Meier curves of healthy and unhealthy subjects obtained by log-rank tests. }
    \label{fig:kmf-dm}
\end{figure}

In a similar manner, the onset of DM, which is known to be closely related to the biological age \cite{grant2017longitudinal, bahour2022diabetes}, was evaluated depending on the predicted biological ages in Figure \ref{fig:kmf-dm}.
Although healthy and unhealthy groups defined by KDM biological ages also show a statistically significant difference in the pattern of DM onset, biological ages estimated through the proposed model demonstrate their potential to be used as a mediator of DM onset.

%% file: 7_conclusion.tex
In conclusion, the proposed model effectively learns latent representations of biological aging while considering both morbidity and mortality in the subject population. 
We demonstrated that biological ages estimated via the proposed model successfully reflect the morbidity states of several diseases even in the case that the disease was not considered during the training of the model.
The obtained biological ages are also able to accurately predict aging-related outcomes, i.e., mortality and onset of diabetes, and outperform existing methods in the field. 
This research has the potential to improve our understanding of the aging process and help the development of interventions to improve health outcomes in older adults. 
Further experimentation and validation on diverse aging aspects are still needed to enhance the robustness of the model.

%% file: appendix.tex
\renewcommand{\thetable}{\Alph{section}\arabic{table}}
\renewcommand{\thefigure}{\Alph{section}\arabic{figure}}

\section{Average gap values depending on morbidity}
\label{appsec:morbidity_gap}
\setcounter{table}{0}
\setcounter{figure}{0}


\begin{table}[bh]
    \centering
    \caption{Average BA gap values of healthy (Hlty), pre-disease (preD), and with disease (withD) male subjects for morbidity of DM, HBP, and DLP. }
    \small
    \setlength\tabcolsep{5pt}
    \def\arraystretch{0.7}
    \begin{tabular}{c c c r r r r r r r r r}
        \toprule \midrule
        
        \multirow{2}{*}{Population} & \multirow{2}{*}{Model} & \multirow{2}{*}{Disease} & \multicolumn{3}{c}{\lfeature} & \multicolumn{3}{c}{\mfeature} & \multicolumn{3}{c}{\sfeature} \\ \cmidrule(lr){4-12}
        & & & Hlty & preD & withD & Hlty & preD & withD & Hlty & preD & withD \\ \midrule
        
        \multirow{12}{*}{\textit{Whole}} & 
            \multirow{3}{*}{{KDM}} & 
                DM & \multirow{3}{*}{6.14} & 0.64 & 16.0 & \multirow{3}{*}{-0.80} & -3.48 & 20.4 & \multirow{3}{*}{5.59} & -1.47 & 11.43 \\ & & HBP &  & -1.10 & 0.18 &  & -4.16 & 4.82 &  & -4.58 & -1.45 \\ & & DLP &  & -0.96 & -3.78 &  & -5.94 & -0.06 &  & -2.85 & -1.67 \\ \cmidrule(lr){2-12}
                
            & \multirow{3}{*}{{CAC}} & 
                DM & \multirow{3}{*}{-1.62} & -2.62 & -3.87 & \multirow{3}{*}{\underline{-10.74}} & \underline{-4.90} & \underline{1.95} & \multirow{3}{*}{-0.04} & -3.00 & -1.31 \\
                & & HBP &  & -1.72 & -4.63 &  & \underline{-1.03} & \underline{1.41} &  & -3.45 & -3.13 \\
                & & DLP &  & -2.16 & -4.04 &  & \underline{-5.24} & \underline{-0.58} &  & -3.01 & -2.86 \\ \cmidrule(lr){2-12}
                
            & \multirow{3}{*}{{DNN}} & 
                DM & \multirow{3}{*}{0.02} & 0.77 & 0.23 & \multirow{3}{*}{1.19} & 0.88 & -0.18 & \multirow{3}{*}{3.31} & 0.54 & -1.40 \\ & & HBP &  & 0.91 & -0.26 &  & 0.77 & -1.06 &  & 0.70 & -2.49 \\ & & DLP &  & 0.59 & 0.14 &  & 0.48 & -0.26 &  & 0.98 & -0.45 \\ \cmidrule(lr){2-12}
            
            & \multirow{3}{*}{{proposed}} & 
                DM & \multirow{3}{*}{\textbf{\underline{-3.73}}} & \textbf{\underline{-0.47}} & \textbf{\underline{1.26}} & \multirow{3}{*}{\textbf{\underline{-4.21}}} & \textbf{\underline{-0.89}} & \textbf{\underline{1.69}} & \multirow{3}{*}{\textbf{\underline{-2.23}}} & \textbf{\underline{0.39}} & \textbf{\underline{4.35}} \\ 
                & & HBP &  & \textbf{\underline{-0.09}} & \textbf{\underline{0.98}} &  & \textbf{\underline{-0.79}} & \textbf{\underline{0.92}} &  & \textbf{\underline{-0.65}} & \textbf{\underline{0.69}} \\
                & & DLP &  & \textbf{\underline{-0.64}} & \textbf{\underline{1.47}} &  & \underline{-1.41} & \underline{1.01} &  & \textbf{\underline{-0.71}} & \textbf{\underline{0.80}} \\ \midrule

        \multirow{12}{*}{\textit{Average}} & 
            \multirow{3}{*}{{KDM}} & 
                DM & \multirow{3}{*}{2.72} & 0.93 & 11.41 & \multirow{3}{*}{\textbf{\underline{-4.73}}} & \textbf{\underline{-0.15}} & \textbf{\underline{18.75}} & \multirow{3}{*}{\underline{0.40}} & \underline{1.19} & \underline{11.97} \\
                & & HBP &  & 0.46 & 1.36 &  & \textbf{\underline{-0.31}} & \textbf{\underline{5.06}} &  & -1.81 & 0.18 \\
                & & DLP &  & -0.47 & -0.49 &  & \underline{-3.44} & \underline{1.74} &  & -0.93 & 0.03 \\ \cmidrule(lr){2-12}
                
            & \multirow{3}{*}{{CAC}} & 
                DM & \multirow{3}{*}{\underline{-3.44}} & \underline{-2.53} & \underline{-1.78} & \multirow{3}{*}{\underline{-12.85}} & \underline{-3.45} & \underline{3.29} & \multirow{3}{*}{\underline{-6.35}} & \underline{-2.59} & \underline{2.36} \\
                & & HBP &  & -0.12 & -1.41 &  & \underline{1.35} & \underline{3.36} &  & \underline{-3.06} & \underline{-2.40} \\
                & & DLP &  & \underline{-1.88} & \underline{-1.62} &  & \underline{-4.37} & \underline{0.24} &  & \underline{-3.64} & \underline{-1.49} \\ \cmidrule(lr){2-12}
                
            & \multirow{3}{*}{{DNN}} & 
                DM & \multirow{3}{*}{-0.24} & 0.09 & -0.44 & \multirow{3}{*}{1.62} & 1.12 & -0.10 & \multirow{3}{*}{2.68} & 0.48 & -0.62 \\ & & HBP &  & 0.24 & -1.06 &  & 1.03 & -0.74 &  & 0.41 & -2.19 \\ & & DLP &  & 0.05 & -0.58 &  & 1.02 & 0.08 &  & 0.74 & -0.26 \\ \cmidrule(lr){2-12}
            
            & \multirow{3}{*}{{proposed}} & 
                DM & \multirow{3}{*}{\textbf{\underline{-4.28}}} & \textbf{\underline{-0.77}} & \textbf{\underline{0.94}} & \multirow{3}{*}{\underline{-5.32}} & \underline{-1.35} & \underline{0.53} & \multirow{3}{*}{\textbf{\underline{-2.51}}} & \textbf{\underline{0.42}} & \textbf{\underline{3.85}} \\
                & & HBP &  & \textbf{\underline{-0.47}} & \textbf{\underline{1.04}} &  & \underline{-1.27} & \underline{0.64} &  & \textbf{\underline{-0.64}} & \textbf{\underline{1.13}} \\
                & & DLP &  & \underline{-1.07} & \underline{1.80} &  & \underline{-1.79} & \underline{1.35} &  & \textbf{\underline{-0.74}} & \textbf{\underline{0.90}} \\ \midrule

        \multirow{12}{*}{\textit{Normal}} & 
            \multirow{3}{*}{{KDM}} & 
                DM & \multirow{3}{*}{\textbf{\underline{-2.90}}} & \underline{1.77} & \ood{\underline{7.37}} & \multirow{3}{*}{\textbf{\underline{-3.93}}} & \underline{3.68} & \ood{\underline{20.85}} & \multirow{3}{*}{2.37} & 2.18 & \ood{13.91} \\
                & & HBP &  & \textbf{\underline{0.12}} & \ood{\textbf{\underline{0.69}}} &  & \underline{1.15} & \ood{\underline{6.74}} &  & -1.52 & \ood{0.26} \\
                & & DLP &  & \textbf{\underline{0.14}} & \ood{\textbf{\underline{0.86}}} &  & \textbf{\underline{0.01}} & \ood{\textbf{\underline{5.42}}} &  & -0.13 & \ood{-0.29} \\ \cmidrule(lr){2-12}
                
            & \multirow{3}{*}{{CAC}} & 
                DM & \multirow{3}{*}{\textbf{\underline{-3.60}}} & \textbf{\underline{0.76}} & \ood{\textbf{\underline{5.15}}} & \multirow{3}{*}{\textbf{\underline{-9.21}}} & \underline{2.50} & \ood{\underline{10.8}} & \multirow{3}{*}{\underline{-1.94}} & \underline{1.49} & \ood{\underline{9.62}} \\
                & & HBP &  & 1.26 & \ood{0.67} &  & \underline{4.06} & \ood{\underline{6.21}} &  & \underline{-1.24} & \ood{\underline{-0.88}} \\
                & & DLP &  & \textbf{\underline{0.06}} & \ood{\textbf{\underline{0.45}}} &  & \textbf{\underline{0.21}} & \ood{\textbf{\underline{3.67}}} &  & \underline{-0.92} & \ood{\underline{-0.21}} \\ \cmidrule(lr){2-12}
                
            & \multirow{3}{*}{{DNN}} & 
                DM & \multirow{3}{*}{-0.24} & -0.42 & \ood{-2.77} & \multirow{3}{*}{1.57} & 0.19 & \ood{-3.79} & \multirow{3}{*}{3.73} & 1.10 & \ood{-2.30} \\ & & HBP &  & -0.42 & \ood{-2.80} &  & -0.08 & \ood{-3.28} &  & 1.26 & \ood{-2.96} \\ & & DLP &  & -0.20 & \ood{-1.63} &  & 0.11 & \ood{-1.83} &  & 1.63 & \ood{-0.65} \\ \cmidrule(lr){2-12}
            
            & \multirow{3}{*}{{proposed}} & 
                DM & \multirow{3}{*}{\textbf{\underline{-2.13}}} & \textbf{\underline{0.37}} & \ood{\textbf{\underline{1.97}}} & \multirow{3}{*}{\underline{-0.71}} & \underline{2.90} & \ood{\underline{8.59}} & \multirow{3}{*}{\textbf{\underline{-0.67}}} & \textbf{\underline{0.14}} & \ood{\textbf{\underline{0.81}}} \\
                & & HBP &  & \textbf{\underline{0.07}} & \ood{\textbf{\underline{1.51}}} &  &\underline{1.91} & \ood{\underline{5.43}} &  & \textbf{\underline{-0.34}} & \ood{\textbf{\underline{0.44}}} \\
                & & DLP &  & \textbf{\underline{0.02}} & \ood{\textbf{\underline{1.46}}} &  & \underline{1.80} & \ood{\underline{3.59}} &  & \underline{-0.14} & \ood{\underline{-0.10}} \\ \midrule
    \end{tabular}
\end{table}
                
\begin{table}
    \centering
    \caption*{(continued) Average BA gap values of healthy (Hlty), pre-disease (preD), and with disease (withD) male subjects for morbidity of DM, HBP, and DLP. }
    \small
    \setlength\tabcolsep{5pt}
    \def\arraystretch{0.7}
    \begin{tabular}{c c c r r r r r r r r r}
        \midrule
        
        \multirow{2}{*}{Population} & \multirow{2}{*}{Model} & \multirow{2}{*}{Disease} & \multicolumn{3}{c}{\lfeature} & \multicolumn{3}{c}{\mfeature} & \multicolumn{3}{c}{\sfeature} \\ \cmidrule(lr){4-12}
        & & & Hlty & preD & withD & Hlty & preD & withD & Hlty & preD & withD \\ \midrule
        
        \multirow{12}{*}{\makecell{\textit{Super} \\ \textit{normal}}} & 
            \multirow{3}{*}{{KDM}} & 
                DM & \multirow{3}{*}{\textbf{\underline{-0.81}}} & \ood{\textbf{\underline{0.88}}} & \ood{\textbf{\underline{1.87}}} & \multirow{3}{*}{\underline{-0.10}} & \ood{\underline{7.53}} & \ood{\underline{17.67}} & \multirow{3}{*}{0.85} & \ood{0.33} & \ood{12.15} \\
                & & HBP &  & \ood{0.47} & \ood{-0.39} &  & \ood{\underline{6.25}} & \ood{\underline{10.93}} &  & \ood{-3.78} & \ood{-1.93} \\
                & & DLP &  & \ood{\textbf{\underline{0.21}}} & \ood{\textbf{\underline{0.39}}} &  & \ood{\underline{4.47}} & \ood{\underline{9.30}} &  & \ood{-2.72} & \ood{-2.22} \\ \cmidrule(lr){2-12}
                
            & \multirow{3}{*}{{CAC}} & 
                DM & \multirow{3}{*}{-0.96} & \ood{-5.72} & \ood{-13.32} & \multirow{3}{*}{1.64} & \ood{-3.36} & \ood{-11.97} & \multirow{3}{*}{0.80} & \ood{-2.44} & \ood{-10.6} \\ & & HBP &  & \ood{-4.29} & \ood{-10.75} &  & \ood{-1.76} & \ood{-8.25} &  & \ood{-2.46} & \ood{-8.26} \\ & & DLP &  & \ood{-3.76} & \ood{-7.28} &  & \ood{-1.38} & \ood{-4.49} &  & \ood{-1.79} & \ood{-5.11} \\ \cmidrule(lr){2-12}
                
            & \multirow{3}{*}{{DNN}} & 
                DM & \multirow{3}{*}{-0.73} & \ood{-1.73} & \ood{-13.86} & \multirow{3}{*}{0.26} & \ood{-9.60} & \ood{-30.88} & \multirow{3}{*}{-0.44} & \ood{-2.50} & \ood{-5.34} \\ & & HBP &  & \ood{-1.24} & \ood{-8.60} &  & \ood{-8.28} & \ood{-21.58} &  & \ood{-2.87} & \ood{-8.60} \\ & & DLP &  & \ood{-0.14} & \ood{-5.83} &  & \ood{-4.95} & \ood{-15.76} &  & \ood{-2.28} & \ood{-5.20} \\ \cmidrule(lr){2-12}
            
            & \multirow{3}{*}{{proposed}} & 
                DM & \multirow{3}{*}{0.54} & \ood{0.04} & \ood{1.51} & \multirow{3}{*}{\underline{1.87}} & \ood{\underline{4.56}} & \ood{\underline{7.90}} & \multirow{3}{*}{\underline{0.59}} & \ood{\underline{3.16}} & \ood{\underline{6.59}} \\
                & & HBP &  & \ood{-0.42} & \ood{-0.19} &  & \ood{\underline{3.63}} & \ood{\underline{6.80}} &  & \ood{\underline{2.40}} & \ood{\underline{4.99}} \\
                & & DLP &  & \ood{-0.06} & \ood{0.24} &  & \ood{\underline{3.47}} & \ood{\underline{5.55}} &  & \ood{\underline{2.23}} & \ood{\underline{4.13}} \\ 
                
    \midrule \bottomrule
    \end{tabular}
    \label{tab:whole-morbidity1-m}
\end{table}

\begin{table}[hb]
    \centering
    \caption{Average BA gap values of healthy (Hlty), pre-disease (preD), and with disease (withD) female subjects for morbidity of DM, HBP, and DLP.}
    \small
    \def\arraystretch{0.7}
    \setlength\tabcolsep{5pt}
    \begin{tabular}{c c c r r r r r r r r r}
        \toprule \midrule
        
        \multirow{2}{*}{Population} & \multirow{2}{*}{Model} & \multirow{2}{*}{Disease} & \multicolumn{3}{c}{\lfeature} & \multicolumn{3}{c}{\mfeature} & \multicolumn{3}{c}{\sfeature} \\ \cmidrule(lr){4-12}
        & & & Hlty & preD & withD & Hlty & preD & withD & Hlty & preD & withD \\ \midrule
        
        \multirow{12}{*}{\textit{Whole}} & 
            \multirow{3}{*}{{KDM}} & 
                DM & \multirow{3}{*}{\textbf{\underline{-1.80}}} & \textbf{\underline{-0.25}} & \textbf{\underline{2.92}} & \multirow{3}{*}{\underline{-22.68}} & \underline{-3.71} & \underline{45.89} & \multirow{3}{*}{\underline{-22.25}} & \underline{1.24} & \underline{61.12} \\
                & & HBP &  & \textbf{\underline{0.20}} & \textbf{\underline{1.60}} &  & \underline{4.22} & \underline{24.52} &  & \underline{-4.96} & \underline{21.13} \\
                & & DLP &  & \textbf{\underline{-0.60}} & \textbf{\underline{1.30}} &  & \underline{-7.78} & \underline{17.29} &  & \underline{-8.63} & \underline{17.29} \\ \cmidrule(lr){2-12}
                
            & \multirow{3}{*}{{CAC}} & 
                DM & \multirow{3}{*}{-4.15} & -5.46 & -3.42 & \multirow{3}{*}{\underline{-7.78}} & \underline{-4.41} & \underline{4.29} & \multirow{3}{*}{-3.14} & -3.38 & 1.04 \\
                & & HBP &  & -4.97 & -4.21 &  & \underline{2.73} & \underline{4.12} &  & -4.69 & -2.07 \\
                & & DLP &  & -4.90 & -5.78 &  & \underline{-4.64} & \underline{-0.03} &  & -3.62 & -3.04 \\ \cmidrule(lr){2-12}
                
            & \multirow{3}{*}{{DNN}} & 
                DM & \multirow{3}{*}{0.31} & 0.43 & -0.60 & \multirow{3}{*}{0.52} & 1.16 & -0.06 & \multirow{3}{*}{1.76} & 0.78 & -1.12 \\ & & HBP &  & 0.45 & -1.33 &  & 0.86 & -1.37 &  & -1.36 & -3.69 \\ & & DLP &  & 0.39 & -0.77 &  & 0.53 & -0.59 &  & 0.21 & -2.32 \\ \cmidrule(lr){2-12}
            
            & \multirow{3}{*}{{proposed}} & 
                DM & \multirow{3}{*}{\textbf{\underline{-3.29}}} & \textbf{\underline{-0.20}} & \textbf{\underline{3.39}} & \multirow{3}{*}{\textbf{\underline{-3.11}}} & \textbf{\underline{-0.05}} & \textbf{\underline{2.98}} & \multirow{3}{*}{\textbf{\underline{-0.59}}} & \textbf{\underline{0.83}} & \textbf{\underline{3.29}} \\
                & & HBP &  & \underline{1.03} & \underline{2.80} &  & \textbf{\underline{0.80}} & \textbf{\underline{2.29}} &  &\textbf{\underline{-0.72}} & \textbf{\underline{0.27}} \\
                & & DLP &  & \textbf{\underline{-0.54}} & \textbf{\underline{2.49}} &  & \textbf{\underline{-0.73}} & \textbf{\underline{2.17}} &  & \textbf{\underline{-0.35}} & \textbf{\underline{0.13}} \\ \midrule

        \multirow{12}{*}{\textit{Average}} & 
            \multirow{3}{*}{{KDM}} & 
                DM & \multirow{3}{*}{\textbf{\underline{-2.05}}} & \textbf{\underline{0.34}} & \textbf{\underline{3.80}} & \multirow{3}{*}{\underline{-9.43}} & \underline{1.60} & \underline{22.29} & \multirow{3}{*}{\textbf{\underline{-7.62}}} & \underline{3.30} & \underline{23.78} \\
                & & HBP &  & \underline{1.01} & \underline{2.26} &  & \underline{6.22} & \underline{11.39} &  & \textbf{\underline{-0.60}} & \textbf{\underline{8.83}} \\
                & & DLP &  & \textbf{\underline{-0.28}} & \textbf{\underline{1.93}} &  & \underline{-1.26} & \underline{8.50} &  & \underline{-1.20} & \underline{7.55} \\ \cmidrule(lr){2-12}
                
            & \multirow{3}{*}{{CAC}} & 
                DM & \multirow{3}{*}{\underline{-6.03}} & \underline{-2.93} & \underline{3.36} & \multirow{3}{*}{\underline{-7.98}} & \underline{-1.87} & \underline{5.49} & \multirow{3}{*}{\textbf{\underline{-4.67}}} & \textbf{\underline{-0.50}} & \textbf{\underline{5.09}} \\ & & HBP &  & \underline{1.97} & \underline{3.63} &  & 5.58 & 5.09 &  & \underline{-2.54} & \underline{0.55} \\ & & DLP &  & \underline{-3.18} & \underline{1.46} &  & \underline{-2.22} & \underline{2.11} &  & \underline{-2.44} & \underline{0.88} \\ \cmidrule(lr){2-12}
                
            & \multirow{3}{*}{{DNN}} & 
                DM & \multirow{3}{*}{0.85} & 1.11 & 0.09 & \multirow{3}{*}{0.08} & 0.37 & -0.41 & \multirow{3}{*}{3.22} & 1.54 & -0.50 \\ & & HBP &  & 1.44 & -0.33 &  & 0.51 & -1.97 &  & -0.20 & -3.03 \\ & & DLP &  & 1.11 & 0.04 &  & -0.07 & -1.54 &  & 1.65 & -1.75 \\ \cmidrule(lr){2-12}
            
            & \multirow{3}{*}{{proposed}} & 
                DM & \multirow{3}{*}{\textbf{\underline{-3.47}}} & \textbf{\underline{0.21}} & \textbf{\underline{3.08}} & \multirow{3}{*}{\textbf{\underline{-4.04}}} & \textbf{\underline{0.69}} & \textbf{\underline{3.60}} & \multirow{3}{*}{\textbf{\underline{-1.70}}} & \underline{1.88} & \underline{10.64} \\ & & HBP &  & \underline{1.06} & \underline{2.37} &  & \underline{1.49} & \underline{2.99} &  & \textbf{\underline{0.19}} & \textbf{\underline{3.65}} \\ & & DLP &  & \textbf{\underline{-\underline{0.19}}} & \textbf{\underline{2.71}} &  & \textbf{\underline{0.12}} & \textbf{\underline{3.37}} &  & \textbf{\underline{-0.42}} & \textbf{\underline{2.83}} \\ \midrule
    \end{tabular}
    \label{tab:whole-morbidity1-f}
\end{table}

\begin{table}[ht]
    \centering
    \caption*{(continued) Average BA gap values of healthy (Hlty), pre-disease (preD), and with disease (withD) female subjects for morbidity of DM, HBP, and DLP.}
    \small
    \def\arraystretch{0.7}
    \setlength\tabcolsep{5pt}
    \begin{tabular}{c c c r r r r r r r r r}
        \midrule
        
        \multirow{2}{*}{Population} & \multirow{2}{*}{Model} & \multirow{2}{*}{Disease} & \multicolumn{3}{c}{\lfeature} & \multicolumn{3}{c}{\mfeature} & \multicolumn{3}{c}{\sfeature} \\ \cmidrule(lr){4-12}
        & & & Hlty & preD & withD & Hlty & preD & withD & Hlty & preD & withD \\ \midrule

        \multirow{12}{*}{\textit{Normal}} & 
            \multirow{3}{*}{{KDM}} & 
                DM & \multirow{3}{*}{\underline{-8.60}} & \underline{5.10} & \ood{1\underline{9.34}} & \multirow{3}{*}{\underline{-12.29}} & \underline{7.11} & \ood{\underline{34.95}} & \multirow{3}{*}{\underline{-9.70}} & \underline{9.34} & \ood{\underline{40.16}} \\ & & HBP &  & \underline{7.42} & \ood{\underline{11.95}} &  & \underline{11.28} & \ood{\underline{20.06}} &  & \underline{4.43} & \ood{\underline{17.09}} \\ & & DLP &  & \underline{1.80} & \ood{\underline{12.32}} &  & \underline{1.80} & \ood{\underline{17.19}} &  & \underline{1.46} & \ood{\underline{14.09}} \\ \cmidrule(lr){2-12}
                
            & \multirow{3}{*}{{CAC}} & 
                DM & \multirow{3}{*}{\textbf{\underline{-6.01}}} & \textbf{\underline{0.38}} & \ood{\textbf{\underline{5.69}}} & \multirow{3}{*}{\textbf{\underline{-9.80}}} & \underline{1.00} & \ood{\underline{8.06}} & \multirow{3}{*}{\textbf{\underline{-4.47}}} & \textbf{\underline{0.91}} & \ood{\textbf{\underline{5.72}}} \\ & & HBP &  & 5.97 & \ood{5.71} &  & 10.76 & \ood{7.93} &  & \underline{-1.90} & \ood{\underline{0.40}} \\ & & DLP &  & \textbf{\underline{-0.98}} & \ood{\textbf{\underline{4.36}}} &  & \textbf{\underline{0.07}} & \ood{\textbf{\underline{5.62}}} &  & \underline{-1.92} & \ood{\underline{0.30}} \\ \cmidrule(lr){2-12}
                
            & \multirow{3}{*}{{DNN}} & 
                DM & \multirow{3}{*}{1.44} & 1.02 & \ood{-3.31} & \multirow{3}{*}{-0.11} & 1.11 & \ood{-3.19} & \multirow{3}{*}{1.31} & 0.73 & \ood{-4.39} \\ & & HBP &  & 1.04 & \ood{-2.98} &  & 0.78 & \ood{-4.03} &  & -1.43 & \ood{-5.69} \\ & & DLP &  & 1.15 & \ood{-1.76} &  & 0.66 & \ood{-2.32} &  & 0.19 & \ood{-4.68} \\ \cmidrule(lr){2-12}
            
            & \multirow{3}{*}{{proposed}} & 
                DM & \multirow{3}{*}{\textbf{\underline{-1.56}}} & \underline{1.34} & \ood{\underline{4.39}} & \multirow{3}{*}{0.31} & 0.30 & \ood{-0.06} & \multirow{3}{*}{\textbf{\underline{-0.66}}} & \underline{1.10} & \ood{\underline{1.17}} \\ & & HBP &  & \underline{1.55} & \ood{\underline{3.82}} &  & -0.47 & \ood{0.39} &  & \textbf{\underline{0.39}} & \ood{\textbf{\underline{1.51}}} \\ & & DLP &  & \textbf{\underline{0.58}} & \ood{\textbf{\underline{3.53}}} &  & 0.24 & \ood{0.38} &  & \textbf{\underline{0.26}} & \ood{\textbf{\underline{1.01}}} \\ \midrule

        \multirow{12}{*}{\makecell{\textit{Super} \\ \textit{normal}}} & 
            \multirow{3}{*}{{KDM}} & 
                DM & \multirow{3}{*}{\underline{-0.16}} & \ood{\underline{9.34}} & \ood{\underline{14.63}} & \multirow{3}{*}{\underline{-0.79}} & \ood{\underline{19.71}} & \ood{\underline{43.96}} & \multirow{3}{*}{\underline{-0.24}} & \ood{\underline{13.74}} & \ood{\underline{34.35}} \\ & & HBP &  & \ood{\underline{8.39}} & \ood{\underline{10.80}} &  & \ood{\underline{19.60}} & \ood{\underline{31.30}} &  & \ood{\underline{11.25}} & \ood{\underline{21.83}} \\ & & DLP &  & \ood{\underline{6.59}} & \ood{\underline{12.22}} &  & \ood{\underline{13.13}} & \ood{\underline{29.38}} &  & \ood{\underline{8.07}} & \ood{\underline{19.15}} \\ \cmidrule(lr){2-12}
                
            & \multirow{3}{*}{{CAC}} & 
                DM & \multirow{3}{*}{-0.28} & \ood{1.52} & \ood{-5.15} & \multirow{3}{*}{1.10} & \ood{6.41} & \ood{-3.47} & \multirow{3}{*}{1.17} & \ood{3.53} & \ood{-4.27} \\ & & HBP &  & \ood{2.00} & \ood{-3.56} &  & \ood{6.03} & \ood{-1.33} &  & \ood{-0.24} & \ood{-4.22} \\ & & DLP &  & \ood{1.50} & \ood{-1.88} &  & \ood{5.58} & \ood{0.80} &  & \ood{1.63} & \ood{-2.54} \\ \cmidrule(lr){2-12}
                
            & \multirow{3}{*}{{DNN}} & 
                DM & \multirow{3}{*}{-1.82} & \ood{-5.07} & \ood{-16.44} & \multirow{3}{*}{-0.33} & \ood{-5.00} & \ood{-20.13} & \multirow{3}{*}{-0.10} & \ood{-7.19} & \ood{-21.52} \\ & & HBP &  & \ood{-5.61} & \ood{-13.27} &  & \ood{-6.17} & \ood{-16.07} &  & \ood{-7.33} & \ood{-15.93} \\ & & DLP &  & \ood{-3.89} & \ood{-12.06} &  & \ood{-3.00} & \ood{-13.99} &  & \ood{-4.08} & \ood{-13.72} \\ \cmidrule(lr){2-12}
            
            & \multirow{3}{*}{{proposed}} & 
                DM & \multirow{3}{*}{\underline{0.05}} & \ood{\underline{1.39}} & \ood{\underline{1.71}} & \multirow{3}{*}{\underline{2.08}} & \ood{\underline{7.28}} & \ood{\underline{10.09}} & \multirow{3}{*}{\textbf{\underline{-0.22}}} & \ood{0.38} & \ood{0.22} \\ & & HBP &  & \ood{\underline{1.13}} & \ood{\underline{1.58}} &  & \ood{\underline{6.16}} & \ood{\underline{8.26}} &  & \ood{\textbf{\underline{0.27}}} & \ood{\textbf{\underline{1.11}}} \\ & & DLP &  & \ood{\underline{1.28}} & \ood{\underline{2.21}} &  & \ood{\underline{6.25}} & \ood{\underline{8.79}} &  & \ood{\textbf{\underline{0.02}}} & \ood{\textbf{\underline{0.73}}} \\ 
    \midrule \bottomrule
    \end{tabular}
\end{table}

\begin{table}[hb]
    \centering
    \caption{Average BA gap values of healthy (Hlty) and with disease (withD) male subjects for morbidity of MS, Cancer, CVD, and CVA. }
    \label{tab:male-binary}
    \small
    \setlength\tabcolsep{5pt}
    \def\arraystretch{0.7}
    \begin{tabular}{c c c r r r r r r}
        \toprule \midrule
        \multirow{2}{*}{Population} & \multirow{2}{*}{Model} & \multirow{2}{*}{Disease} & \multicolumn{2}{c}{\lfeature} & \multicolumn{2}{c}{\mfeature} & \multicolumn{2}{c}{\sfeature} \\ \cmidrule(lr){4-9}
        & & & Hlty & withD & Hlty & withD & Hlty & withD \\ \midrule
        
        \multirow{16}{*}{\textit{Whole}} &
            \multirow{4}{*}{{KDM}} & 
                MS & \multirow{4}{*}{\underline{6.14}} & -5.05 & \multirow{4}{*}{\textbf{\underline{-0.80}}} & \textbf{\underline{4.58}} & \multirow{4}{*}{5.59} & -2.09 \\ & & Cancer &  & 2.23 &  & \textbf{\underline{0.46}} &  & 0.48 \\ & & CVD &  & \underline{7.50} &  & \textbf{\underline{2.12}} &  & 3.68 \\ & & CVA &  & 0.29 &  & \textbf{\underline{5.90}} &  & -0.55 \\ \cmidrule(lr){2-9}
                
            & \multirow{4}{*}{{CAC}} & 
                MS & \multirow{4}{*}{-1.62} & -4.07 & \multirow{4}{*}{\textbf{\underline{-10.74}}} & \textbf{\underline{0.84}} & \multirow{4}{*}{-0.04} & -1.99 \\ & & Cancer &  & -7.38 &  & \underline{-2.81} &  & -3.94 \\ & & CVD &  & -4.06 &  & \underline{-1.53} &  & -2.01 \\ & & CVA &  & -6.62 &  & \underline{-0.12} &  & -5.30 \\ \cmidrule(lr){2-9}
                
            & \multirow{4}{*}{{DNN}} & 
                MS & \multirow{4}{*}{\underline{0.02}} & \underline{0.33} & \multirow{4}{*}{1.19} & 0.05 & \multirow{4}{*}{3.31} & -0.56 \\ & & Cancer &  & -1.87 &  & -2.78 &  & -3.34 \\ & & CVD &  & -0.70 &  & -2.27 &  & -2.80 \\ & & CVA &  & -2.32 &  & -3.79 &  & -5.11 \\ \cmidrule(lr){2-9}
            
            & \multirow{4}{*}{{proposed}} & 
                MS & \multirow{4}{*}{\textbf{\underline{-3.73}}} & \textbf{\underline{2.07}} & \multirow{4}{*}{\textbf{\underline{-4.21}}} & \textbf{\underline{1.91}} & \multirow{4}{*}{\textbf{\underline{-2.23}}} & \textbf{\underline{1.91}} \\ & & Cancer &  & \textbf{\underline{0.38}} &  & \textbf{\underline{0.13}} &  & \textbf{\underline{0.56}} \\ & & CVD &  & \underline{-0.32} &  & \underline{-0.06} &  & \textbf{\underline{1.19}} \\ & & CVA &  & \textbf{\underline{0.03}} &  & \textbf{\underline{0.23}} &  & \textbf{\underline{1.62}} \\ \midrule
        
    \end{tabular}    
\end{table}

\begin{table}
    \centering
    \caption*{(continued) Average BA gap values of healthy (Hlty) and with disease (withD) male subjects for morbidity of MS, Cancer, CVD, and CVA.}
    \small
    \setlength\tabcolsep{5pt}
    \def\arraystretch{0.7}
    \begin{tabular}{c c c r r r r r r}
        \midrule
        \multirow{2}{*}{Population} & \multirow{2}{*}{Model} & \multirow{2}{*}{Disease} & \multicolumn{2}{c}{\lfeature} & \multicolumn{2}{c}{\mfeature} & \multicolumn{2}{c}{\sfeature} \\ \cmidrule(lr){4-9}
        & & & Hlty & withD & Hlty & withD & Hlty & withD \\ \midrule
        
        \multirow{8}{*}{\textit{Average}} & 
            \multirow{4}{*}{{KDM}} & 
                MS & \multirow{4}{*}{\underline{2.72}} & 1.09 & \multirow{4}{*}{\textbf{\underline{-4.73}}} & \textbf{\underline{6.85}} & \multirow{4}{*}{\underline{0.40}} & \underline{0.73} \\ & & Cancer &  & 1.66 &  & \textbf{\underline{1.70}} &  & \underline{1.44} \\ & & CVD &  & \underline{3.29} &  & \textbf{\underline{3.53}} &  & \underline{1.79} \\ & & CVA &  & \underline{5.36} &  & \textbf{\underline{2.78}} &  & \underline{5.45} \\ \cmidrule(lr){2-9}
                
            & \multirow{4}{*}{{CAC}} & 
                MS & \multirow{4}{*}{\underline{-3.44}} & \underline{-1.53} & \multirow{4}{*}{\textbf{\underline{-12.85}}} & \textbf{\underline{2.05}} & \multirow{4}{*}{\underline{-6.35}} & \underline{-0.02} \\ & & Cancer &  & \underline{-2.97} &  & \underline{-3.21} &  & \underline{-3.23} \\ & & CVD &  & \underline{-1.65} &  & \textbf{\underline{0.43}} &  & \underline{-3.01} \\ & & CVA &  & \underline{-1.24} &  & \underline{-1.40} &  & \underline{-1.77} \\ \cmidrule(lr){2-9}
                
            & \multirow{4}{*}{{DNN}} & 
                MS & \multirow{4}{*}{\underline{-0.24}} & \underline{-0.12} & \multirow{4}{*}{1.62} & 0.46 & \multirow{4}{*}{2.68} & 0.17 \\ & & Cancer &  & -2.30 &  & -2.08 &  & -2.19 \\ & & CVD &  & -2.12 &  & -1.95 &  & -2.62 \\ & & CVA &  & -2.77 &  & -3.91 &  & -1.23 \\ \cmidrule(lr){2-9}
            
            & \multirow{4}{*}{{proposed}} & 
                MS & \multirow{4}{*}{\textbf{\underline{-4.28}}} & \textbf{\underline{2.71}} & \multirow{4}{*}{\textbf{\underline{-5.32}}} & \textbf{\underline{2.52}} & \multirow{4}{*}{\textbf{\underline{-2.51}}} & \textbf{\underline{2.02}} \\ & & Cancer &  & \textbf{\underline{0.14}} &  & \underline{-0.63} &  & \textbf{\underline{1.01}} \\ & & CVD &  & \underline{-0.64} &  & \underline{-1.30} &  & \textbf{\underline{1.58}} \\ & & CVA &  & \underline{-0.11} &  & \underline{-0.81} &  & \textbf{\underline{2.04}} \\ \midrule
            
        \multirow{16}{*}{\textit{Normal}} & 
            \multirow{4}{*}{{KDM}} & 
                MS & \multirow{4}{*}{\textbf{\underline{-2.90}}} & \ood{\textbf{\underline{0.96}}} & \multirow{4}{*}{\textbf{\underline{-3.93}}} & \ood{\textbf{\underline{8.92}}} & \multirow{4}{*}{\underline{2.37}} & \ood{0.05} \\ & & Cancer &  & \ood{\textbf{\underline{1.58}}} &  & \ood{\textbf{\underline{5.46}}} &  & \ood{\underline{2.96}} \\ & & CVD &  & \ood{\textbf{\underline{0.84}}} &  & \ood{\textbf{\underline{5.33}}} &  & \ood{\underline{4.40}} \\ & & CVA &  & \ood{\textbf{\underline{1.46}}} &  & \ood{\textbf{\underline{5.50}}} &  & \ood{\underline{4.34}} \\ \cmidrule(lr){2-9}
                
            & \multirow{4}{*}{{CAC}} & 
                MS & \multirow{4}{*}{\textbf{\underline{-3.60}}} & \ood{\textbf{\underline{0.77}}} & \multirow{4}{*}{\textbf{\underline{-9.21}}} & \ood{\textbf{\underline{4.82}}} & \multirow{4}{*}{\textbf{\underline{-1.94}}} & \ood{\textbf{\underline{1.09}}} \\ & & Cancer &  & \ood{\underline{-\underline{0.86}}} &  & \ood{\textbf{\underline{2.31}}} &  & \ood{\underline{-0.65}} \\ & & CVD &  & \ood{\underline{-0.38}} &  & \ood{\textbf{\underline{2.57}}} &  & \ood{\textbf{\underline{0.86}}} \\ & & CVA &  & \ood{\underline{-0.50}} &  & \ood{\textbf{\underline{2.47}}} &  & \ood{\underline{-0.31}} \\ \cmidrule(lr){2-9}
                
            & \multirow{4}{*}{{DNN}} & 
                MS & \multirow{4}{*}{-0.24} & \ood{-1.61} & \multirow{4}{*}{1.57} & \ood{-1.50} & \multirow{4}{*}{3.73} & \ood{-0.43} \\ & & Cancer &  & \ood{-3.50} &  & \ood{-4.90} &  & \ood{-3.28} \\ & & CVD &  & \ood{-3.66} &  & \ood{-4.34} &  & \ood{-3.48} \\ & & CVA &  & \ood{-3.87} &  & \ood{-5.13} &  & \ood{-3.51} \\ \cmidrule(lr){2-9}
            
            & \multirow{4}{*}{{proposed}} & 
                MS & \multirow{4}{*}{\textbf{\underline{-2.13}}} & \ood{\textbf{\underline{1.95}}} & \multirow{4}{*}{\textbf{\underline{-0.71}}} & \ood{\textbf{\underline{3.75}}} & \multirow{4}{*}{\textbf{\underline{-0.67}}} & \ood{\underline{-0.21}} \\ & & Cancer &  & \ood{\textbf{\underline{1.69}}} &  & \ood{\textbf{\underline{5.78}}} &  & \ood{\textbf{\underline{0.43}}} \\ & & CVD &  & \ood{\textbf{\underline{1.89}}} &  & \ood{\textbf{\underline{7.56}}} &  & \ood{\textbf{\underline{1.51}}} \\ & & CVA &  & \ood{\textbf{\underline{2.10}}} &  & \ood{\textbf{\underline{9.10}}} &  & \ood{\textbf{\underline{2.69}}} \\ \midrule
                
        \multirow{16}{*}{\makecell{\textit{Super} \\ \textit{normal}}} & 
            \multirow{4}{*}{{KDM}} & 
                MS & \multirow{4}{*}{\textbf{\underline{-0.81}}} & \ood{\textbf{\underline{0.61}}} & \multirow{4}{*}{\textbf{\underline{-0.10}}} & \ood{\textbf{\underline{12.45}}} & \multirow{4}{*}{\underline{0.85}} & \ood{-1.22} \\ & & Cancer &  & \ood{\underline{-0.81}} &  & \ood{\textbf{\underline{8.38}}} &  & \ood{\underline{0.95}} \\ & & CVD &  & \ood{-2.26} &  & \ood{\textbf{\underline{7.32}}} &  & \ood{\underline{1.98}} \\ & & CVA &  & \ood{-1.85} &  & \ood{\textbf{\underline{7.49}}} &  & \ood{\underline{2.00}} \\ \cmidrule(lr){2-9}
                
            & \multirow{4}{*}{{CAC}} & 
                MS & \multirow{4}{*}{-0.96} & \ood{-7.71} & \multirow{4}{*}{1.64} & \ood{-4.31} & \multirow{4}{*}{0.80} & \ood{-5.37} \\
                & & Cancer &  & \ood{-12.02} &  & \ood{-9.67} &  & \ood{-8.86} \\
                & & CVD &  & \ood{-13.03} &  & \ood{-10.72} &  & \ood{-9.77} \\
                & & CVA &  & \ood{-13.74} &  & \ood{-11.59} &  & \ood{-10.85} \\ \cmidrule(lr){2-9}
                
            & \multirow{4}{*}{{DNN}} & 
                MS & \multirow{4}{*}{-0.73} & \ood{-7.94} & \multirow{4}{*}{0.26} & \ood{-21.64} & \multirow{4}{*}{-0.44} & \ood{-4.71} \\
                & & Cancer &  & \ood{-9.30} &  & \ood{-19.70} &  & \ood{-9.13} \\
                & & CVD &  & \ood{-11.19} &  & \ood{-23.04} &  & \ood{-10.35} \\
                & & CVA &  & \ood{-11.67} &  & \ood{-24.05} &  & \ood{-10.72} \\ \cmidrule(lr){2-9}
            
            & \multirow{4}{*}{{proposed}} & 
                MS & \multirow{4}{*}{\underline{0.54}} & \ood{0.41} & \multirow{4}{*}{\underline{1.87}} & \ood{\underline{5.56}} & \multirow{4}{*}{\underline{0.59}} & \ood{\underline{4.65}} \\ & & Cancer &  & \ood{0.28} &  & \ood{\underline{7.31}} &  & \ood{\underline{5.29}} \\ & & CVD &  & \ood{\underline{0.65}} &  & \ood{\underline{7.57}} &  & \ood{\underline{5.71}} \\ & & CVA &  & \ood{\underline{0.67}} &  & \ood{\underline{8.05}} &  & \ood{\underline{6.01}} \\
            \midrule \bottomrule
    \end{tabular}
\end{table}

\begin{table}
    \centering
    \caption{Average BA gap values of healthy (Hlty) and with disease (withD) female subjects for morbidity of MS, CA, CVD, and CVA.}
    \small
    \setlength\tabcolsep{5pt}
    \def\arraystretch{0.7}
    \begin{tabular}{c c c r r r r r r}
        \toprule \midrule
        \multirow{2}{*}{Population} & \multirow{2}{*}{Model} & \multirow{2}{*}{Disease} & \multicolumn{2}{c}{\lfeature} & \multicolumn{2}{c}{\mfeature} & \multicolumn{2}{c}{\sfeature} \\ \cmidrule(lr){4-9}
        & & & Hlty & withD & Hlty & withD & Hlty & withD \\ \midrule
        
        \multirow{16}{*}{\textit{Whole}} &
            \multirow{4}{*}{{KDM}} & 
                MS & \multirow{4}{*}{\textbf{\underline{-1.80}}} & \textbf{\underline{3.01}} & \multirow{4}{*}{\textbf{\underline{-22.68}}} & \textbf{\underline{42.92}} & \multirow{4}{*}{\textbf{\underline{-22.25}}} & \textbf{\underline{49.36}} \\ & & Cancer &  & \underline{-0.25} &  & \underline{-2.40} &  & \underline{-4.92} \\ & & CVD &  & \textbf{\underline{1.11}} &  & \textbf{\underline{13.36}} &  & \textbf{\underline{20.07}} \\ & & CVA &  & \textbf{\underline{0.55}} &  & \textbf{\underline{16.04}} &  & \textbf{\underline{22.21}} \\ \cmidrule(lr){2-9}
                
            & \multirow{4}{*}{{CAC}} & 
                MS & \multirow{4}{*}{\underline{-4.15}} & \underline{-1.67} & \multirow{4}{*}{\textbf{\underline{-7.78}}} & \textbf{\underline{3.97}} & \multirow{4}{*}{\textbf{\underline{-3.14}}} & \textbf{\underline{0.45}} \\ & & Cancer &  & -8.33 &  & \underline{-2.89} &  & -4.30 \\ & & CVD &  & -4.90 &  & \textbf{\underline{0.01}} &  & \underline{-1.16} \\ & & CVA &  & -5.74 &  & \underline{-1.30} &  & \underline{-2.67} \\ \cmidrule(lr){2-9}
                
            & \multirow{4}{*}{{DNN}} & 
                MS & \multirow{4}{*}{0.31} & -0.64 & \multirow{4}{*}{0.52} & 0.11 & \multirow{4}{*}{1.76} & -1.08 \\ & & Cancer &  & -1.32 &  & -2.01 &  & -3.56 \\ & & CVD &  & -1.45 &  & -2.20 &  & -3.14 \\ & & CVA &  & -1.63 &  & -2.22 &  & -2.39 \\ \cmidrule(lr){2-9}
            
            & \multirow{4}{*}{{proposed}} & 
                MS & \multirow{4}{*}{\textbf{\underline{-3.29}}} & \textbf{\underline{3.66}} & \multirow{4}{*}{\textbf{\underline{-3.11}}} & \textbf{\underline{2.96}} & \multirow{4}{*}{\textbf{\underline{-0.59}}} & \textbf{\underline{0.97}} \\ & & Cancer &  & \textbf{\underline{0.84}} &  & \textbf{\underline{0.84}} &  & \underline{-0.08} \\ & & CVD &  & \textbf{\underline{1.73}} &  & \textbf{\underline{0.90}} &  & \textbf{\underline{0.57}} \\ & & CVA &  & \textbf{\underline{1.32}} &  & \textbf{\underline{1.21}} &  & \textbf{\underline{0.92}} \\ \midrule

        \multirow{16}{*}{\textit{Average}} & 
            \multirow{4}{*}{{KDM}} & 
                MS & \multirow{4}{*}{\textbf{\underline{-2.05}}} & \textbf{\underline{4.74}} & \multirow{4}{*}{\textbf{\underline{-9.43}}} & \textbf{\underline{24.81}} & \multirow{4}{*}{\textbf{\underline{-7.62}}} & \textbf{\underline{23.37}} \\ & & Cancer &  & \underline{-0.07} &  & \textbf{\underline{0.24}} &  & \textbf{\underline{0.89}} \\ & & CVD &  & \textbf{\underline{1.21}} &  & \textbf{\underline{2.98}} &  & \textbf{\underline{4.74}} \\ & & CVA &  & \textbf{\underline{1.52}} &  & \textbf{\underline{4.46}} &  & \textbf{\underline{1.92}} \\ \cmidrule(lr){2-9}
                
            & \multirow{4}{*}{{CAC}} & 
                MS & \multirow{4}{*}{\textbf{\underline{-6.03}}} & \textbf{\underline{7.81}} & \multirow{4}{*}{\textbf{\underline{-7.98}}} & \textbf{\underline{8.61}} & \multirow{4}{*}{\textbf{\underline{-4.67}}} & \textbf{\underline{7.19}} \\ & & Cancer &  & \underline{-3.38} &  & \underline{-1.58} &  & \underline{-2.52} \\ & & CVD &  & \underline{-0.82} &  & \textbf{\underline{0.50}} &  & \underline{-1.17} \\ & & CVA &  & \textbf{\underline{0.25}} &  & \textbf{\underline{1.62}} &  & \underline{-1.14} \\ \cmidrule(lr){2-9}
                
            & \multirow{4}{*}{{DNN}} & 
                MS & \multirow{4}{*}{0.85} & 0.42 & \multirow{4}{*}{0.08} & -0.11 & \multirow{4}{*}{3.22} & -0.63 \\ & & Cancer &  & 0.13 &  & -2.06 &  & -1.46 \\ & & CVD &  & -0.92 &  & -2.50 &  & -2.11 \\ & & CVA &  & -0.46 &  & -1.42 &  & -3.42 \\ \cmidrule(lr){2-9}
            
            & \multirow{4}{*}{{proposed}} & 
                MS & \multirow{4}{*}{\textbf{\underline{-3.47}}} & \textbf{\underline{3.93}} & \multirow{4}{*}{\textbf{\underline{-4.04}}} & \textbf{\underline{4.30}} & \multirow{4}{*}{\textbf{\underline{-1.70}}} & \textbf{\underline{5.84}} \\ & & Cancer &  & \textbf{\underline{0.70}} &  & \textbf{\underline{1.22}} &  & \textbf{\underline{0.93}} \\ & & CVD &  & \textbf{\underline{1.59}} &  & \textbf{\underline{1.89}} &  & \textbf{\underline{1.97}} \\ & & CVA &  & \textbf{\underline{0.41}} &  & \textbf{\underline{1.47}} &  & \textbf{\underline{2.03}} \\ \midrule

        \multirow{16}{*}{\textit{Normal}} & 
            \multirow{4}{*}{{KDM}} & 
                MS & \multirow{4}{*}{\textbf{\underline{-8.60}}} & \ood{\textbf{\underline{20.48}}} & \multirow{4}{*}{\textbf{\underline{-12.29}}} & \ood{\textbf{\underline{33.79}}} & \multirow{4}{*}{\textbf{\underline{-9.70}}} & \ood{\textbf{\underline{32.16}}} \\ & & Cancer &  & \ood{\textbf{\underline{4.54}}} &  & \ood{\textbf{\underline{6.14}}} &  & \ood{\textbf{\underline{4.93}}} \\ & & CVD &  & \ood{\textbf{\underline{8.53}}} &  & \ood{\textbf{\underline{12.87}}} &  & \ood{\textbf{\underline{15.46}}} \\ & & CVA &  & \ood{\textbf{\underline{5.16}}} &  & \ood{\textbf{\underline{8.59}}} &  & \ood{\textbf{\underline{8.42}}} \\ \cmidrule(lr){2-9}
                
            & \multirow{4}{*}{{CAC}} & 
                MS & \multirow{4}{*}{\textbf{\underline{-6.01}}} & \ood{\textbf{\underline{8.71}}} & \multirow{4}{*}{\textbf{\underline{-9.80}}} & \ood{\textbf{\underline{10.07}}} & \multirow{4}{*}{\textbf{\underline{-4.47}}} & \ood{\textbf{\underline{4.26}}} \\ & & Cancer &  & \ood{\underline{-0.45}} &  & \ood{\textbf{\underline{1.03}}} &  & \ood{\underline{-2.24}} \\ & & CVD &  & \ood{\textbf{\underline{1.67}}} &  & \ood{\textbf{\underline{2.46}}} &  & \ood{\underline{-0.40}} \\ & & CVA &  & \ood{\textbf{\underline{0.38}}} &  & \ood{\textbf{\underline{1.92}}} &  & \ood{-\underline{1.92}} \\ \cmidrule(lr){2-9}
                
            & \multirow{4}{*}{{DNN}} & 
                MS & \multirow{4}{*}{1.44} & \ood{-2.65} & \multirow{4}{*}{-0.11} & \ood{-2.68} & \multirow{4}{*}{1.31} & \ood{-4.11} \\ & & Cancer &  & \ood{-1.12} &  & \ood{-3.15} &  & \ood{-3.99} \\ & & CVD &  & \ood{-2.74} &  & \ood{-4.71} &  & \ood{-5.23} \\ & & CVA &  & \ood{-1.62} &  & \ood{-4.05} &  & \ood{-5.04} \\ \cmidrule(lr){2-9}
            
            & \multirow{4}{*}{{proposed}} & 
                MS & \multirow{4}{*}{\textbf{\underline{-1.56}}} & \ood{\textbf{\underline{3.75}}} & \multirow{4}{*}{\underline{0.31}} & \ood{-0.88} & \multirow{4}{*}{\textbf{\underline{-0.66}}} & \ood{\textbf{\underline{0.94}}} \\ & & Cancer &  & \ood{\textbf{\underline{2.27}}} &  & \ood{0.29} &  & \ood{\textbf{\underline{0.44}}} \\ & & CVD &  & \ood{\textbf{\underline{3.60}}} &  & \ood{\underline{0.96}} &  & \ood{\textbf{\underline{1.63}}} \\ & & CVA &  & \ood{\textbf{\underline{2.89}}} &  & \ood{\underline{0.62}} &  & \ood{\textbf{\underline{1.12}}} \\ \midrule
    \end{tabular}
\end{table}

\begin{table}[ht]
    \centering
    \caption*{(continued) Average BA gap values of healthy (Hlty) and with disease (withD) female subjects for morbidity of MS, CA, CVD, and CVA.}
    \small
    \setlength\tabcolsep{5pt}
    \def\arraystretch{0.7}
    \begin{tabular}{c c c r r r r r r}
        \midrule
        \multirow{2}{*}{Population} & \multirow{2}{*}{Model} & \multirow{2}{*}{Disease} & \multicolumn{2}{c}{\lfeature} & \multicolumn{2}{c}{\mfeature} & \multicolumn{2}{c}{\sfeature} \\ \cmidrule(lr){4-9}
        & & & Hlty & withD & Hlty & withD & Hlty & withD \\ \midrule

        \multirow{16}{*}{\makecell{\textit{Super} \\ \textit{normal}}} & 
            \multirow{4}{*}{{KDM}} & 
                MS & \multirow{4}{*}{\textbf{\underline{-0.16}}} & \ood{\textbf{\underline{17.08}}} & \multirow{4}{*}{\textbf{\underline{-0.79}}} & \ood{\textbf{\underline{43.75}}} & \multirow{4}{*}{\textbf{\underline{-0.24}}} & \ood{\textbf{\underline{33.54}}} \\  & & Cancer &  & \ood{\textbf{\underline{8.78}}} &  & \ood{\textbf{\underline{19.70}}} &  & \ood{\textbf{\underline{13.13}}} \\ & & CVD &  & \ood{\textbf{\underline{9.79}}} &  & \ood{\textbf{\underline{27.27}}} &  & \ood{\textbf{\underline{20.66}}} \\ & & CVA &  & \ood{\textbf{\underline{7.47}}} &  & \ood{\textbf{\underline{22.57}}} &  & \ood{\textbf{\underline{16.38}}} \\ \cmidrule(lr){2-9}
                
            & \multirow{4}{*}{{CAC}} & 
                MS & \multirow{4}{*}{-0.28} & \ood{-2.38} & \multirow{4}{*}{\underline{1.10}} & \ood{0.33} & \multirow{4}{*}{1.17} & \ood{-1.91} \\ & & Cancer &  & \ood{-1.22} &  & \ood{\underline{1.74}} &  & \ood{-1.37} \\ & & CVD &  & \ood{-4.10} &  & \ood{-2.08} &  & \ood{-4.38} \\ & & CVA &  & \ood{-3.61} &  & \ood{-1.11} &  & \ood{-3.66} \\ \cmidrule(lr){2-9}
                
            & \multirow{4}{*}{{DNN}} & 
                MS & \multirow{4}{*}{-1.82} & \ood{-13.00} & \multirow{4}{*}{-0.33} & \ood{-16.53} & \multirow{4}{*}{-0.10} & \ood{-16.35} \\ & & Cancer &  & \ood{-8.66} &  & \ood{-9.99} &  & \ood{-10.63} \\ & & CVD &  & \ood{-13.17} &  & \ood{-16.15} &  & \ood{-15.91} \\ & & CVA &  & \ood{-11.46} &  & \ood{-13.06} &  & \ood{-14.08} \\ \cmidrule(lr){2-9}
            
            & \multirow{4}{*}{{proposed}} & 
                MS & \multirow{4}{*}{\underline{0.05}} & \ood{\underline{1.72}} & \multirow{4}{*}{\underline{2.08}} & \ood{\underline{9.29}} & \multirow{4}{*}{\underline{-0.22}} & \ood{\underline{0.85}} \\ & & Cancer &  & \ood{\underline{1.51}} &  & \ood{\underline{6.90}} &  & \ood{\underline{0.46}} \\ & & CVD &  & \ood{\underline{1.75}} &  & \ood{\underline{8.16}} &  & \ood{\underline{1.24}} \\ & & CVA &  & \ood{\underline{1.15}} &  & \ood{\underline{7.63}} &  & \ood{\underline{1.05}} \\
    \midrule \bottomrule
    \end{tabular}
\end{table}

\clearpage
\section{Average gap values depending on chronological ages and morbidity}
\label{appsec:morbidity_age_gap}

\setcounter{table}{0}
\setcounter{figure}{0}

\begin{figure}[hb]
    \centering
    \begin{subfigure}[h]{.45\textwidth}
        \includegraphics[width=\textwidth]{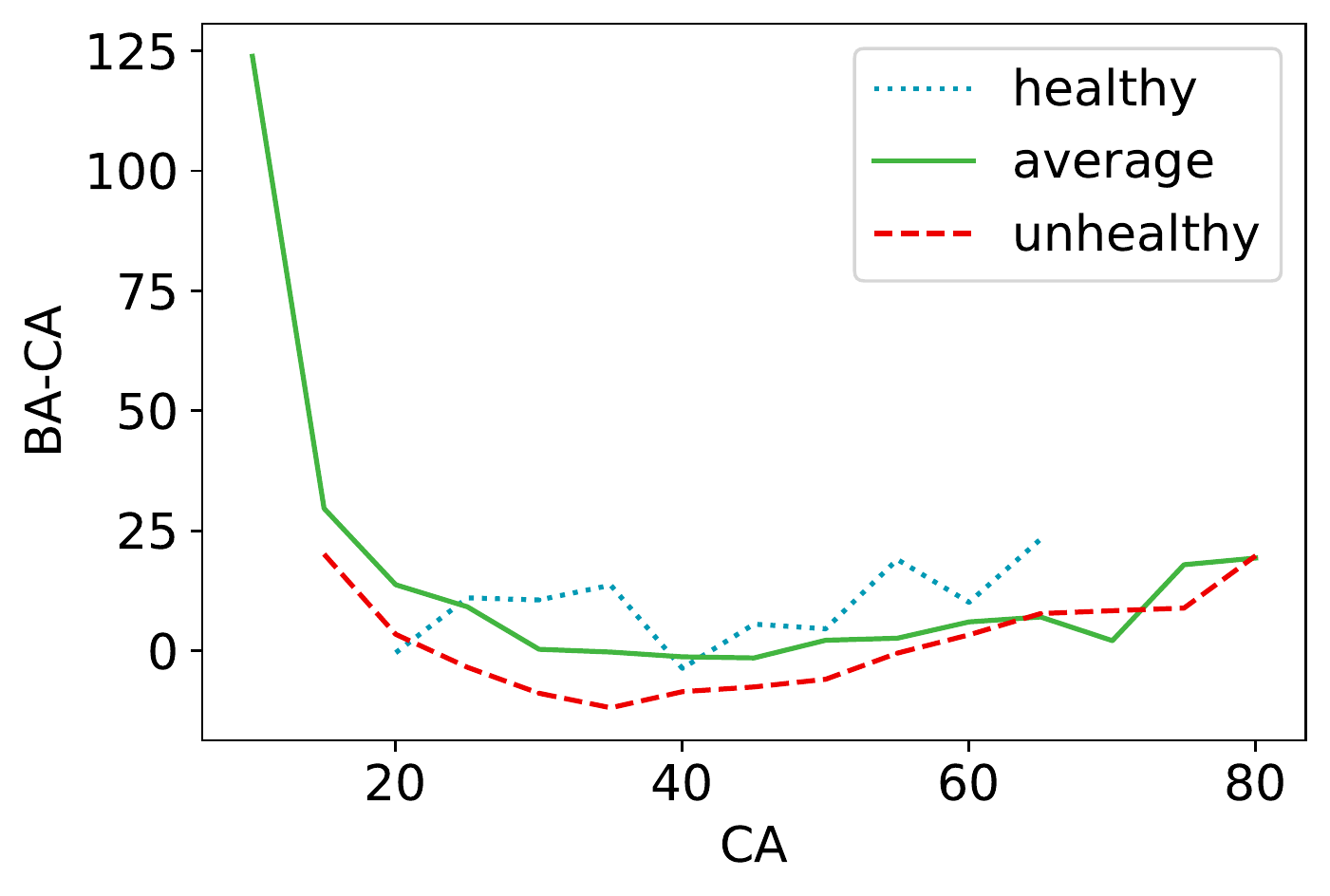}
        \caption{KDM}
    \end{subfigure}
    \begin{subfigure}[h]{.45\textwidth}
        \includegraphics[width=\textwidth]{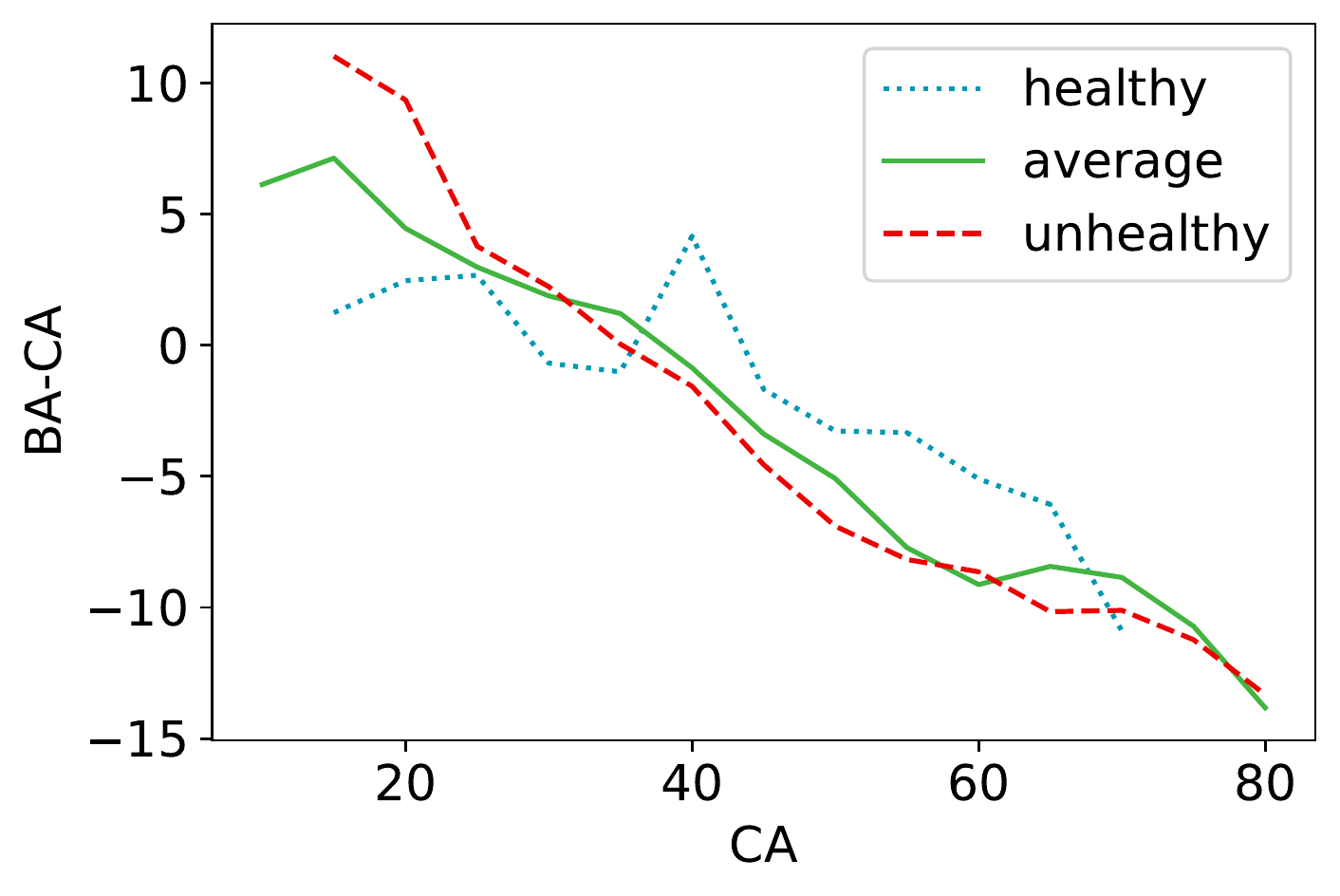}
        \caption{CAC}
    \end{subfigure}
    \begin{subfigure}[h]{.45\textwidth}
        \includegraphics[width=\textwidth]{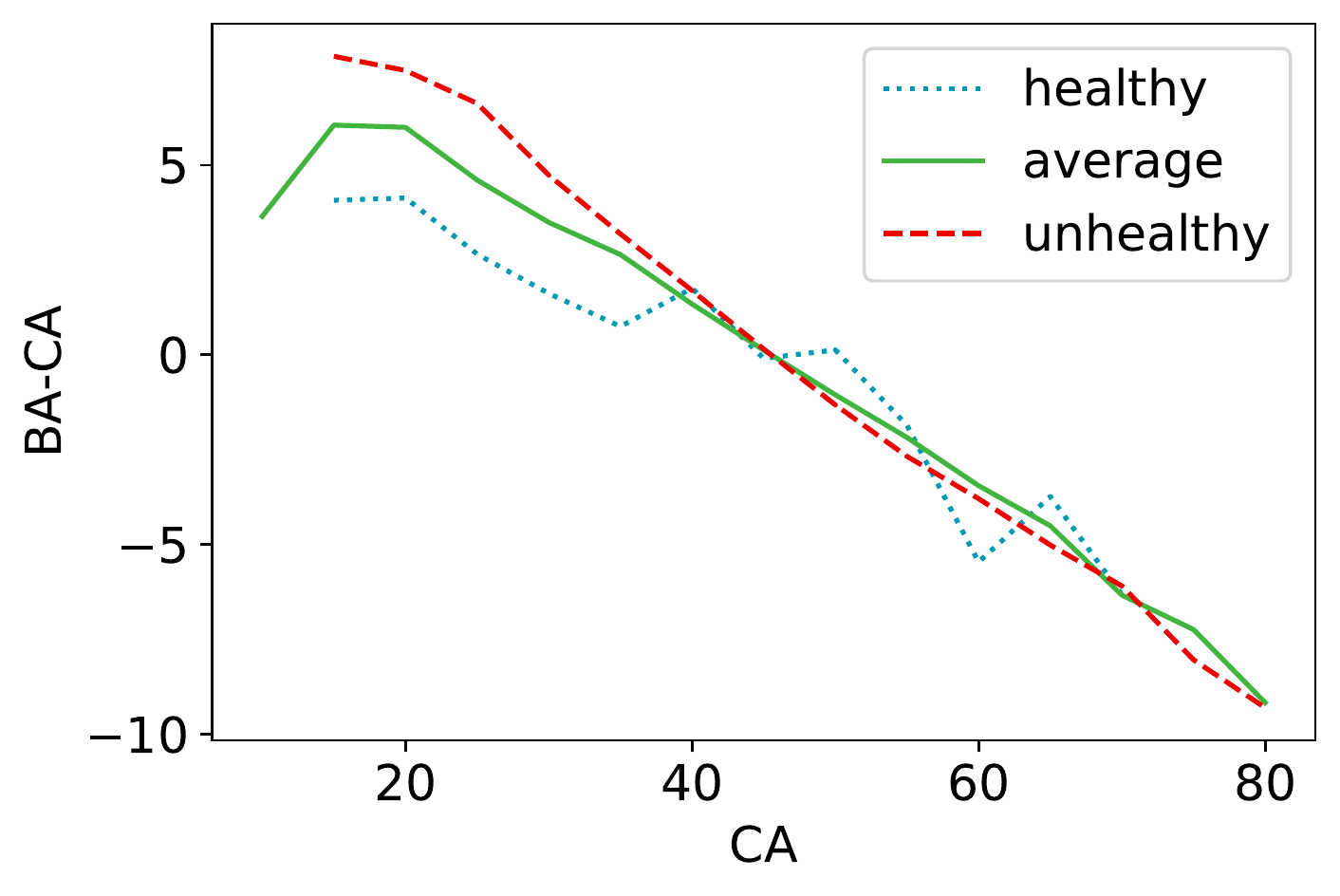}
        \caption{DNN}
    \end{subfigure}
    \begin{subfigure}[h]{.45\textwidth}
        \includegraphics[width=\textwidth]{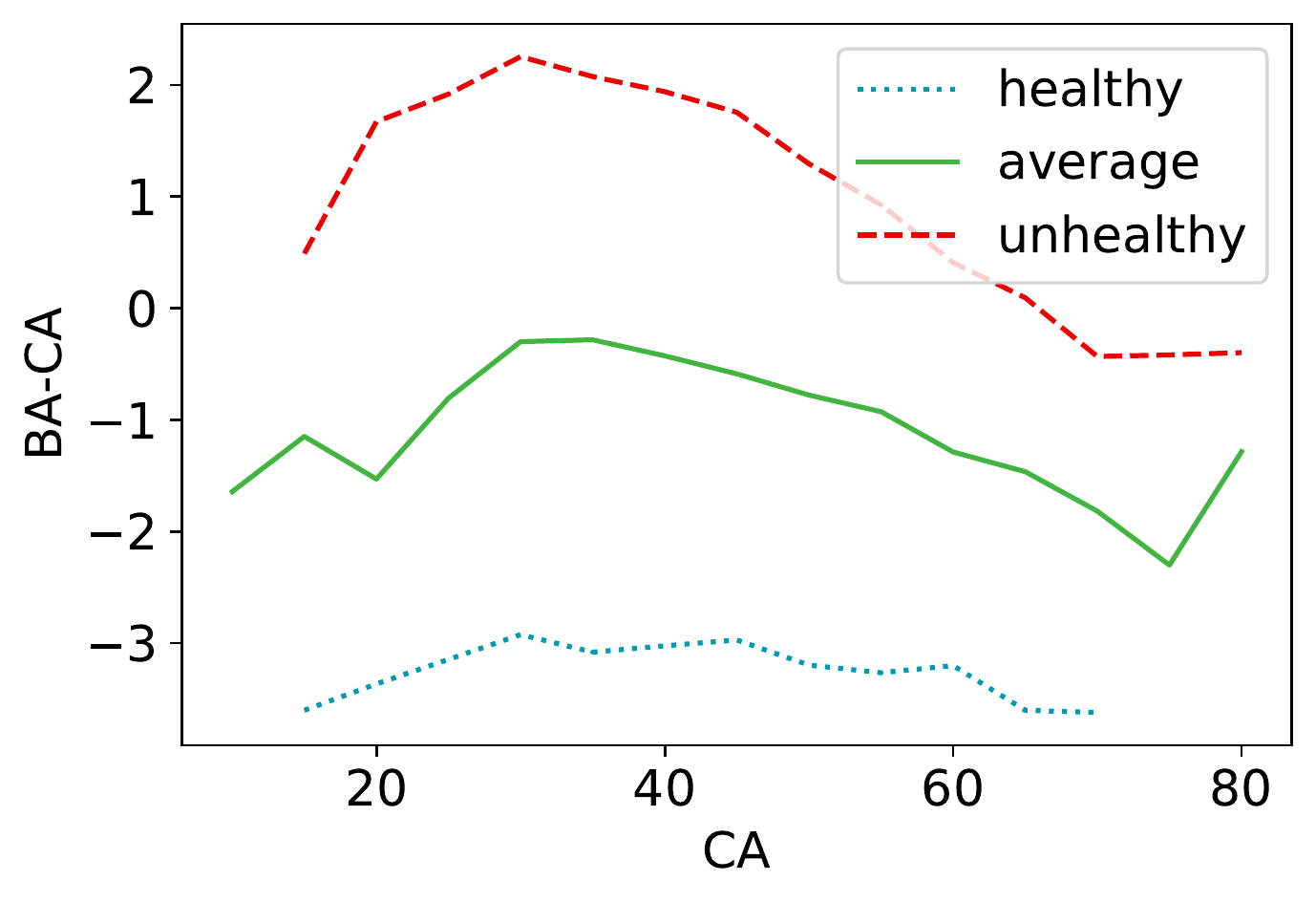}
        \caption{proposed}
    \end{subfigure}
    \caption{The gap between biological and chronological ages depending on the morbidity of \textbf{DLP} in male-\textit{whole}-\lfeature\ case.}
    \label{fig:m-all-l-morbidity-dlp}
\end{figure}

\begin{figure}
    \centering
    \begin{subfigure}[h]{.45\textwidth}
        \includegraphics[width=\textwidth]{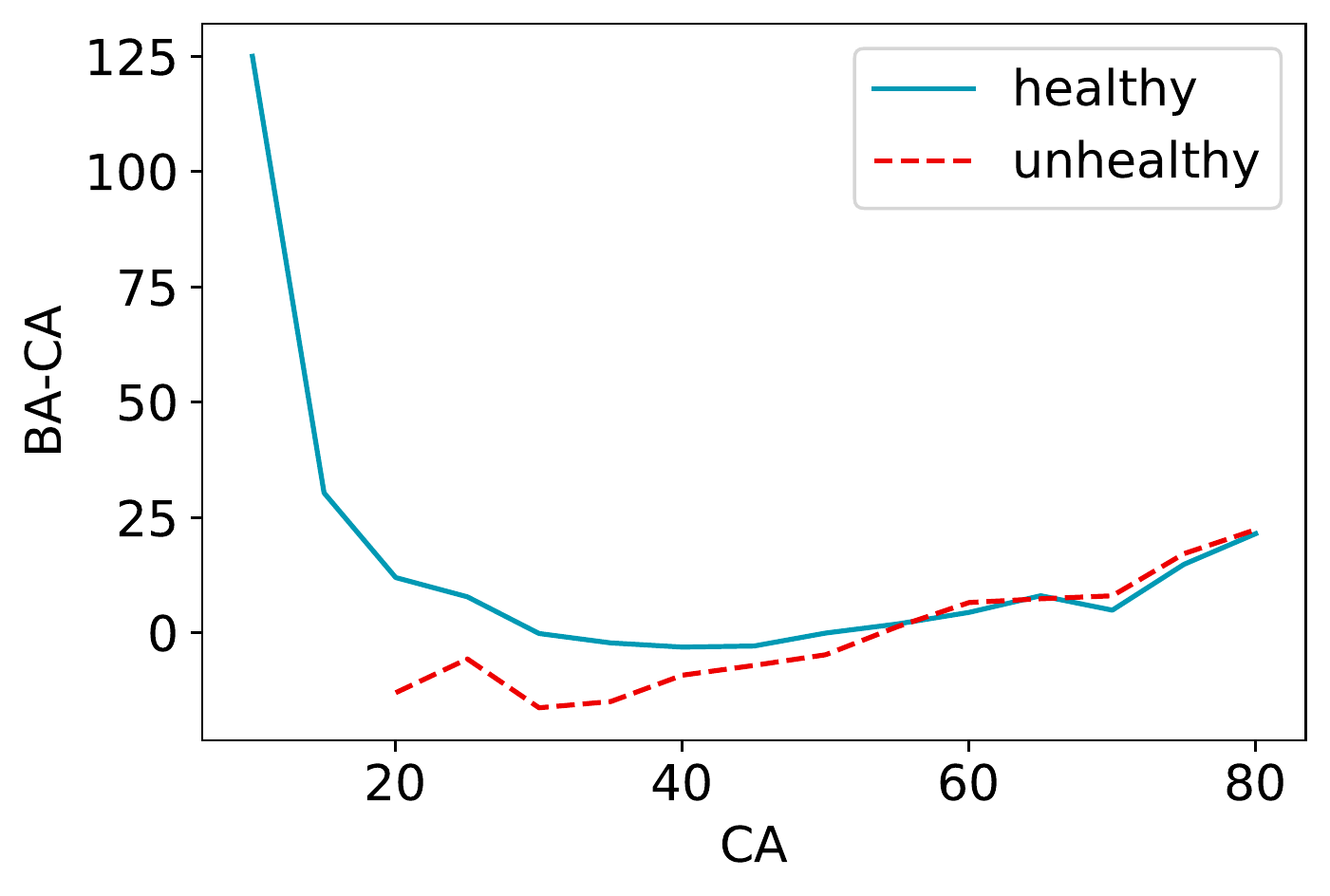}
        \caption{KDM}
    \end{subfigure}
    \begin{subfigure}[h]{.45\textwidth}
        \includegraphics[width=\textwidth]{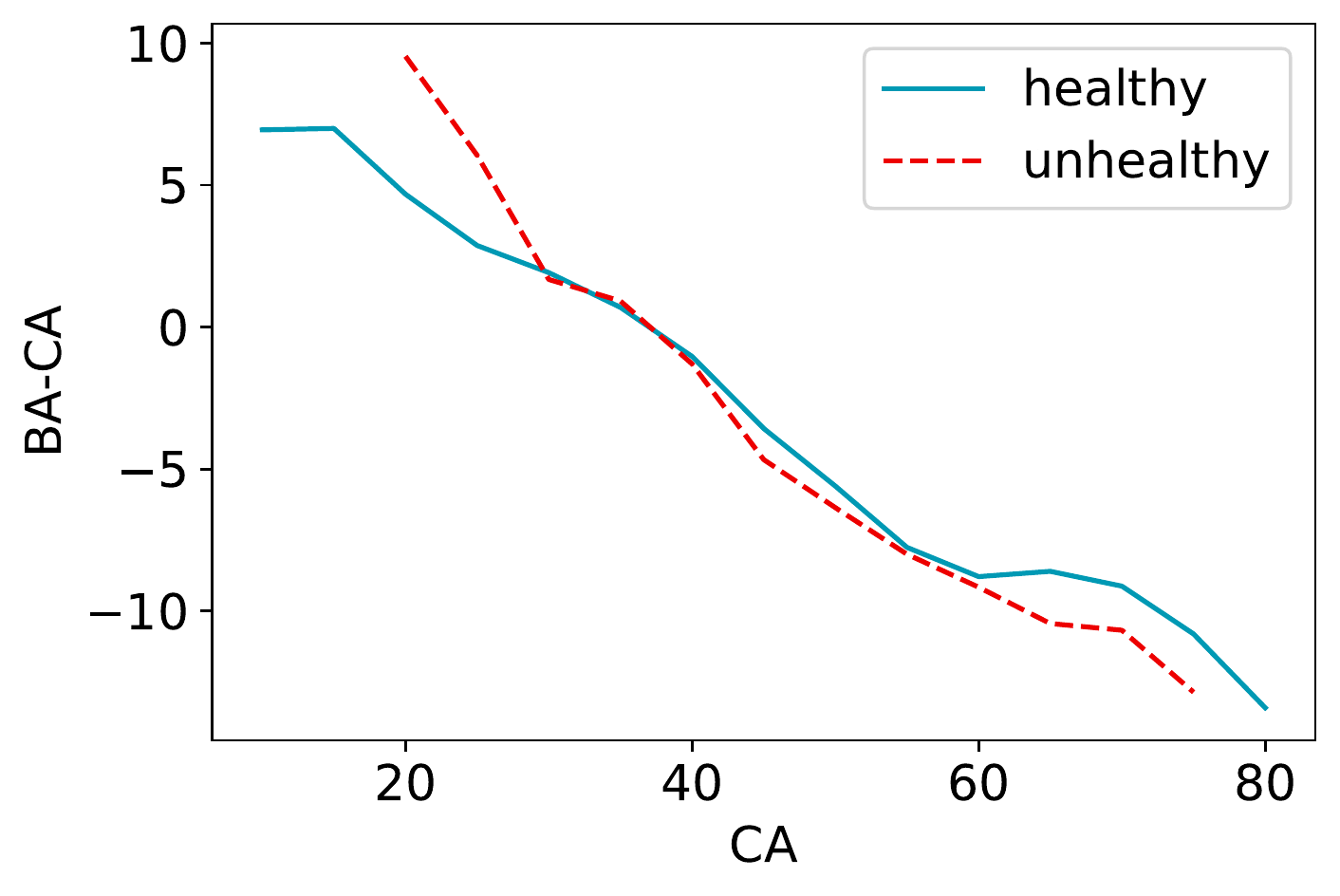}
        \caption{CAC}
    \end{subfigure}
    \begin{subfigure}[h]{.45\textwidth}
        \includegraphics[width=\textwidth]{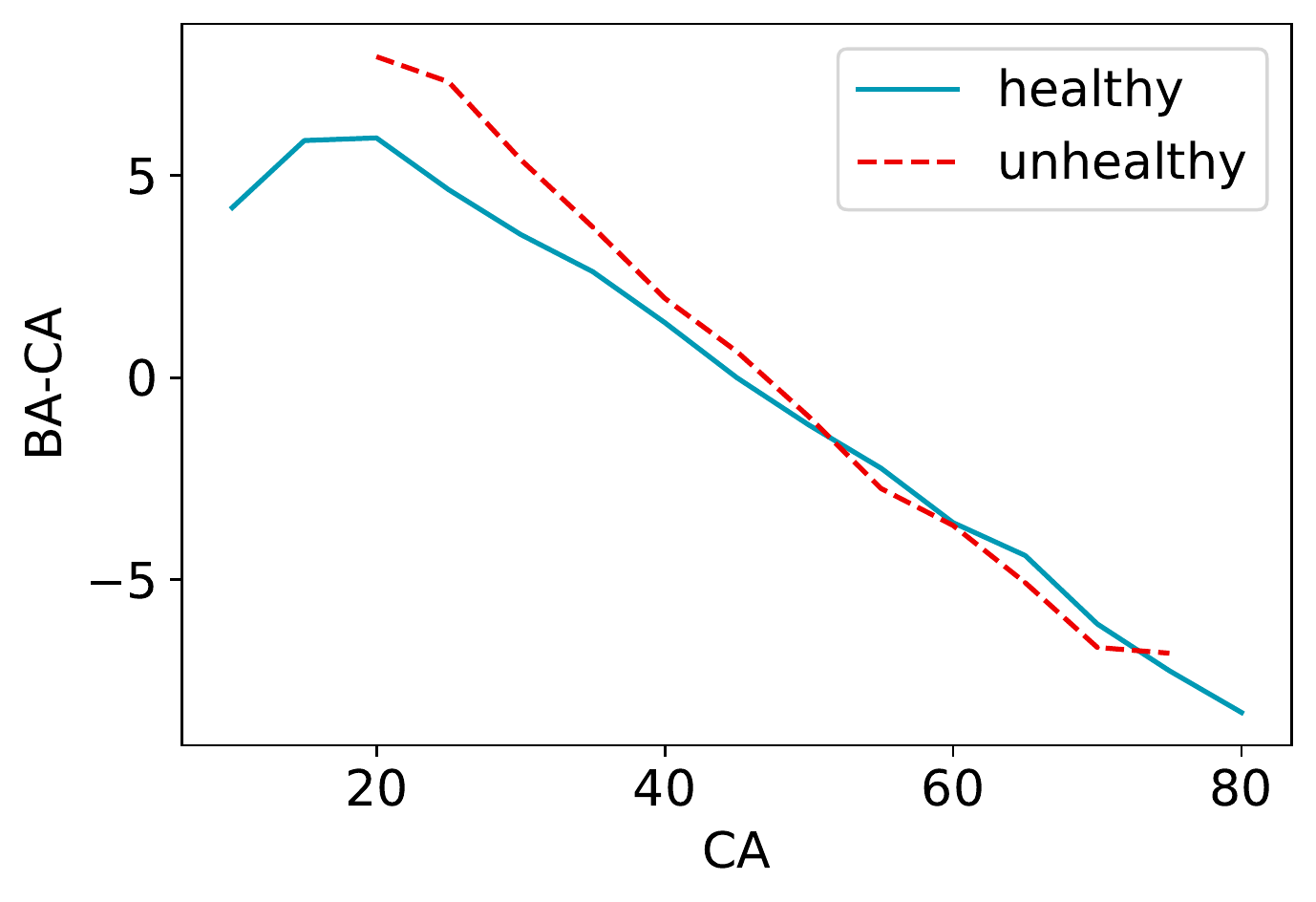}
        \caption{DNN}
    \end{subfigure}
    \begin{subfigure}[h]{.45\textwidth}
        \includegraphics[width=\textwidth]{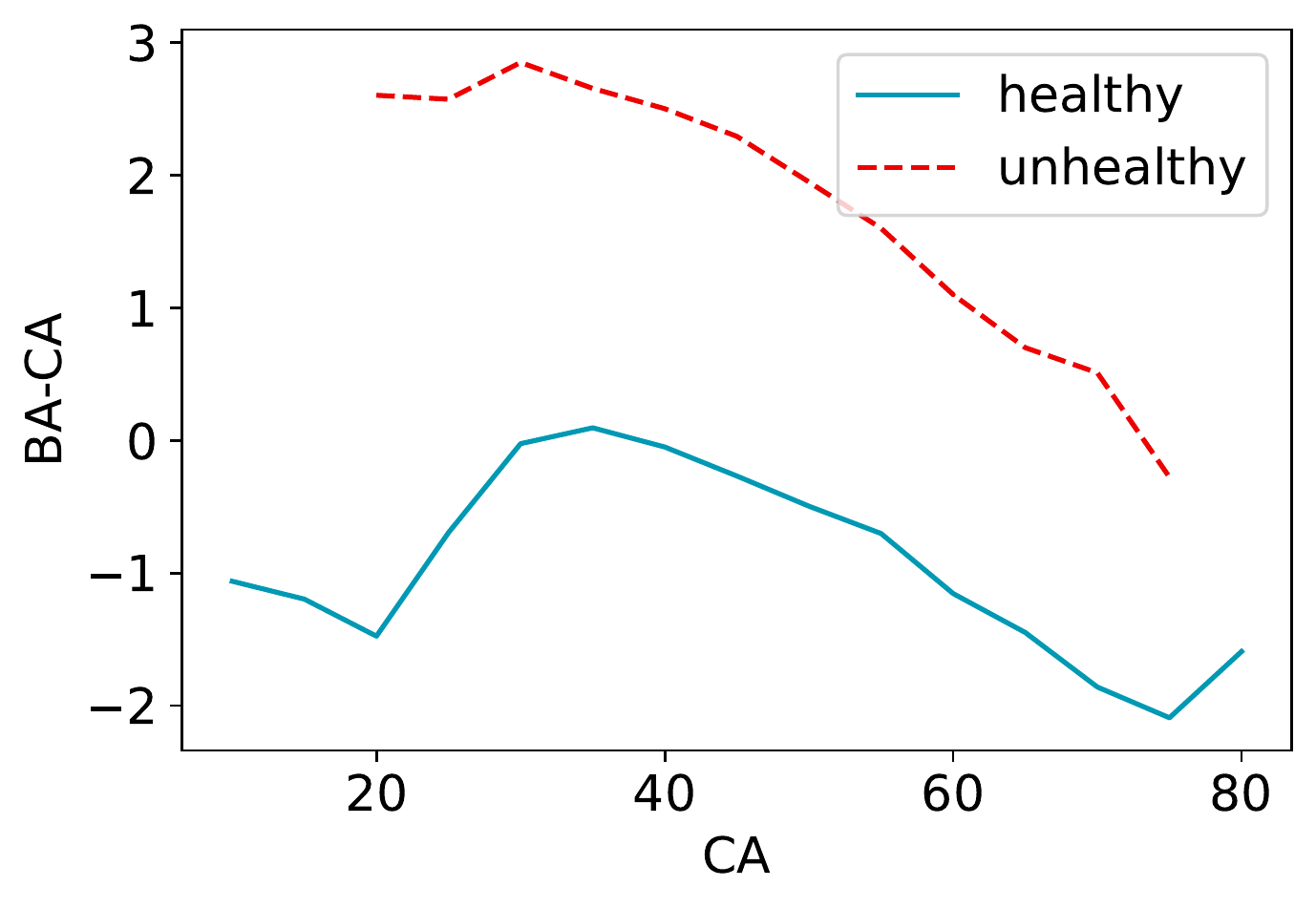}
        \caption{proposed}
    \end{subfigure}
    \caption{The gap between biological and chronological ages depending on the morbidity of \textbf{MS} in male-\textit{whole}-\lfeature\ case.}
    \label{fig:m-all-l-morbidity-ms}
\end{figure}

\begin{figure}
    \centering
    \begin{subfigure}[t]{.45\textwidth}
        \includegraphics[width=\textwidth]{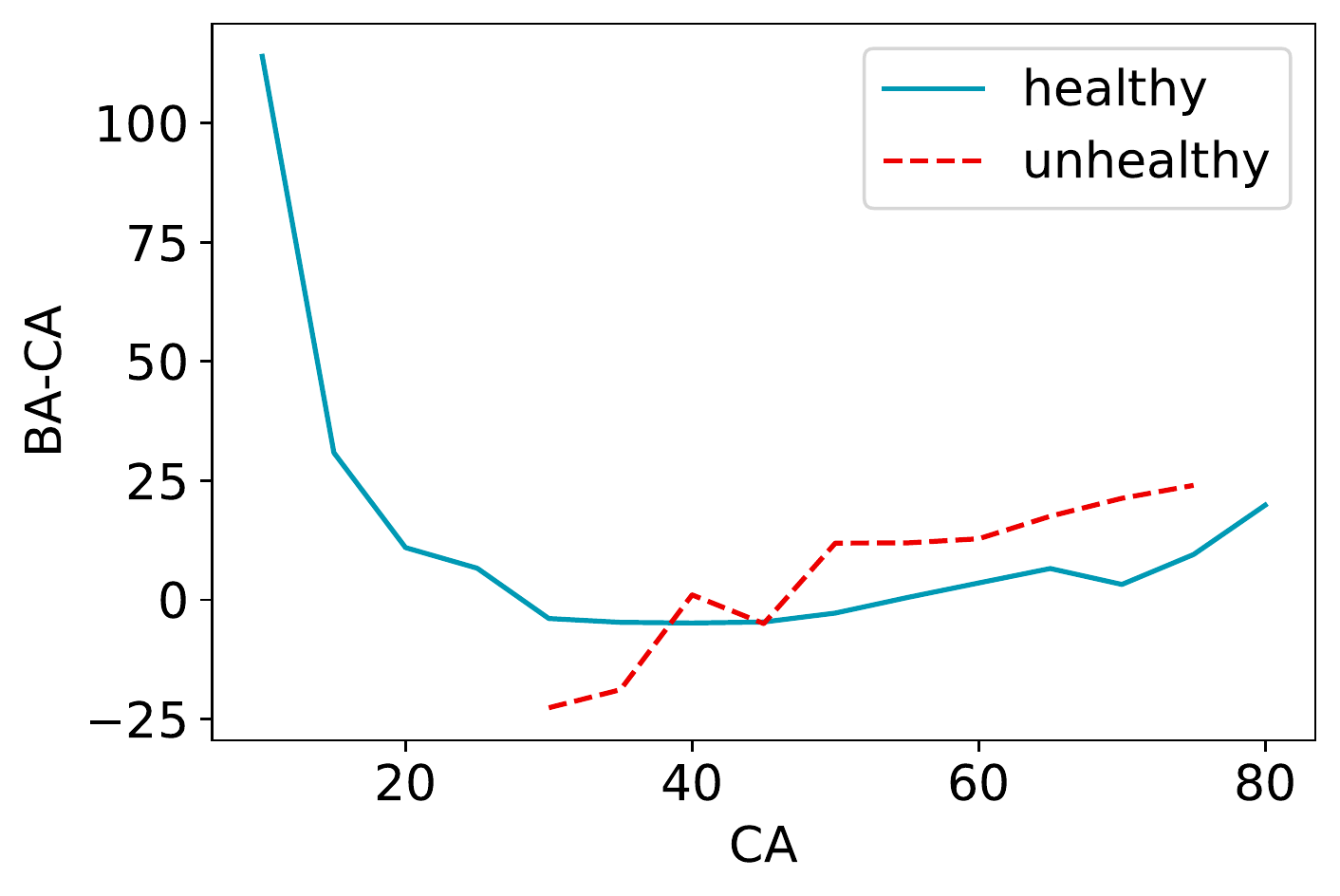}
        \caption{KDM}
    \end{subfigure}
    \begin{subfigure}[t]{.45\textwidth}
        \includegraphics[width=\textwidth]{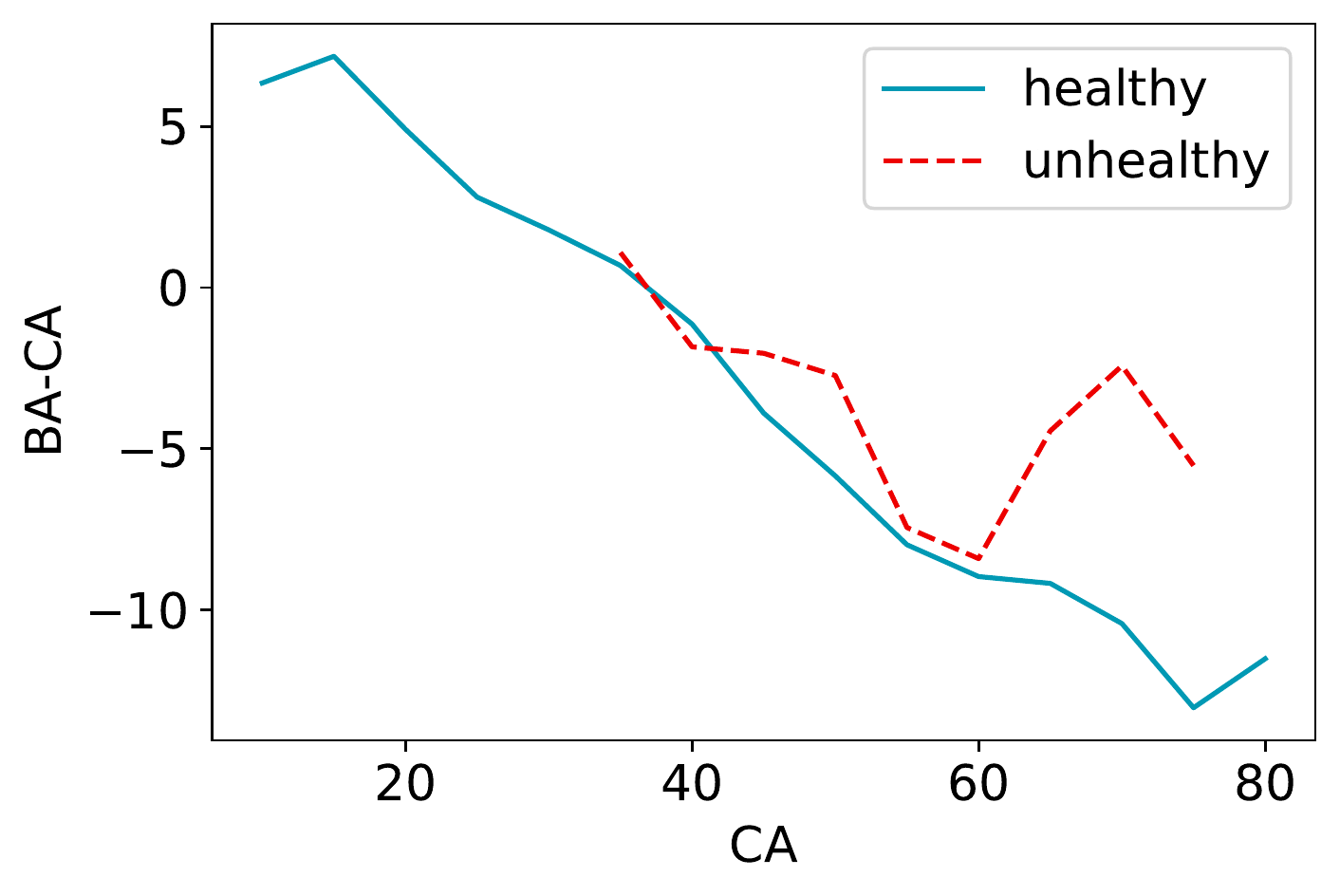}
        \caption{CAC}
    \end{subfigure}
    \begin{subfigure}[t]{.45\textwidth}
        \includegraphics[width=\textwidth]{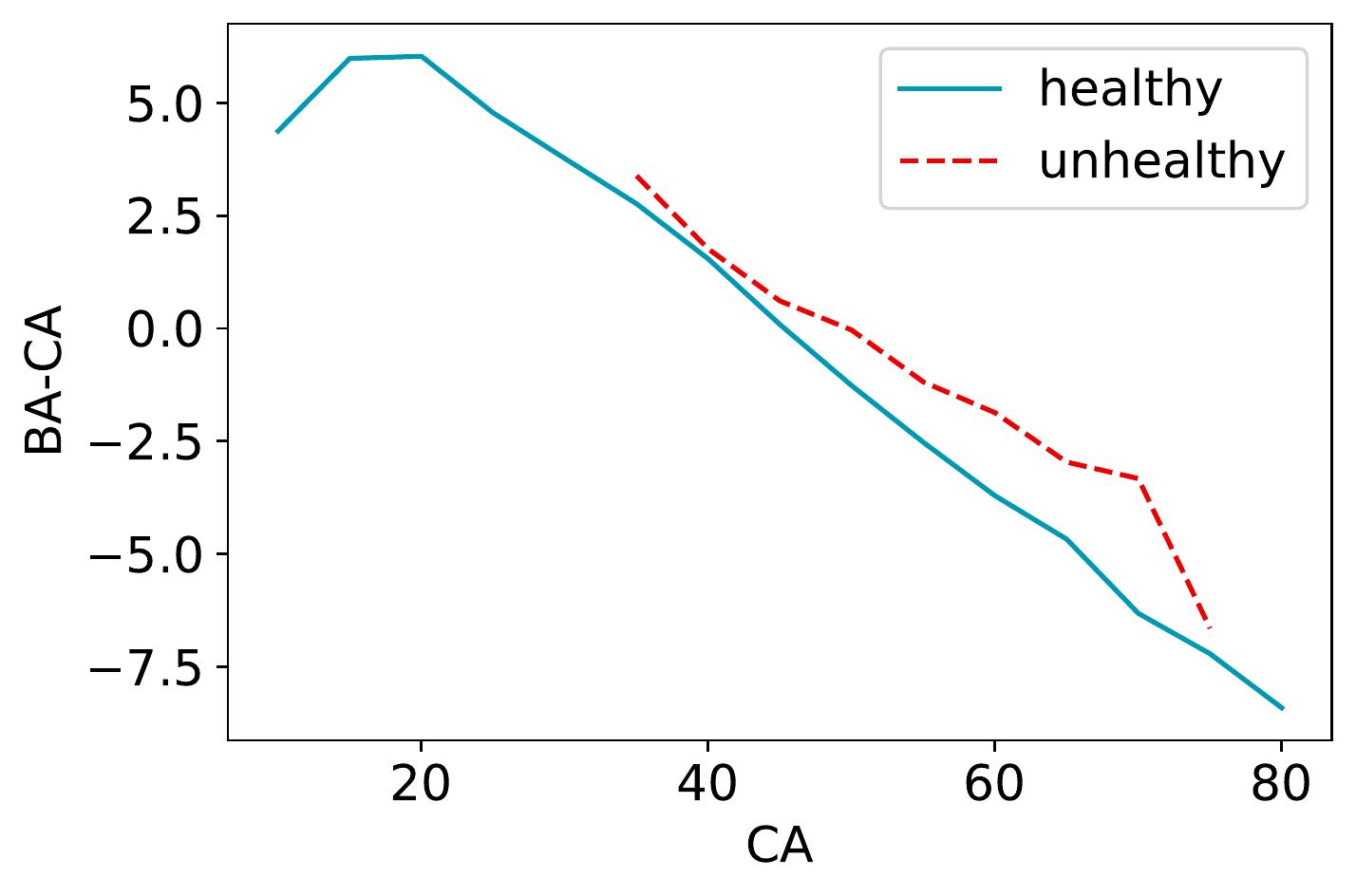}
        \caption{DNN}
    \end{subfigure}
    \begin{subfigure}[t]{.45\textwidth}
        \includegraphics[width=\textwidth]{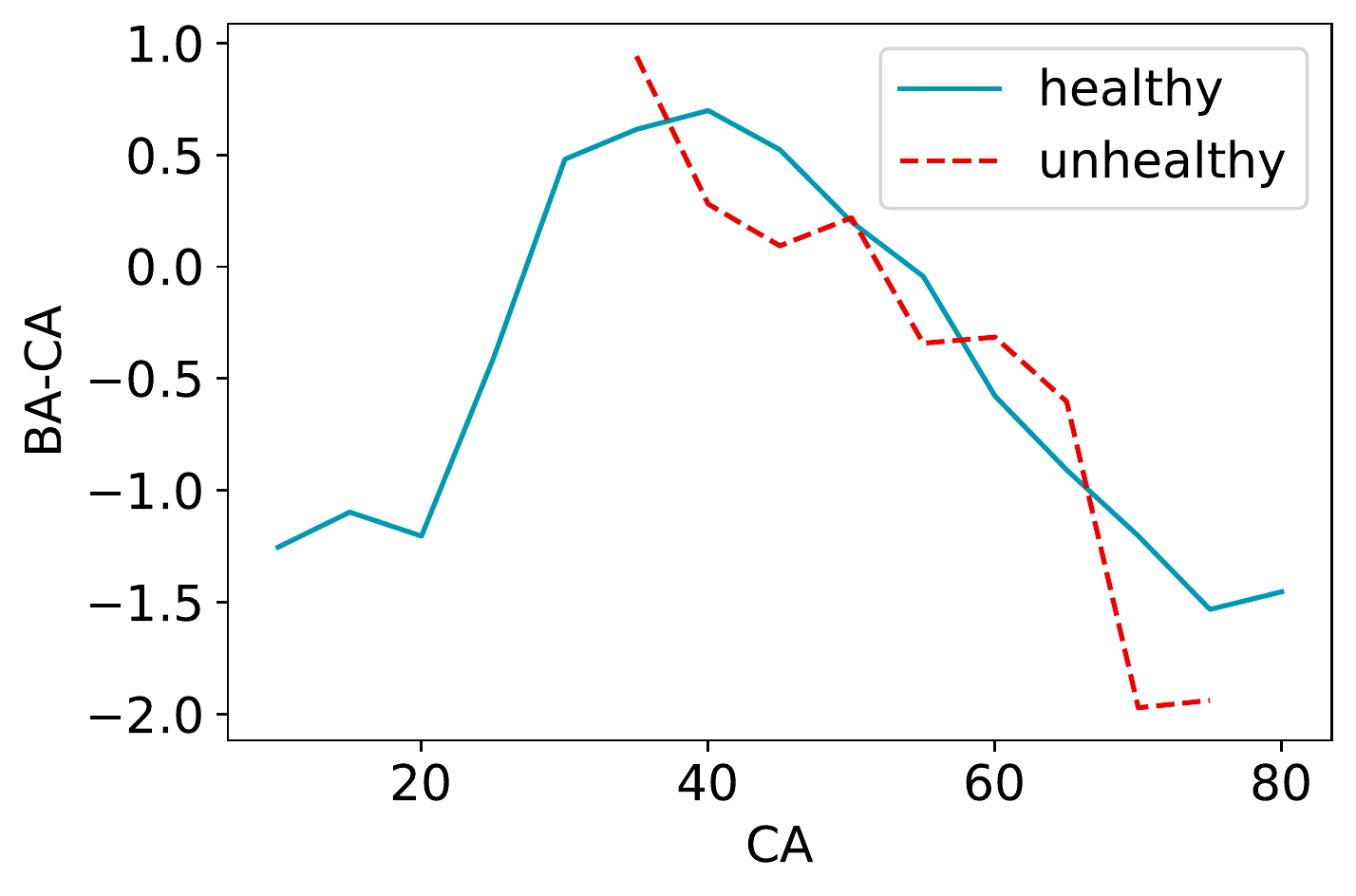}
        \caption{proposed}
    \end{subfigure}
    \caption{The gap between biological and chronological ages depending on the morbidity of \textbf{CVD} in male-\textit{whole}-\lfeature\ case.}
    \label{fig:m-all-l-morbidity-cvd}
\end{figure}

\begin{figure}
    \centering
    \begin{subfigure}[t]{.45\textwidth}
        \includegraphics[width=\textwidth]{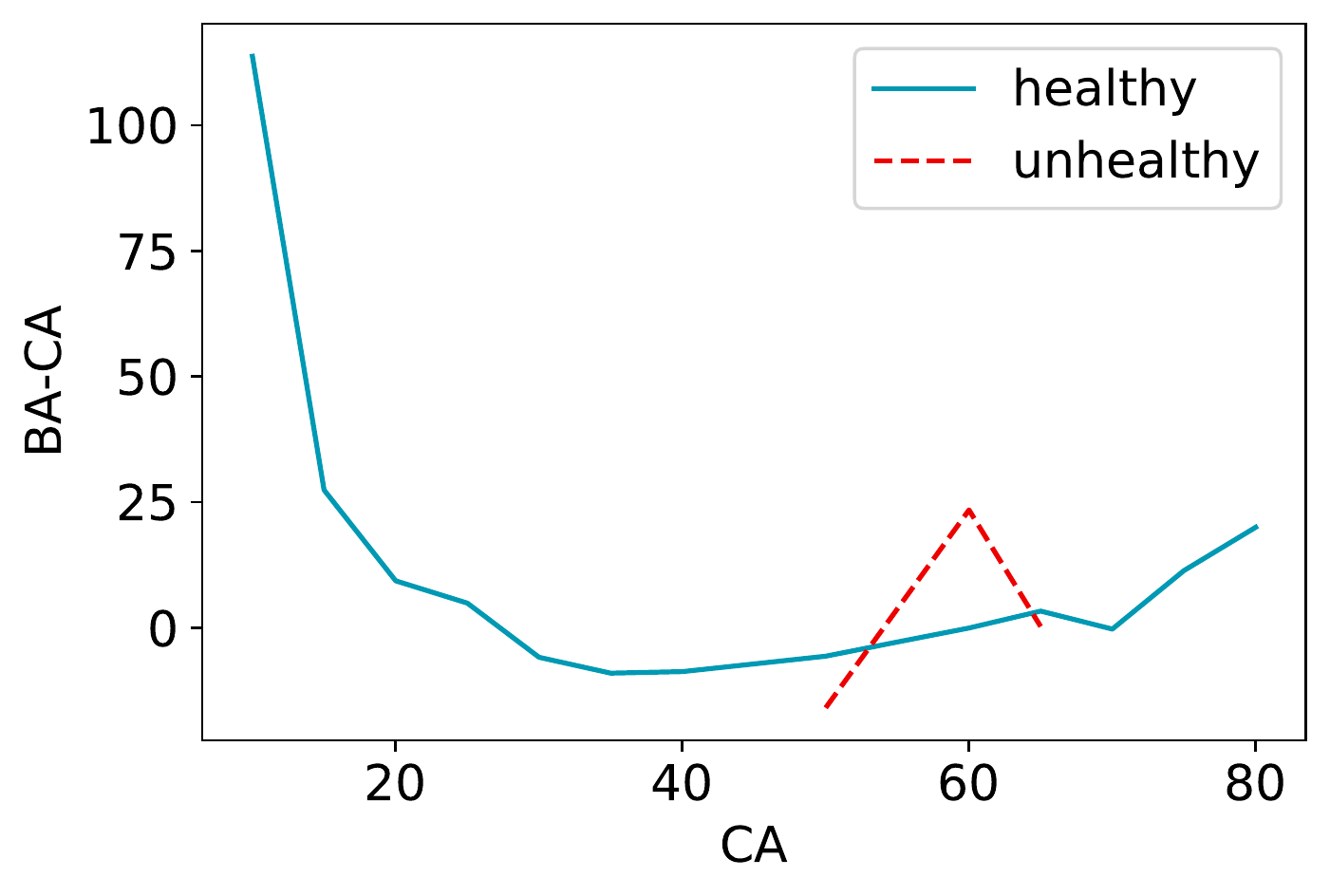}
        \caption{KDM}
    \end{subfigure}
    \begin{subfigure}[t]{.45\textwidth}
        \includegraphics[width=\textwidth]{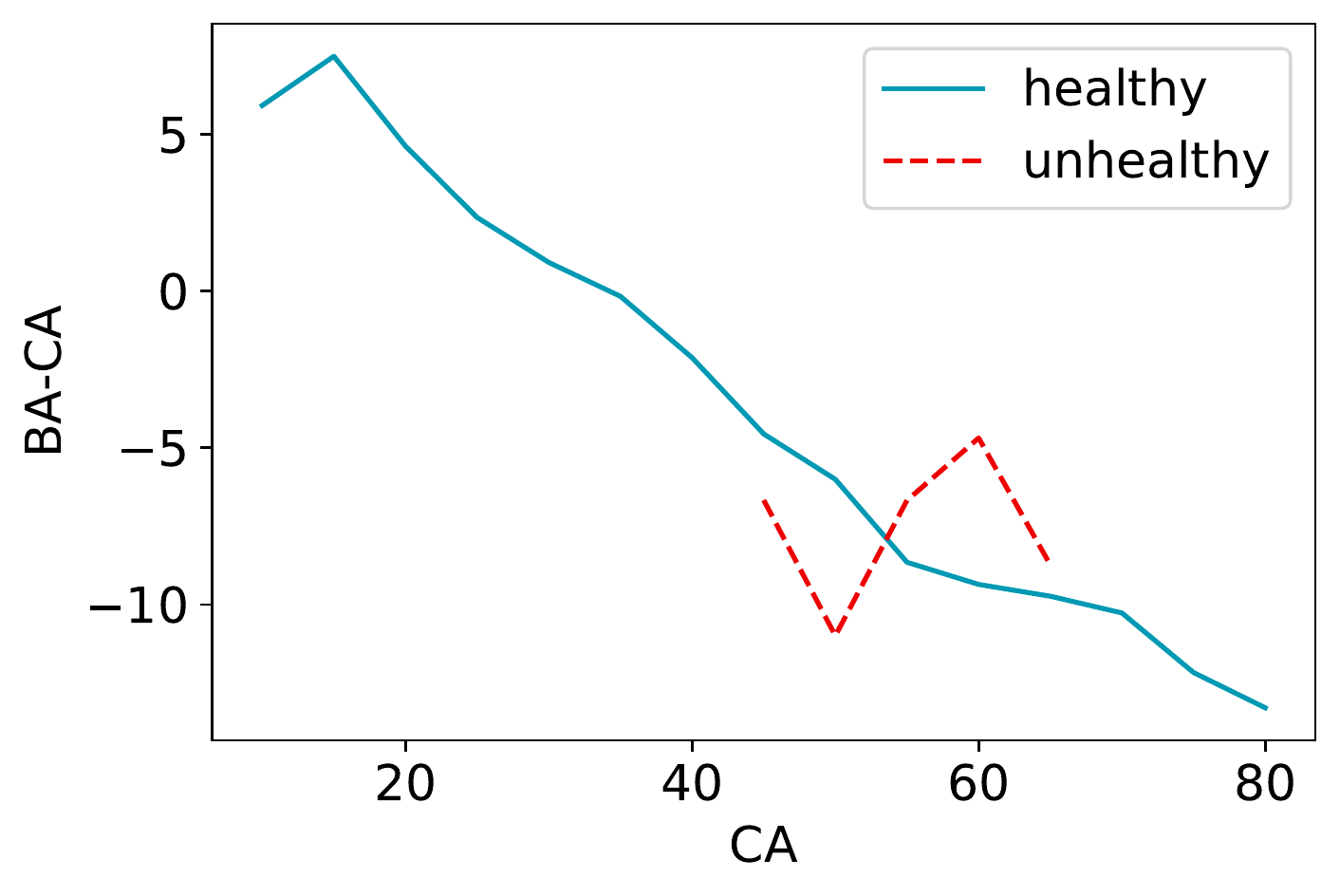}
        \caption{CAC}
    \end{subfigure}
    \begin{subfigure}[t]{.45\textwidth}
        \includegraphics[width=\textwidth]{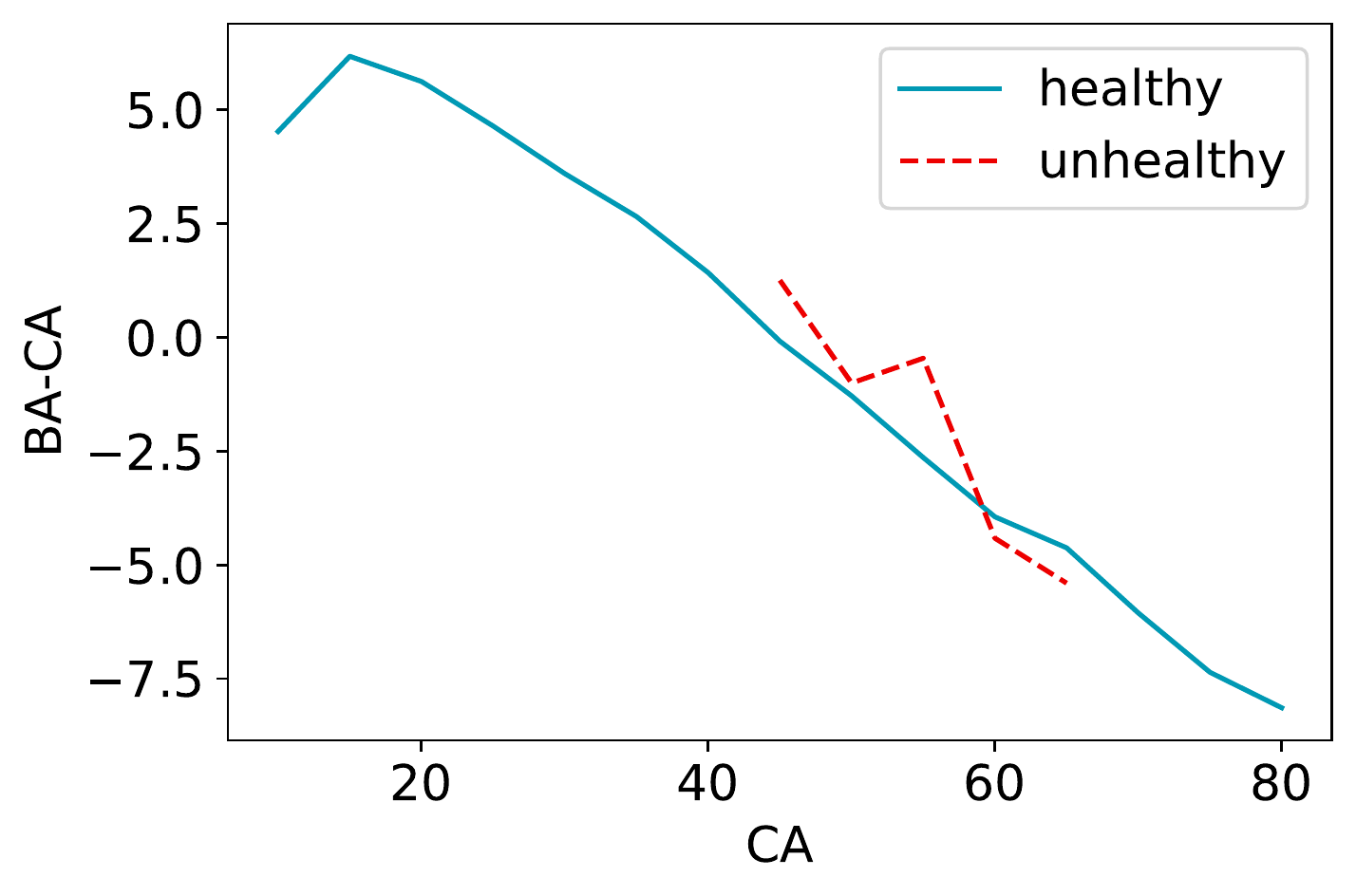}
        \caption{DNN}
    \end{subfigure}
    \begin{subfigure}[t]{.45\textwidth}
        \includegraphics[width=\textwidth]{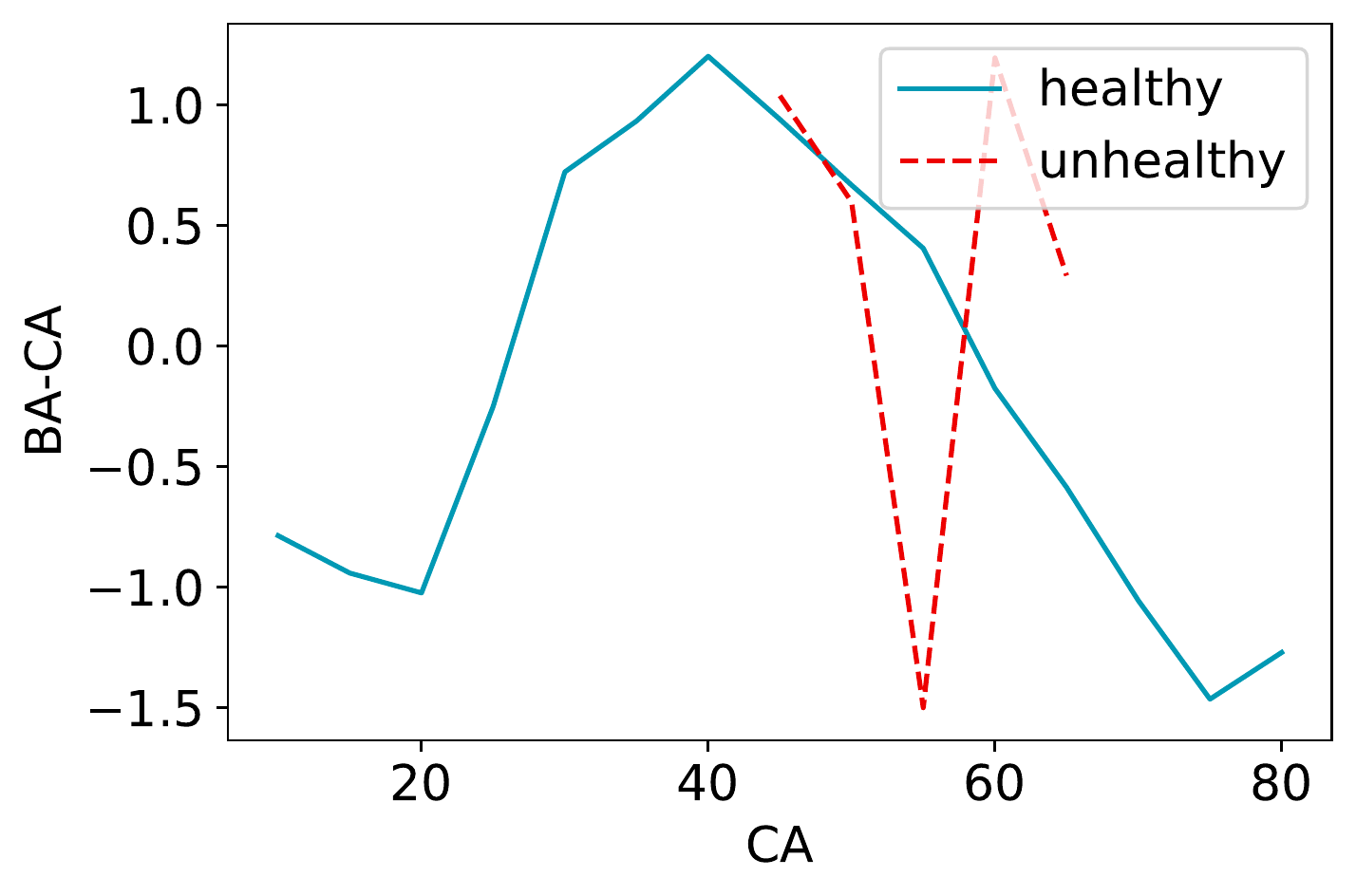}
        \caption{proposed}
    \end{subfigure}
    \caption{The gap between biological and chronological ages depending on the morbidity of \textbf{CVA} in male-\textit{whole}-\lfeature\ case.}
    \label{fig:m-all-l-morbidity-cva}
\end{figure}

\begin{figure}
    \centering
    \begin{subfigure}[h]{.45\textwidth}
        \includegraphics[width=\textwidth]{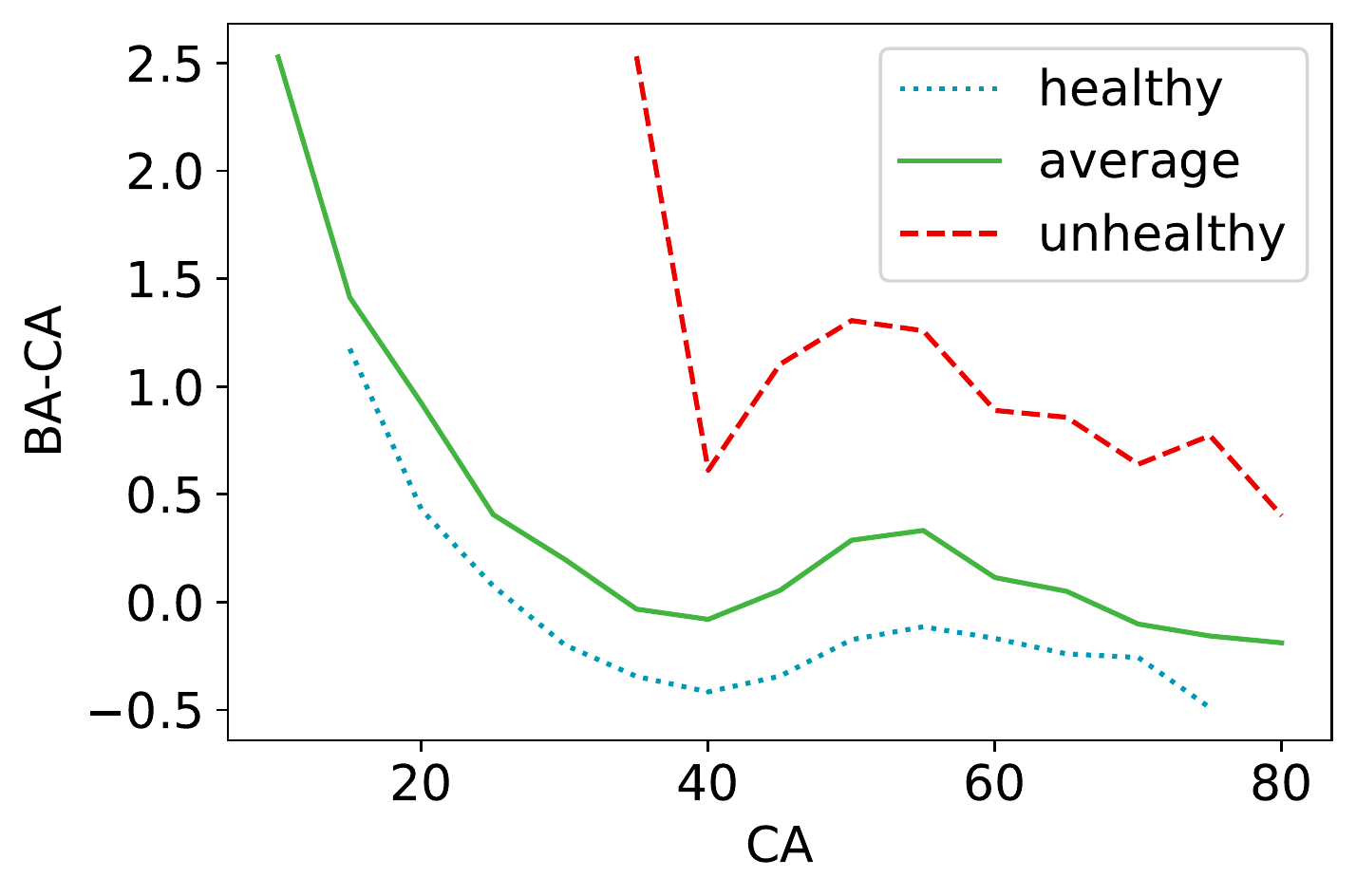}
        \caption{KDM}
    \end{subfigure}
    \begin{subfigure}[h]{.45\textwidth}
        \includegraphics[width=\textwidth]{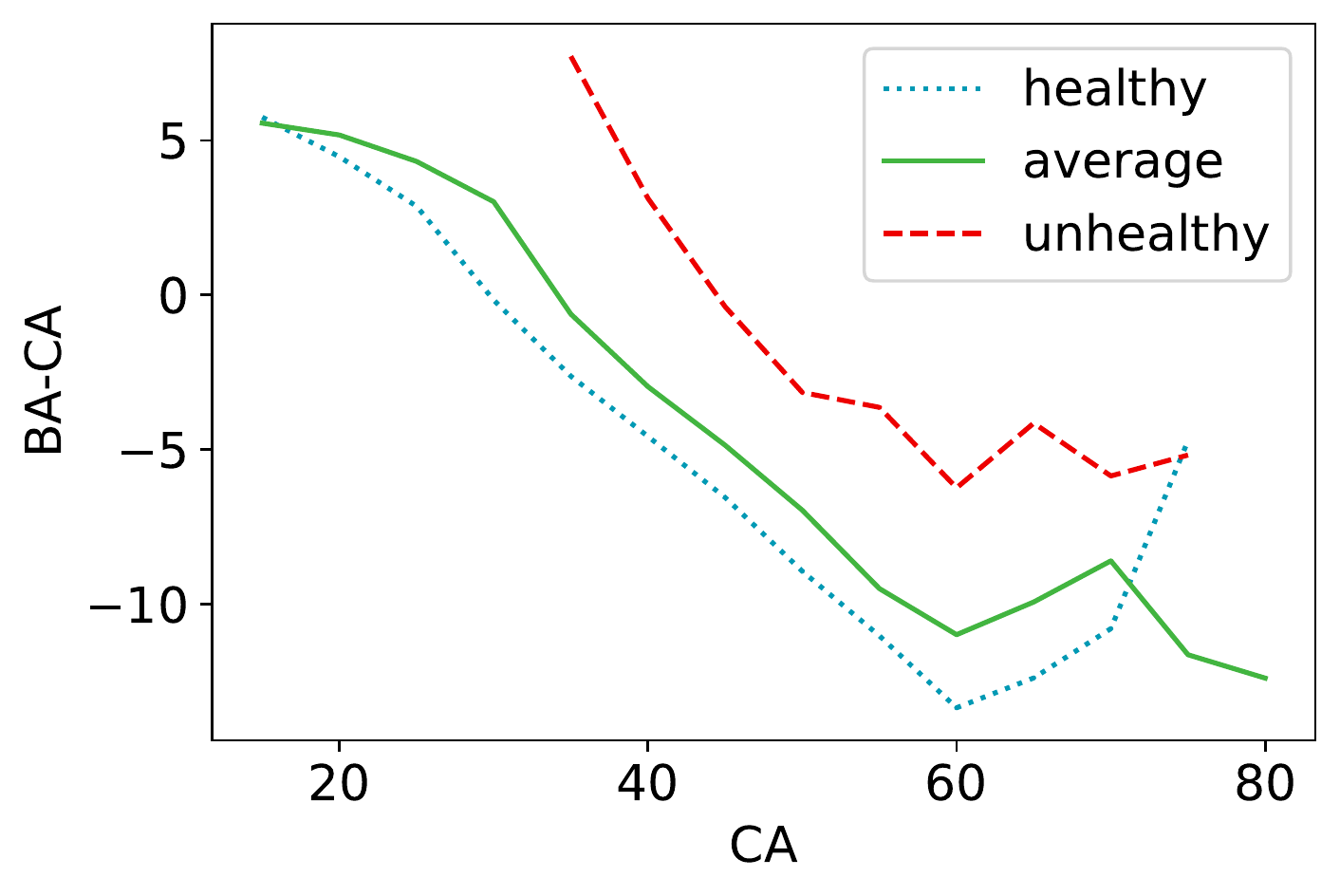}
        \caption{CAC}
    \end{subfigure}
    \begin{subfigure}[h]{.45\textwidth}
        \includegraphics[width=\textwidth]{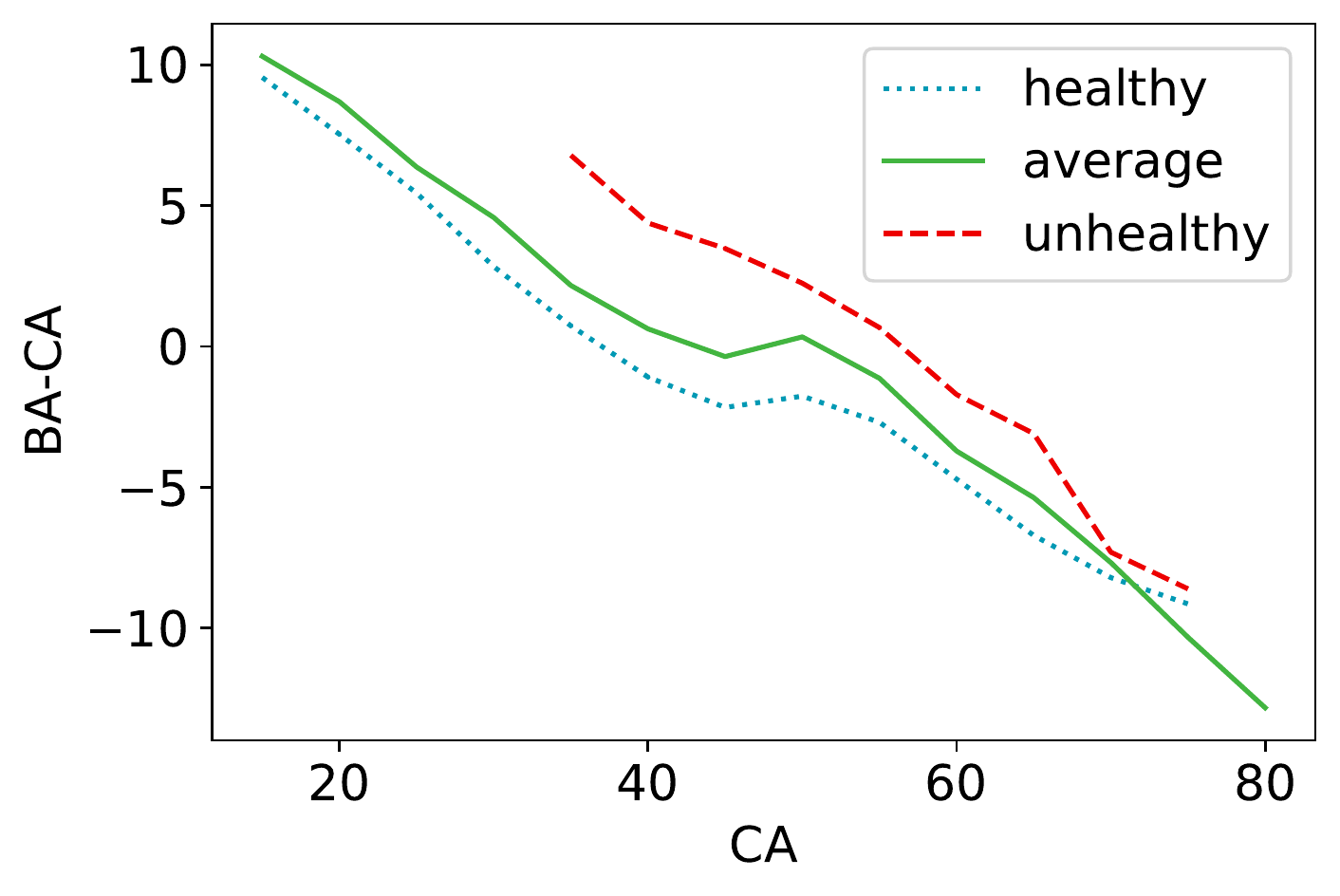}
        \caption{DNN}
    \end{subfigure}
    \begin{subfigure}[h]{.45\textwidth}
        \includegraphics[width=\textwidth]{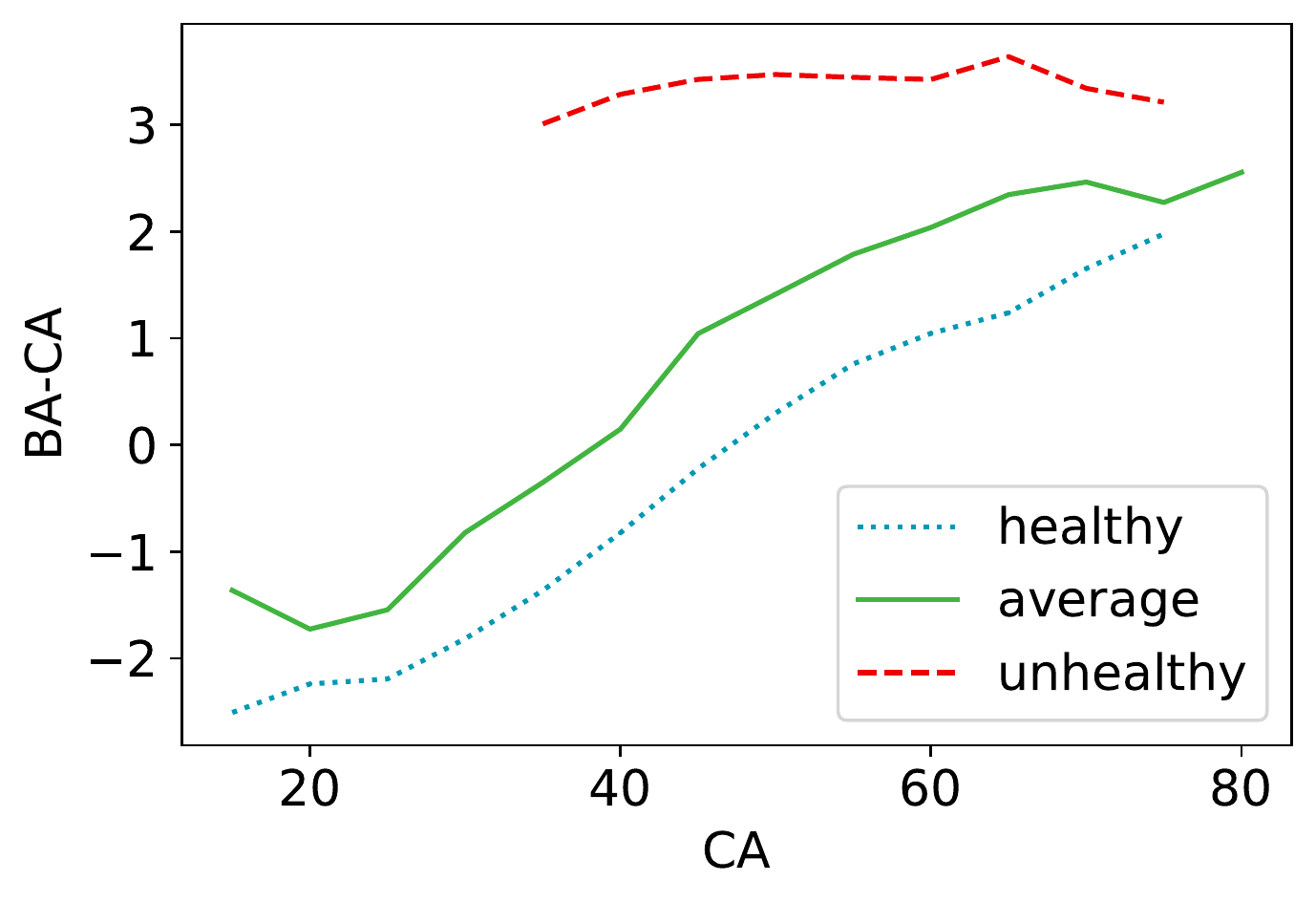}
        \caption{proposed}
    \end{subfigure}
    \caption{The gap between biological and chronological ages depending on the morbidity of \textbf{DM} in female-\textit{whole}-\lfeature\ case.}
    \label{fig:f-all-l-morbidity-dm}
\end{figure}

\begin{figure}
    \centering
    \begin{subfigure}[h]{.45\textwidth}
        \includegraphics[width=\textwidth]{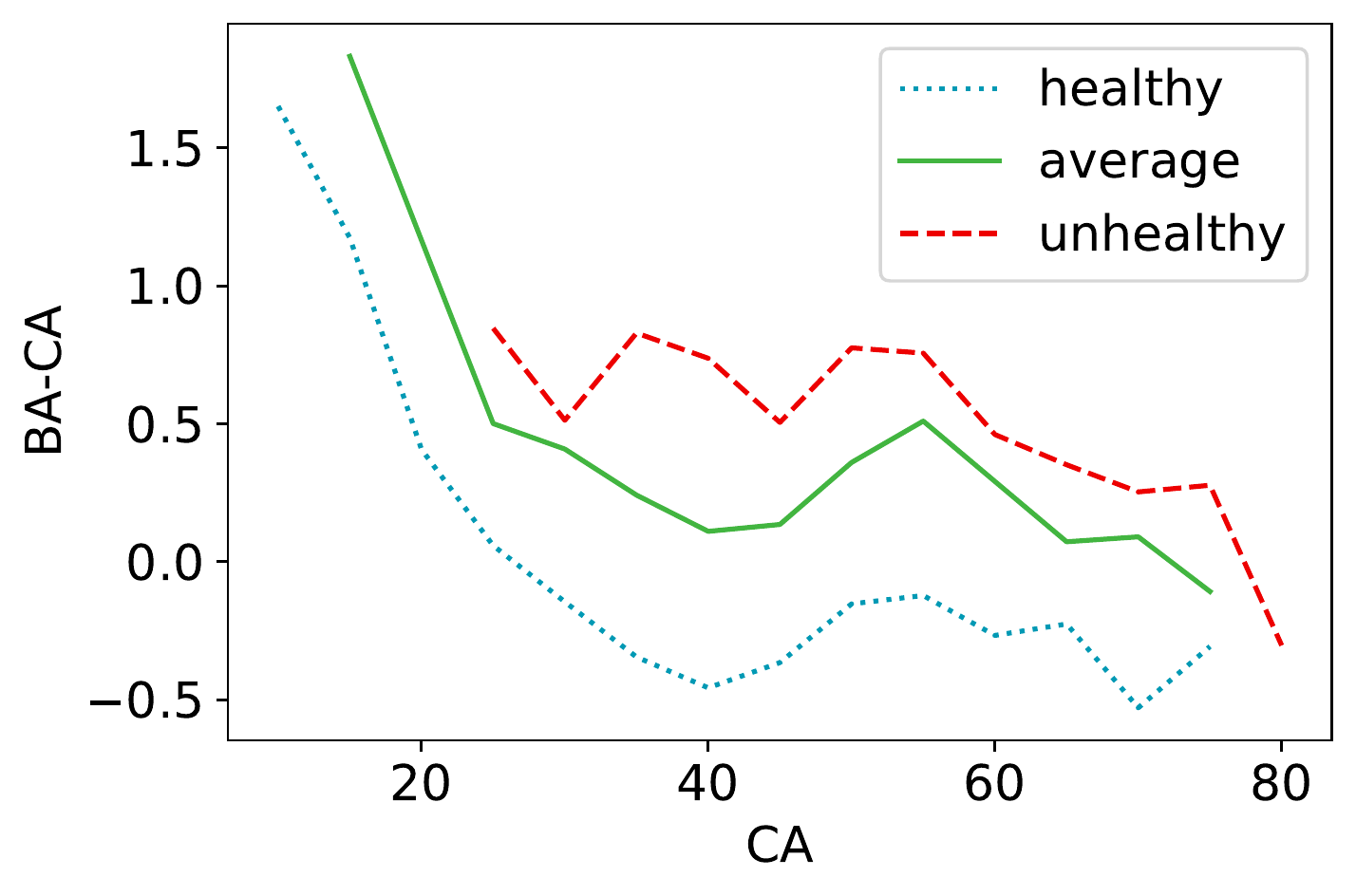}
        \caption{KDM}
    \end{subfigure}
    \begin{subfigure}[h]{.45\textwidth}
        \includegraphics[width=\textwidth]{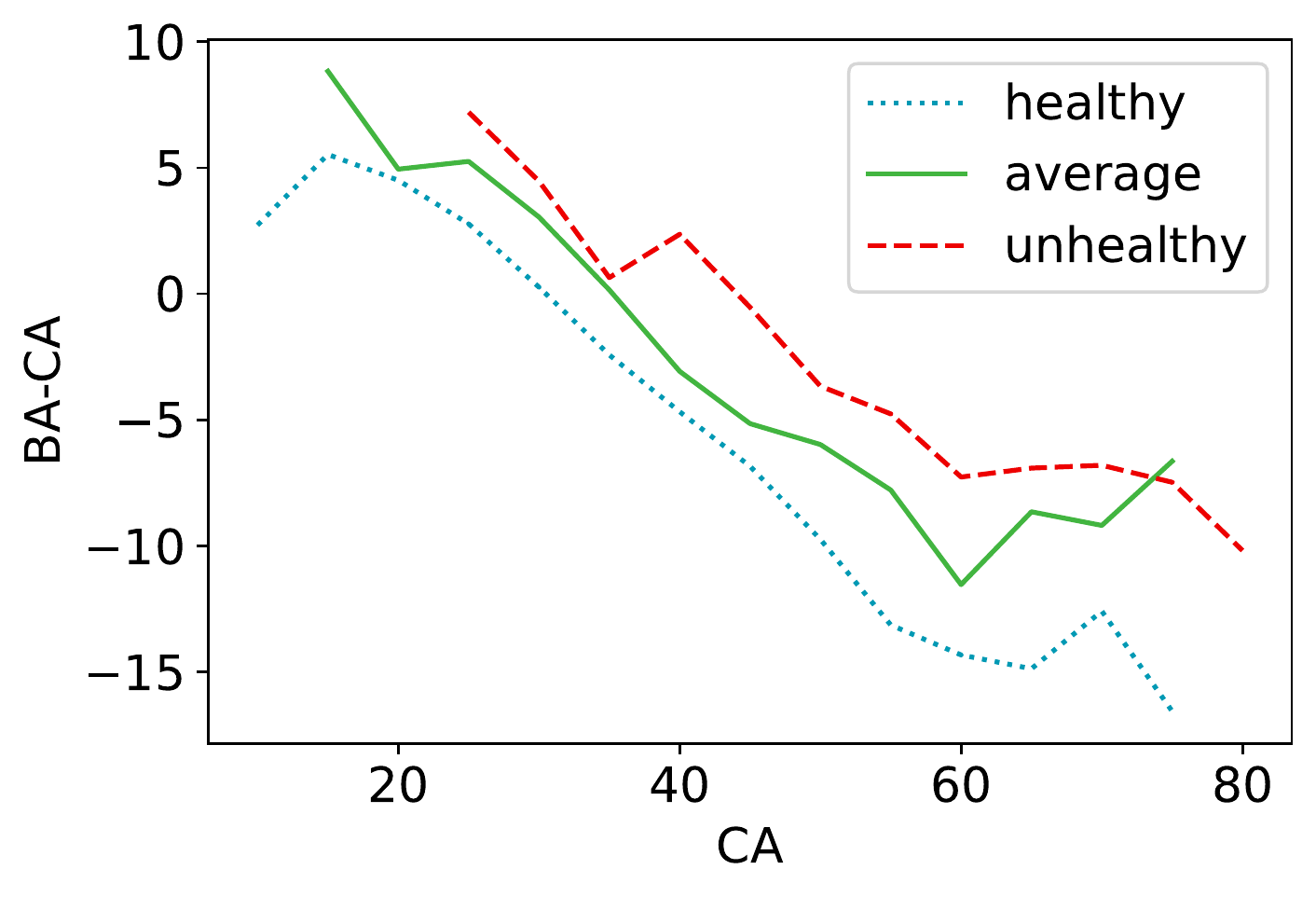}
        \caption{CAC}
    \end{subfigure}
    \begin{subfigure}[h]{.45\textwidth}
        \includegraphics[width=\textwidth]{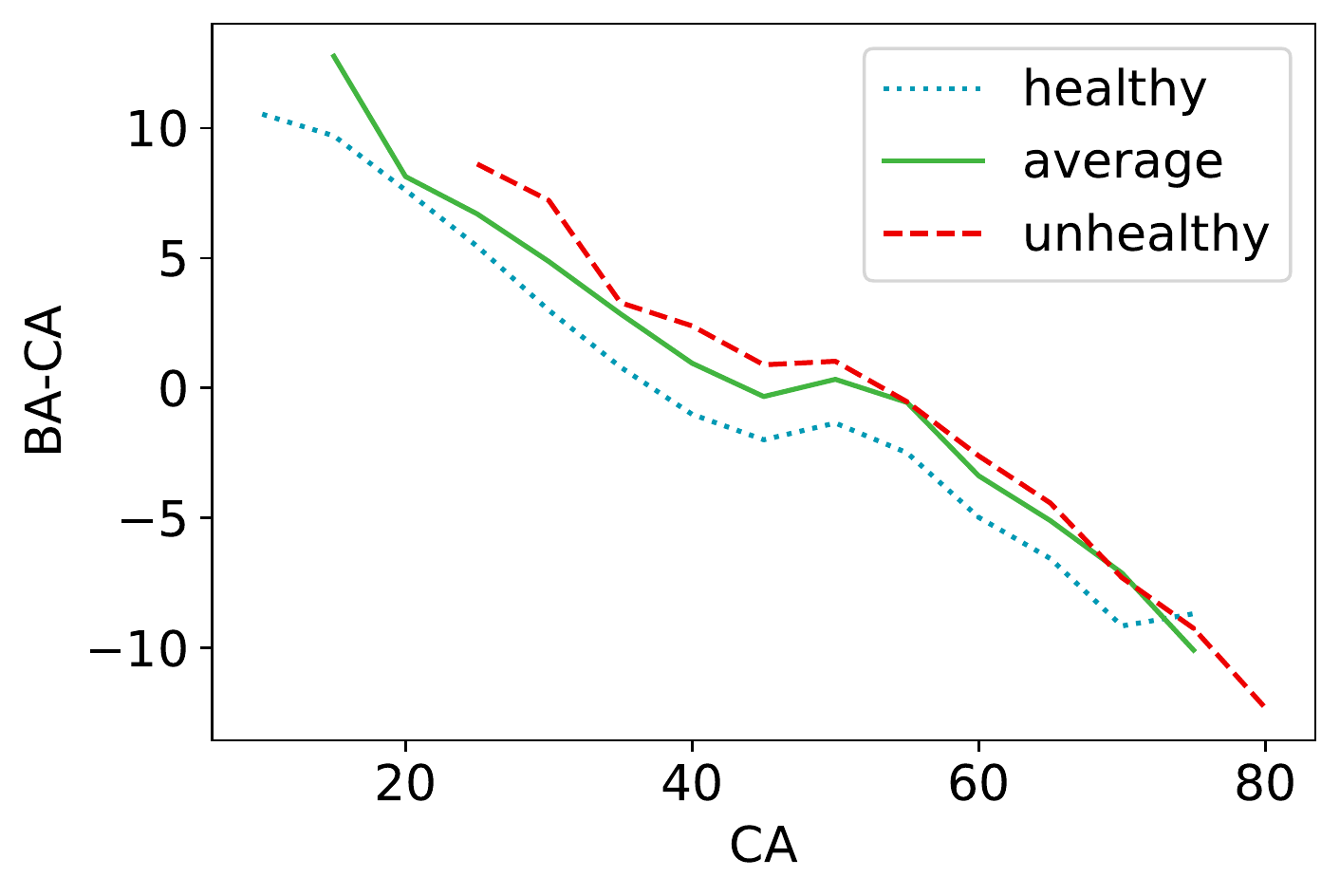}
        \caption{DNN}
    \end{subfigure}
    \begin{subfigure}[h]{.45\textwidth}
        \includegraphics[width=\textwidth]{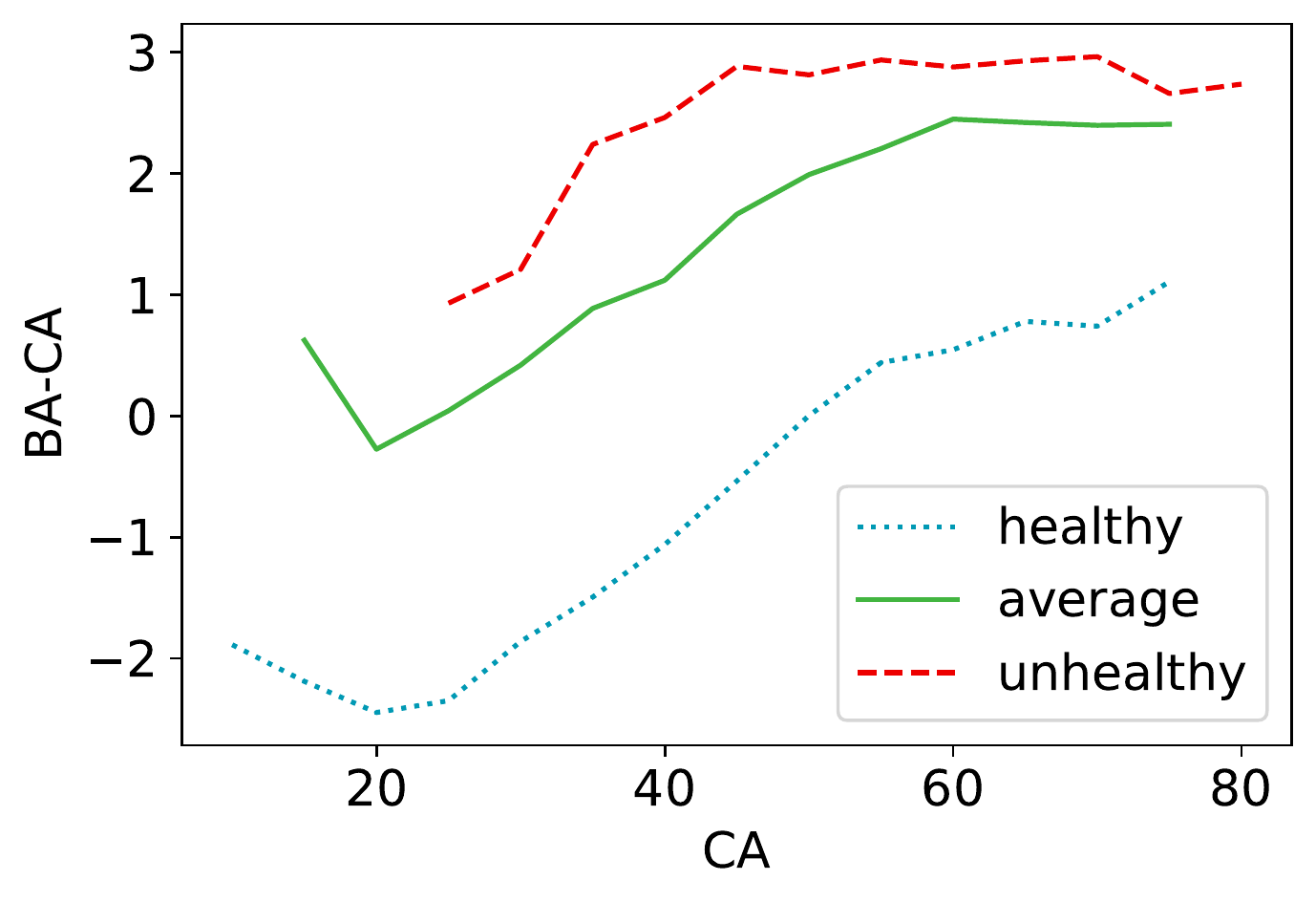}
        \caption{proposed}
    \end{subfigure}
    \caption{The gap between biological and chronological ages depending on the morbidity of \textbf{HBP} in female-\textit{whole}-\lfeature\ case.}
    \label{fig:f-all-l-morbidity-hbp}
\end{figure}

\begin{figure}
    \centering
    \begin{subfigure}[h]{.45\textwidth}
        \includegraphics[width=\textwidth]{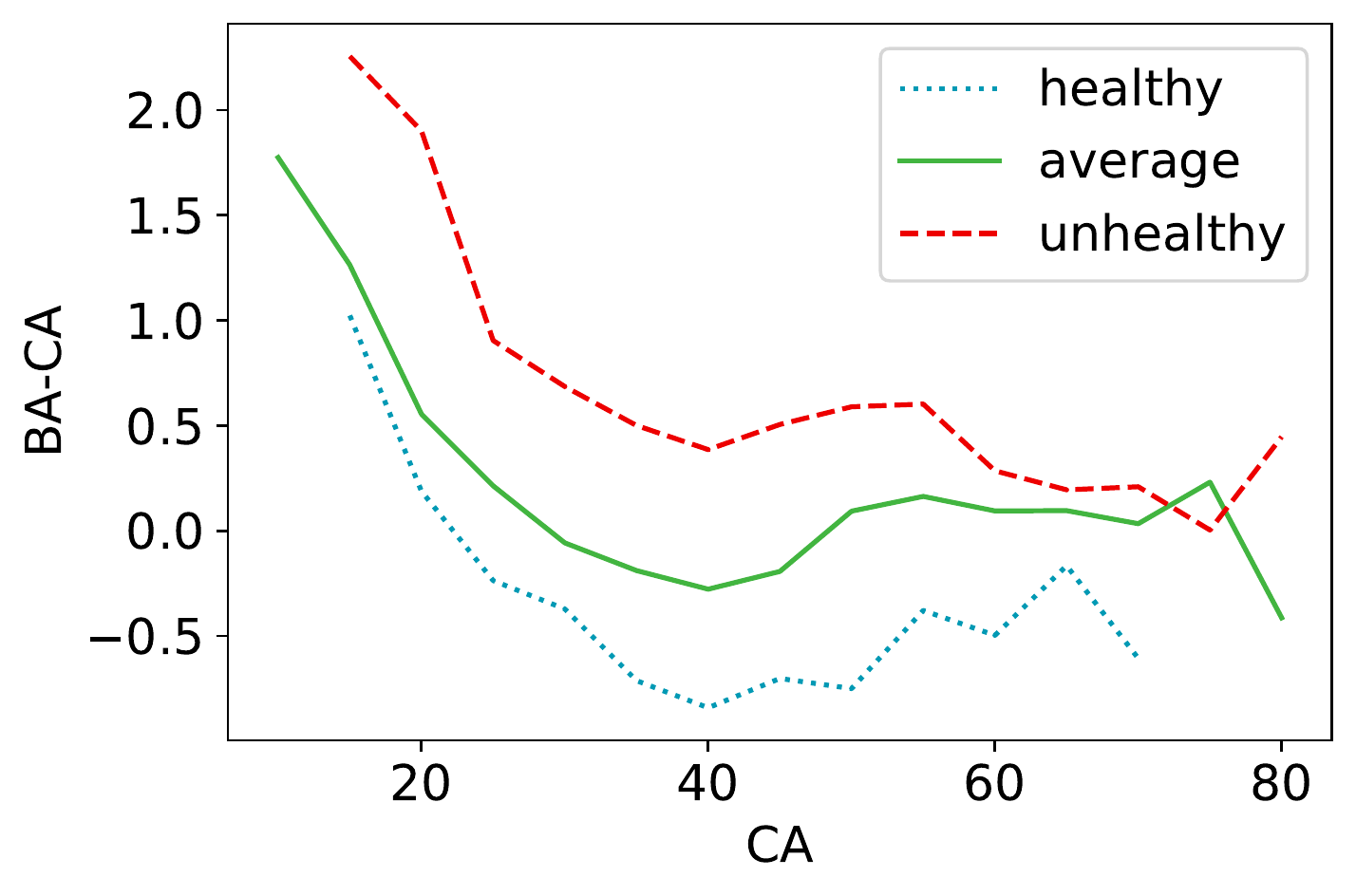}
        \caption{KDM}
    \end{subfigure}
    \begin{subfigure}[h]{.45\textwidth}
        \includegraphics[width=\textwidth]{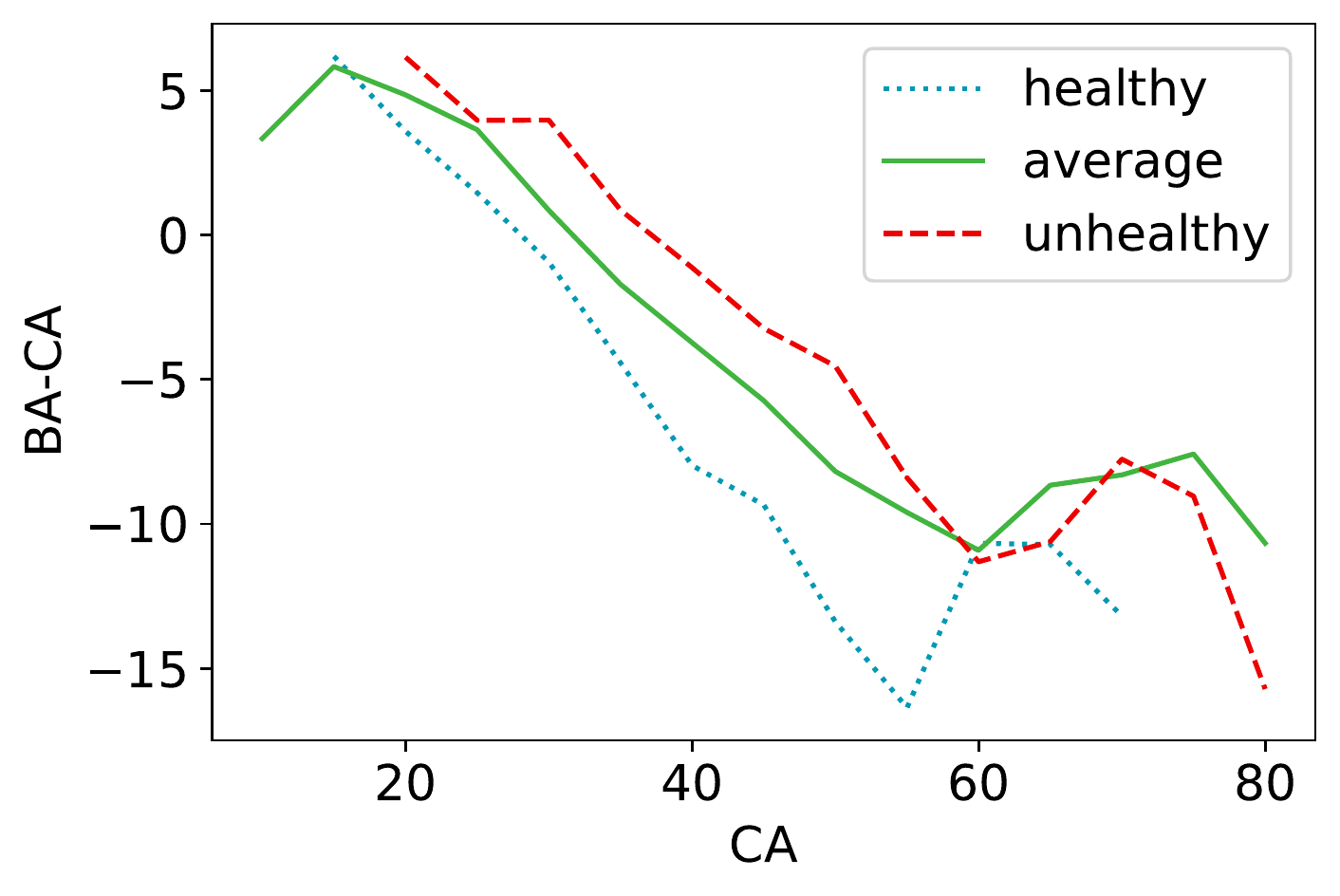}
        \caption{CAC}
    \end{subfigure}
    \begin{subfigure}[h]{.45\textwidth}
        \includegraphics[width=\textwidth]{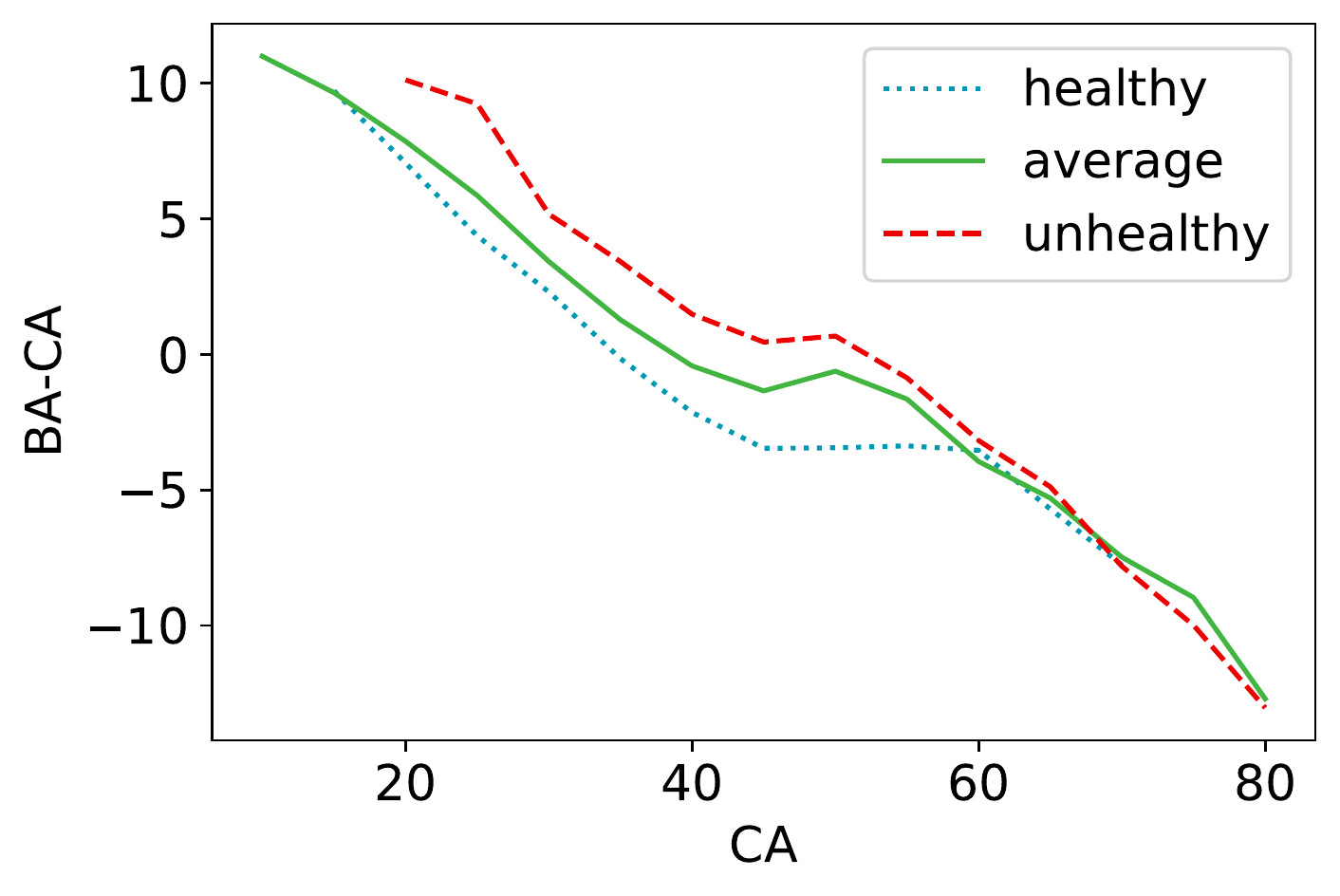}
        \caption{DNN}
    \end{subfigure}
    \begin{subfigure}[h]{.45\textwidth}
        \includegraphics[width=\textwidth]{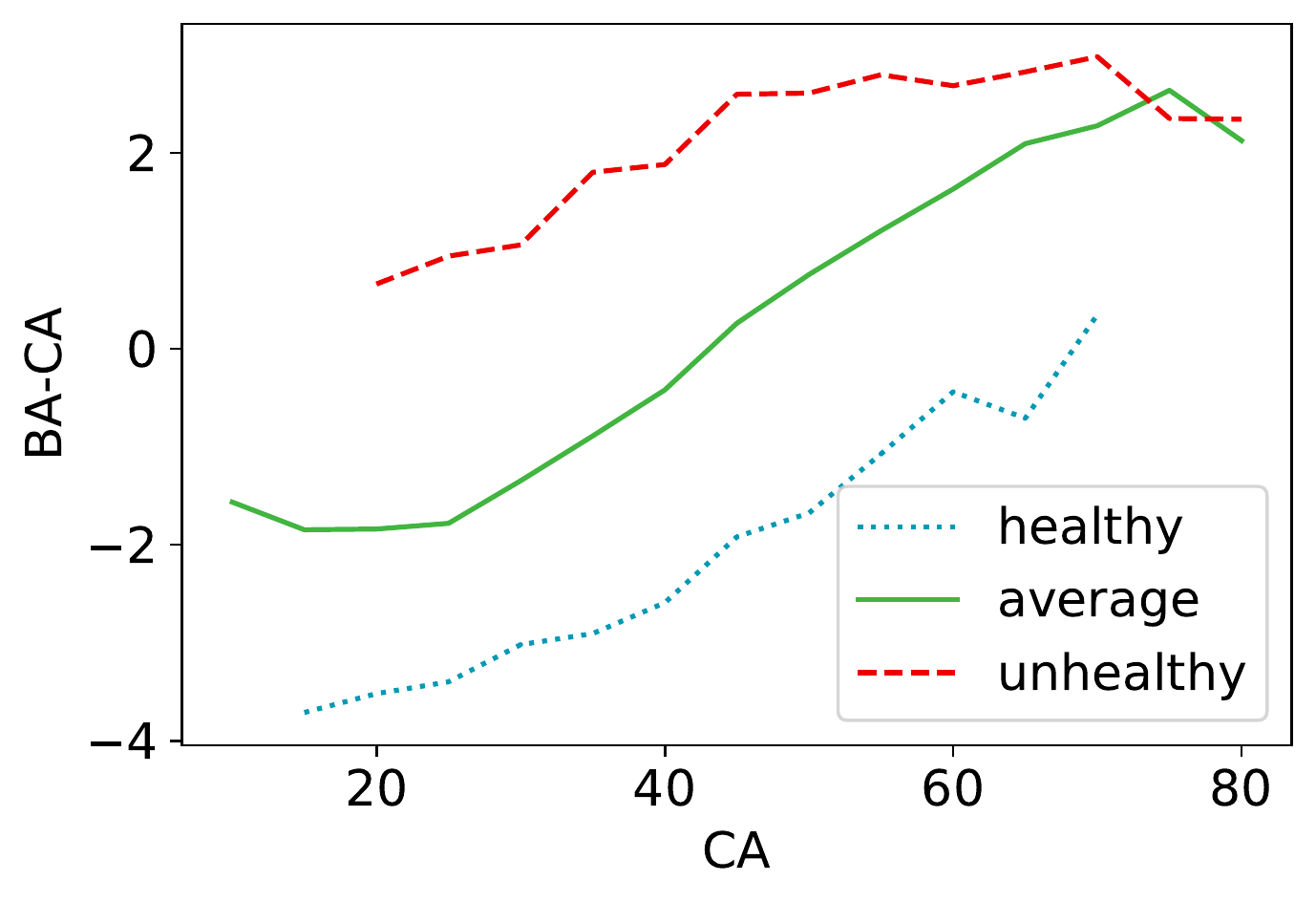}
        \caption{proposed}
    \end{subfigure}
    \caption{The gap between biological and chronological ages depending on the morbidity of \textbf{DLP} in female-\textit{whole}-\lfeature\ case.}
    \label{fig:f-all-l-morbidity-dlp}
\end{figure}

\begin{figure}
    \centering
    \begin{subfigure}[h]{.45\textwidth}
        \includegraphics[width=\textwidth]{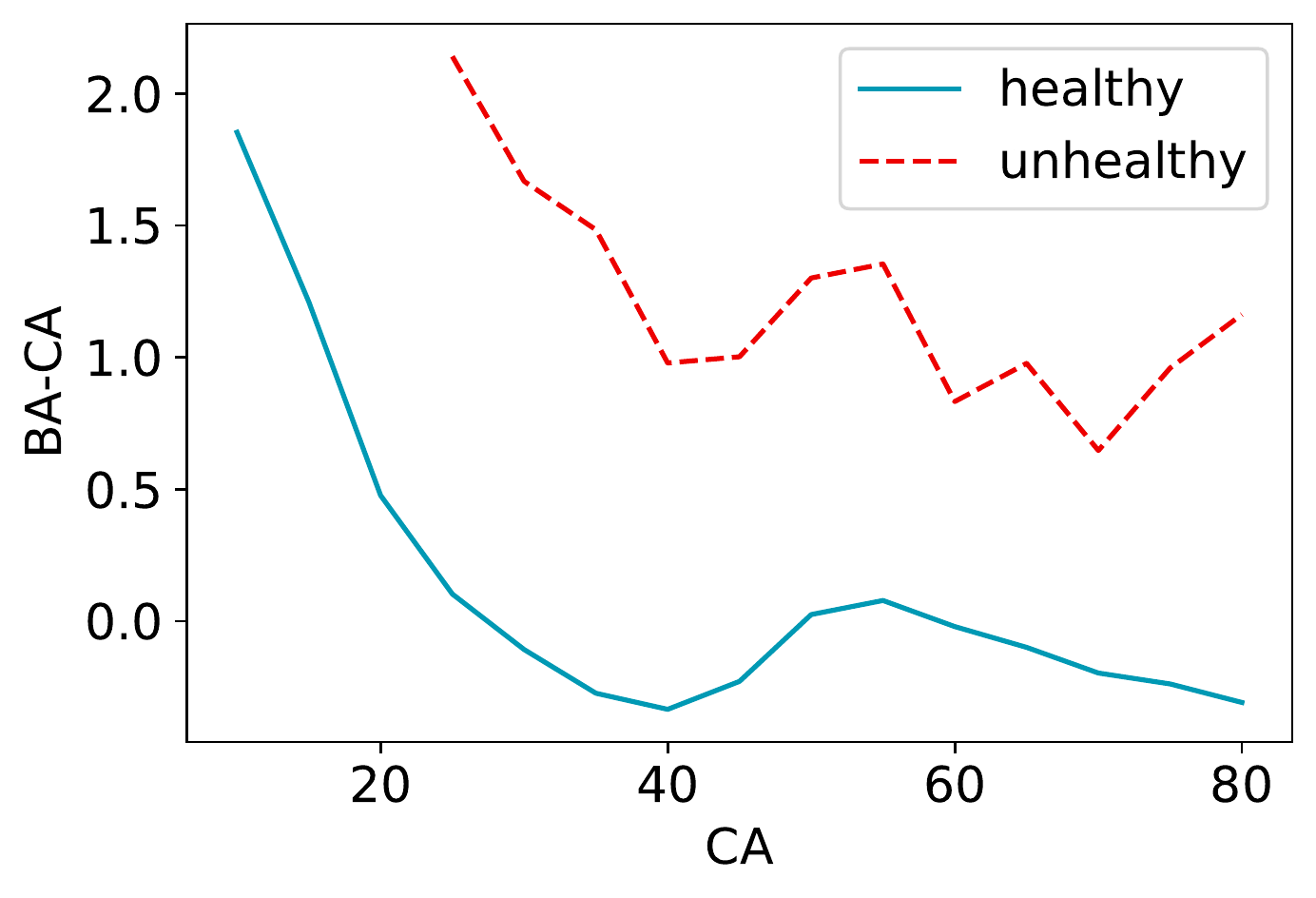}
        \caption{KDM}
    \end{subfigure}
    \begin{subfigure}[h]{.45\textwidth}
        \includegraphics[width=\textwidth]{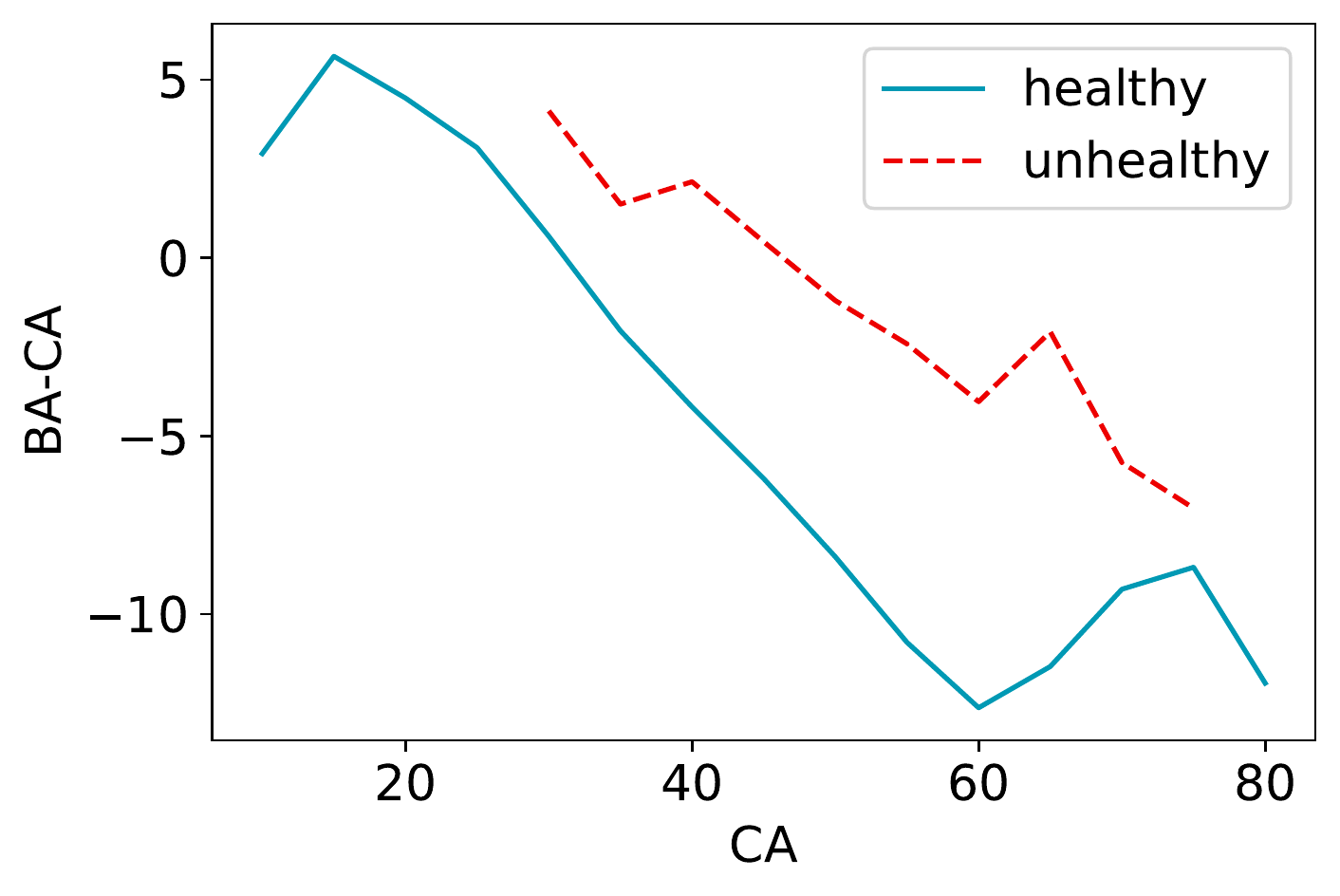}
        \caption{CAC}
    \end{subfigure}
    \begin{subfigure}[h]{.45\textwidth}
        \includegraphics[width=\textwidth]{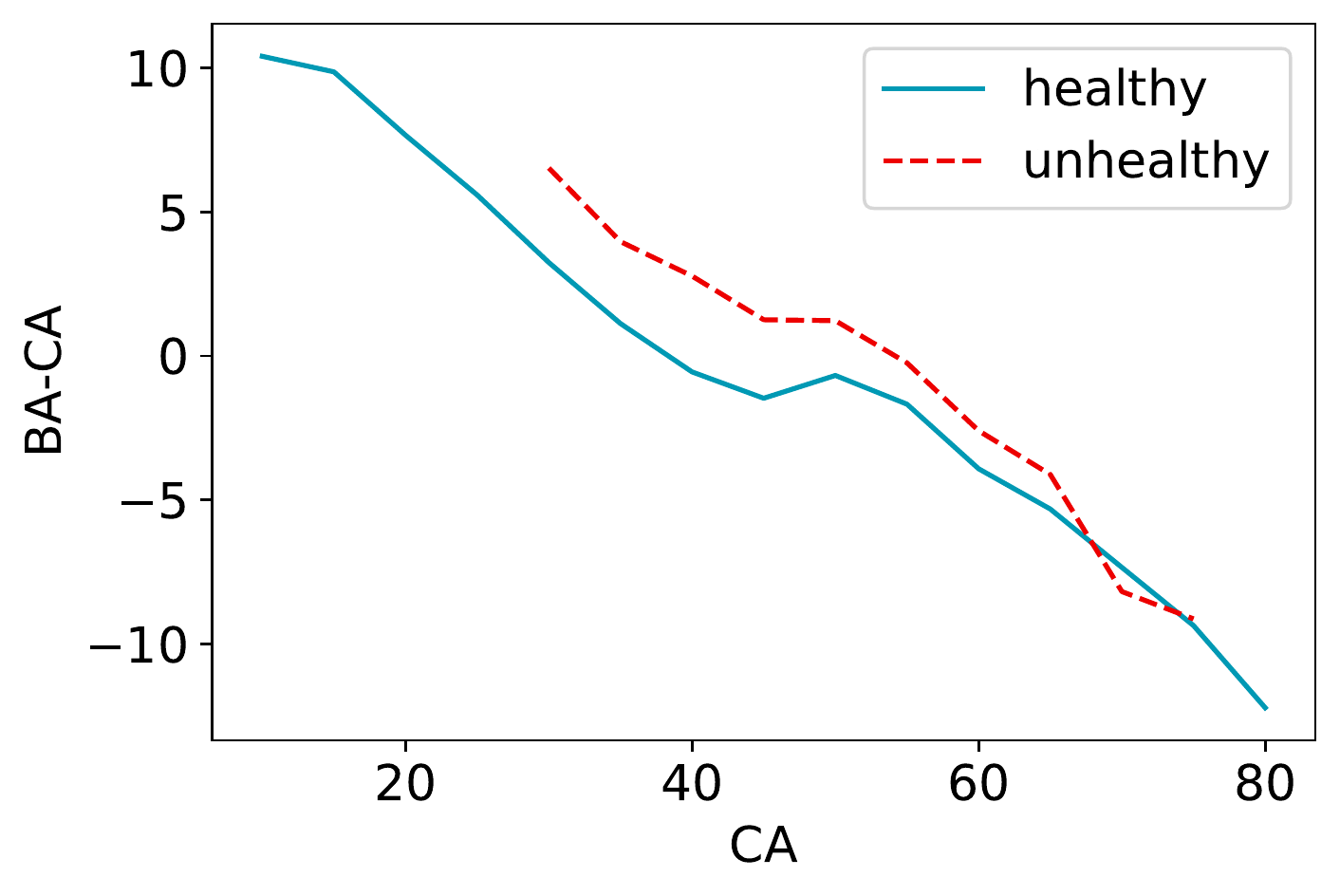}
        \caption{DNN}
    \end{subfigure}
    \begin{subfigure}[h]{.45\textwidth}
        \includegraphics[width=\textwidth]{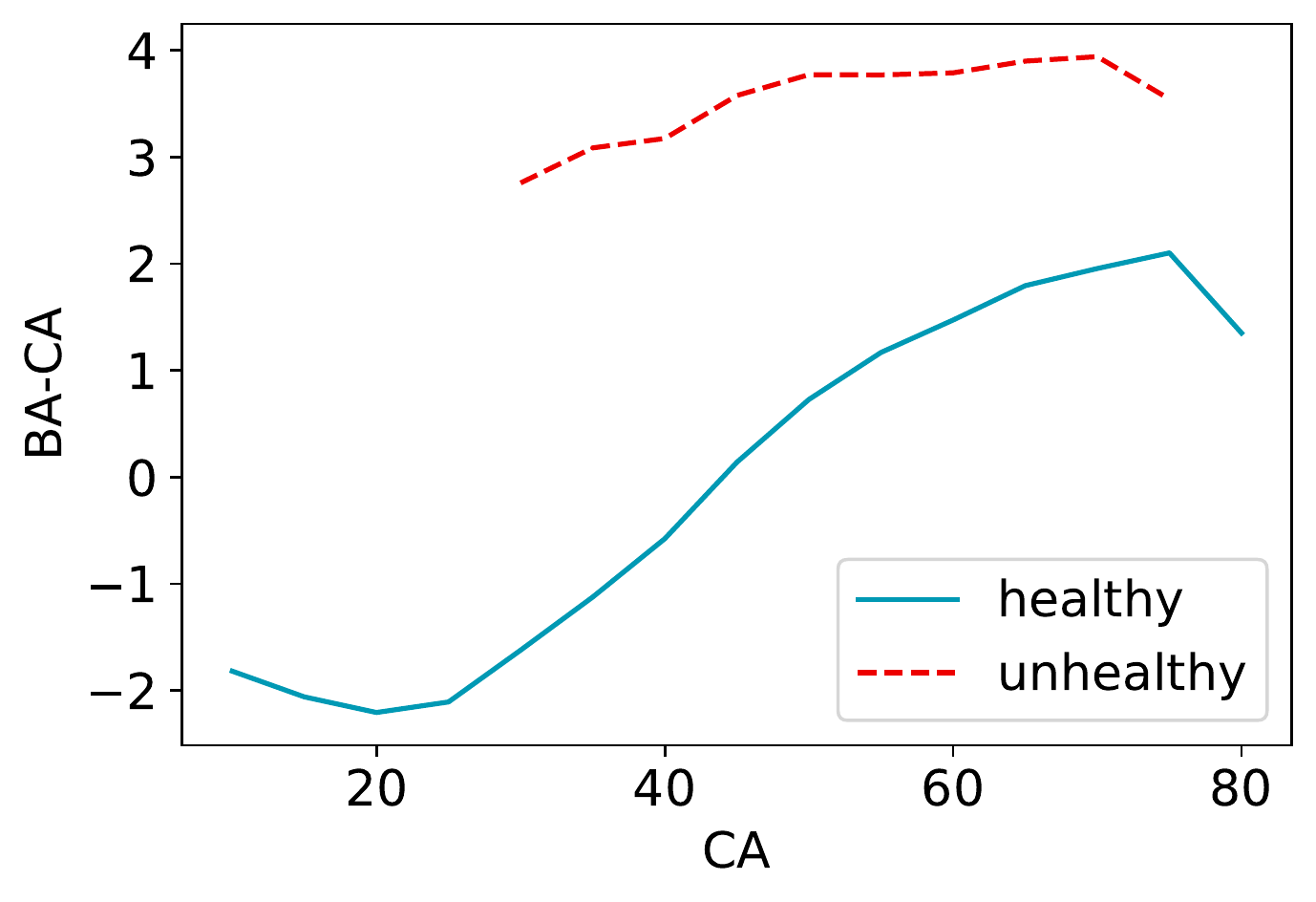}
        \caption{proposed}
    \end{subfigure}
    \caption{The gap between biological and chronological ages depending on the morbidity of \textbf{MS} in female-\textit{whole}-\lfeature\ case.}
    \label{fig:f-all-l-morbidity-ms}
\end{figure}

\begin{figure}
    \centering
    \begin{subfigure}[t]{.45\textwidth}
        \includegraphics[width=\textwidth]{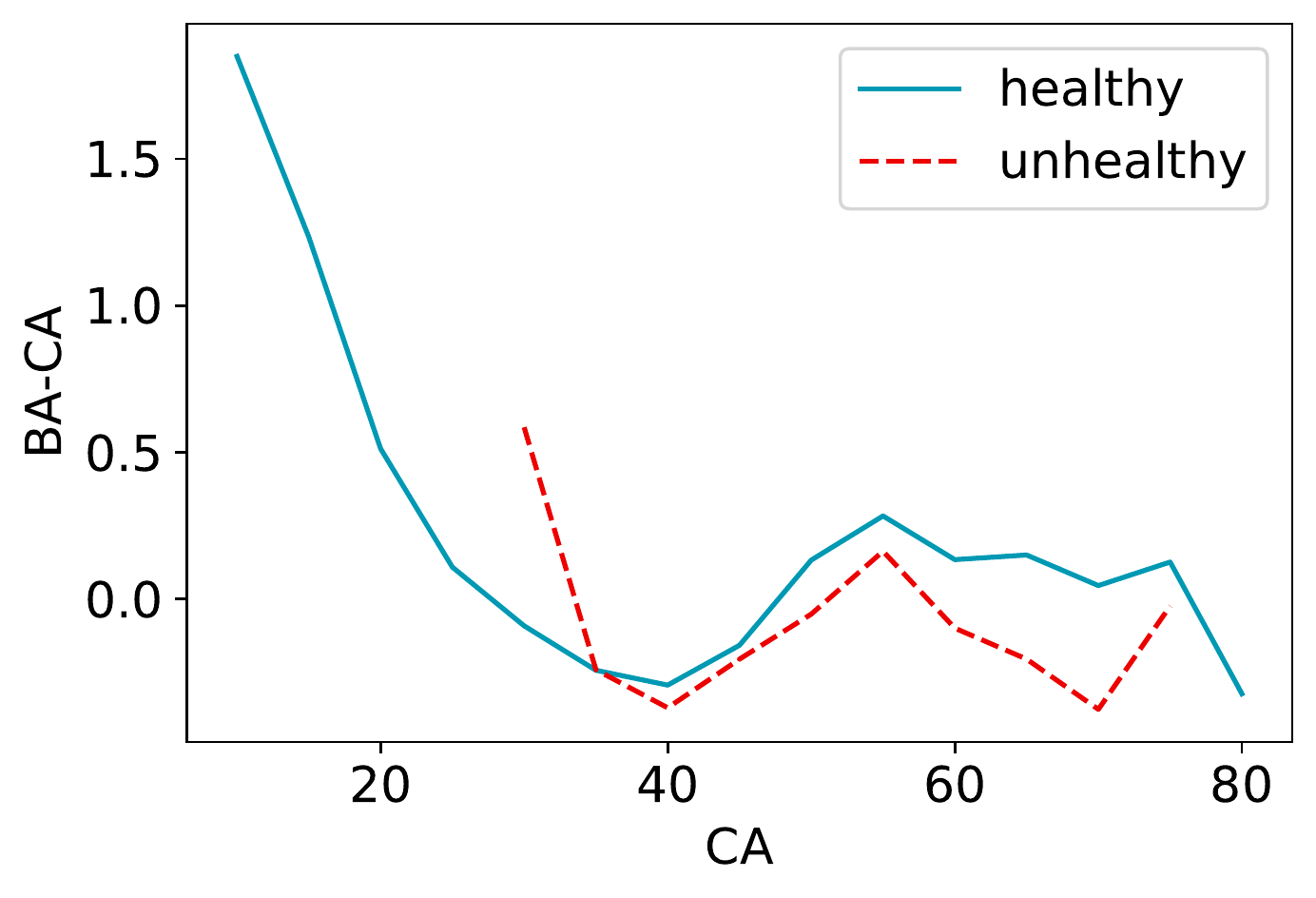}
        \caption{KDM}
    \end{subfigure}
    \begin{subfigure}[t]{.45\textwidth}
        \includegraphics[width=\textwidth]{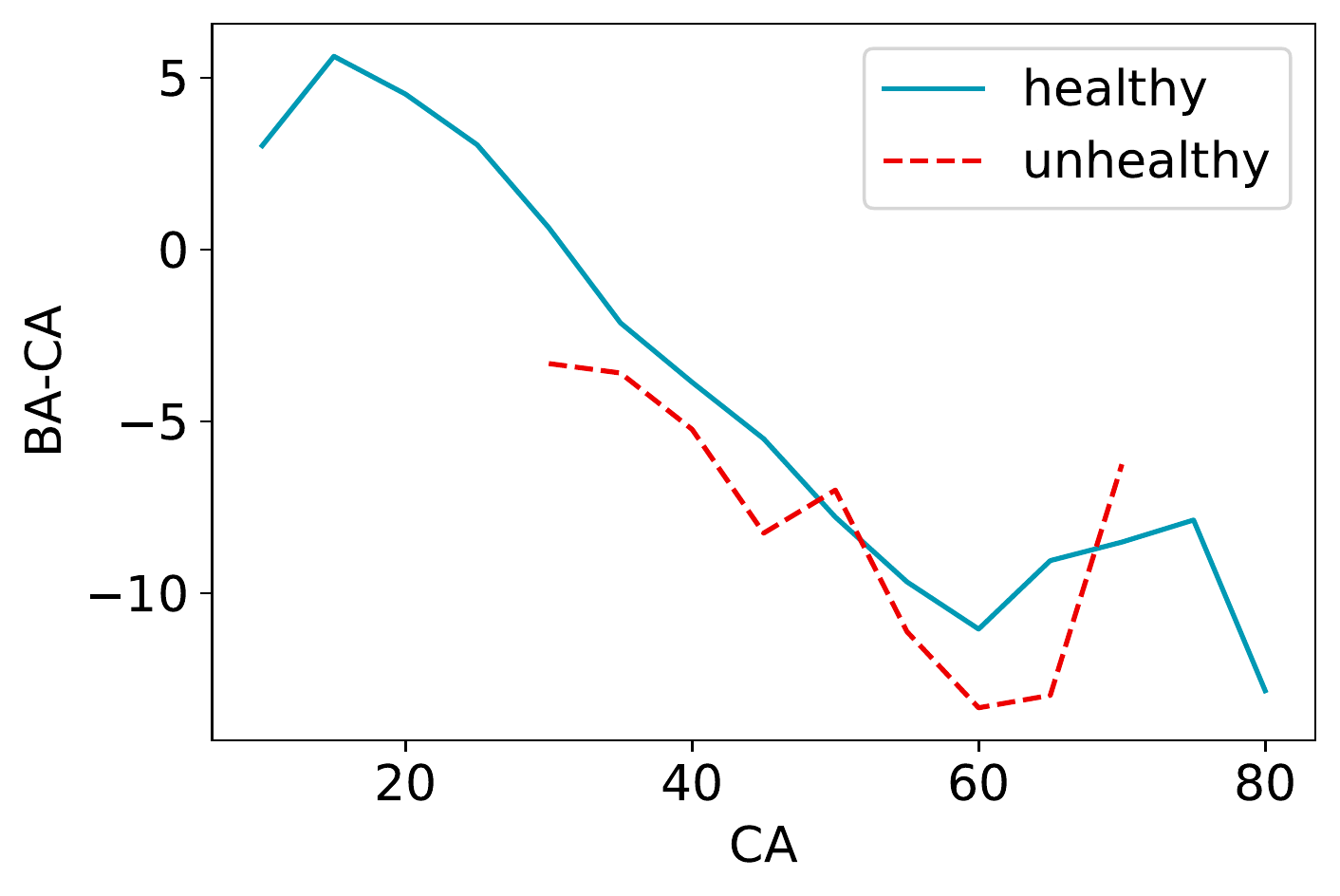}
        \caption{CAC}
    \end{subfigure}
    \begin{subfigure}[t]{.45\textwidth}
        \includegraphics[width=\textwidth]{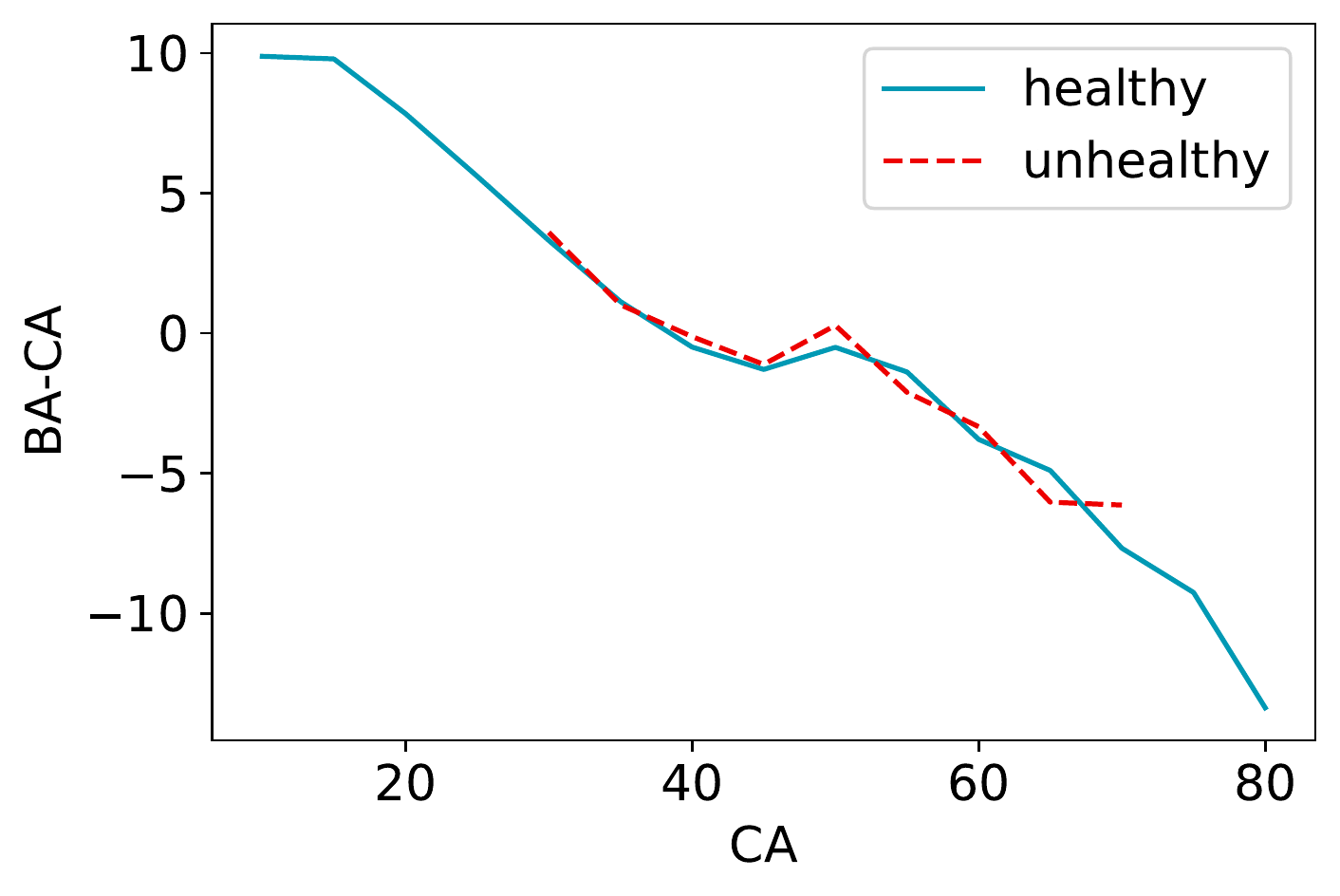}
        \caption{DNN}
    \end{subfigure}
    \begin{subfigure}[t]{.45\textwidth}
        \includegraphics[width=\textwidth]{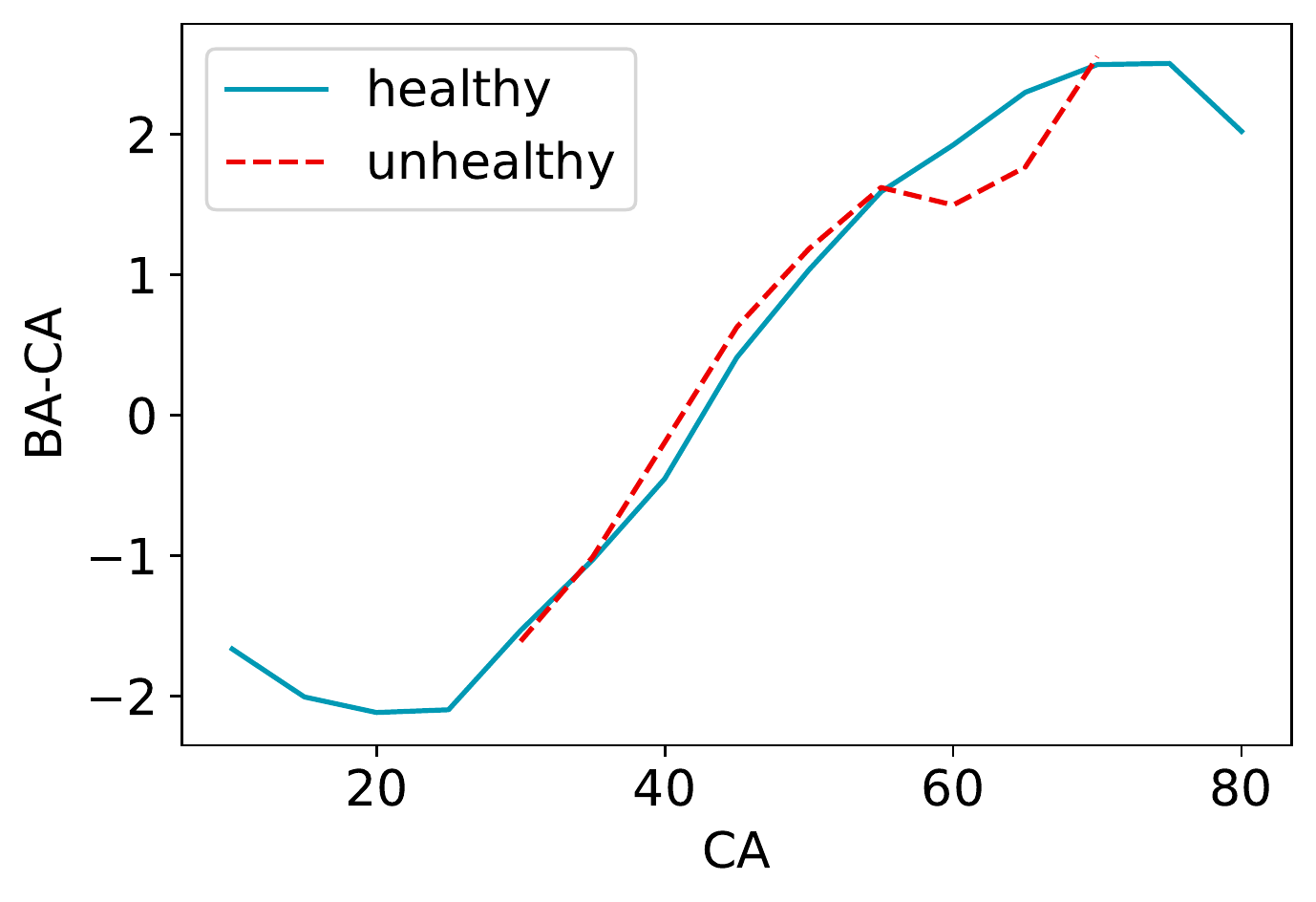}
        \caption{proposed}
    \end{subfigure}
    \caption{The gap between biological and chronological ages depending on the morbidity of \textbf{cancer} in female-\textit{whole}-\lfeature\ case.}
    \label{fig:f-all-l-morbidity-cancer}
\end{figure}

\begin{figure}
    \centering
    \begin{subfigure}[t]{.45\textwidth}
        \includegraphics[width=\textwidth]{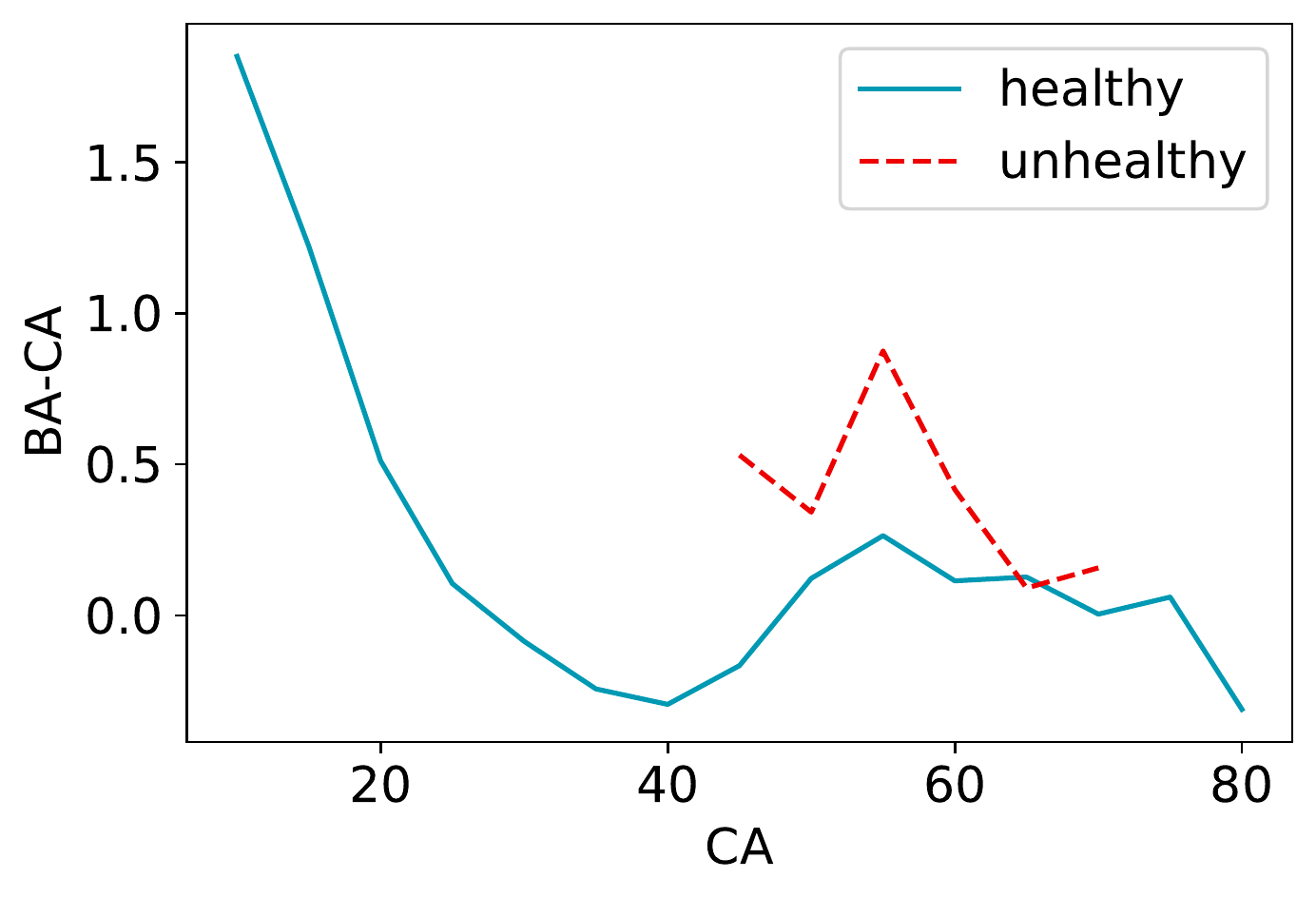}
        \caption{KDM}
    \end{subfigure}
    \begin{subfigure}[t]{.45\textwidth}
        \includegraphics[width=\textwidth]{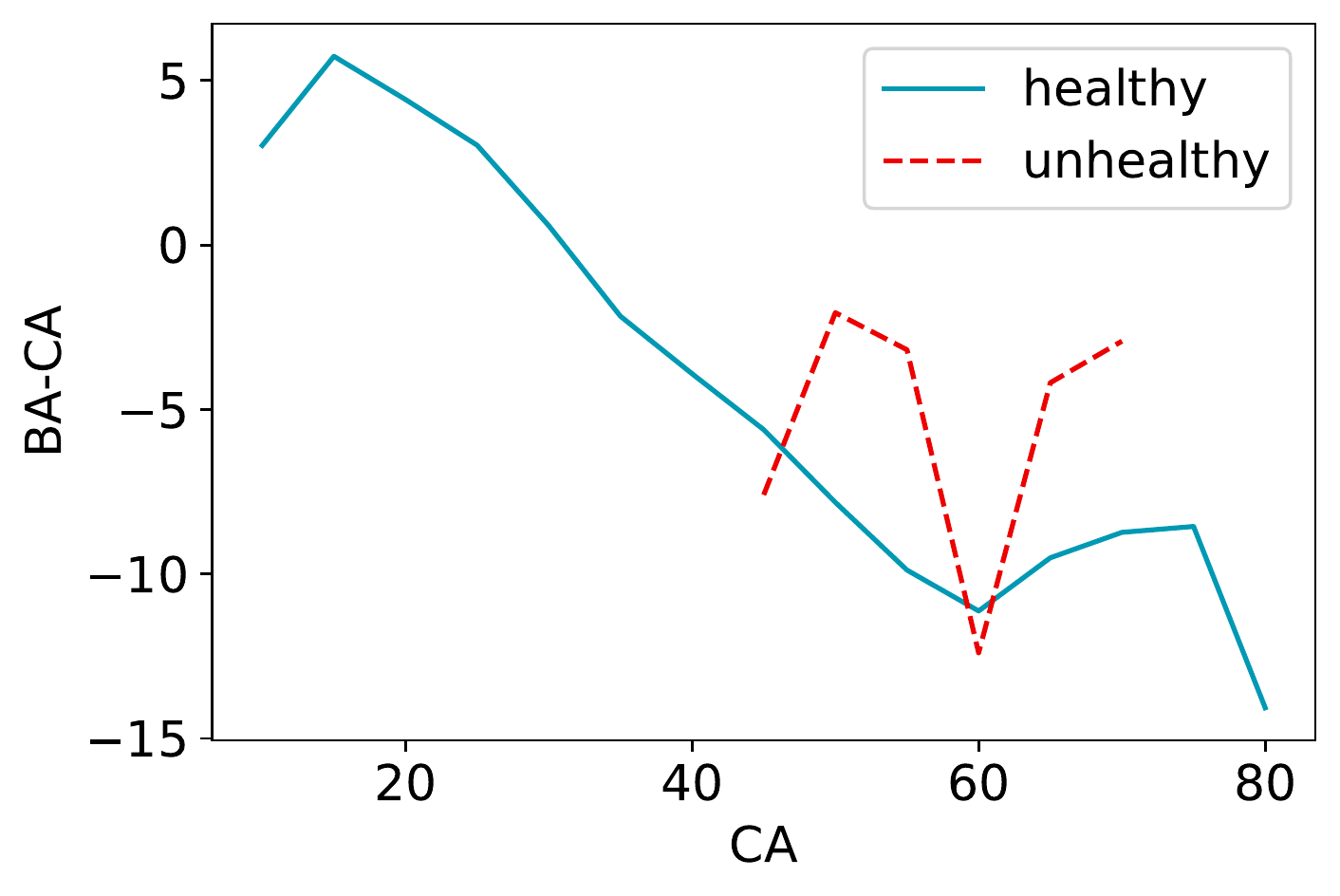}
        \caption{CAC}
    \end{subfigure}
    \begin{subfigure}[t]{.45\textwidth}
        \includegraphics[width=\textwidth]{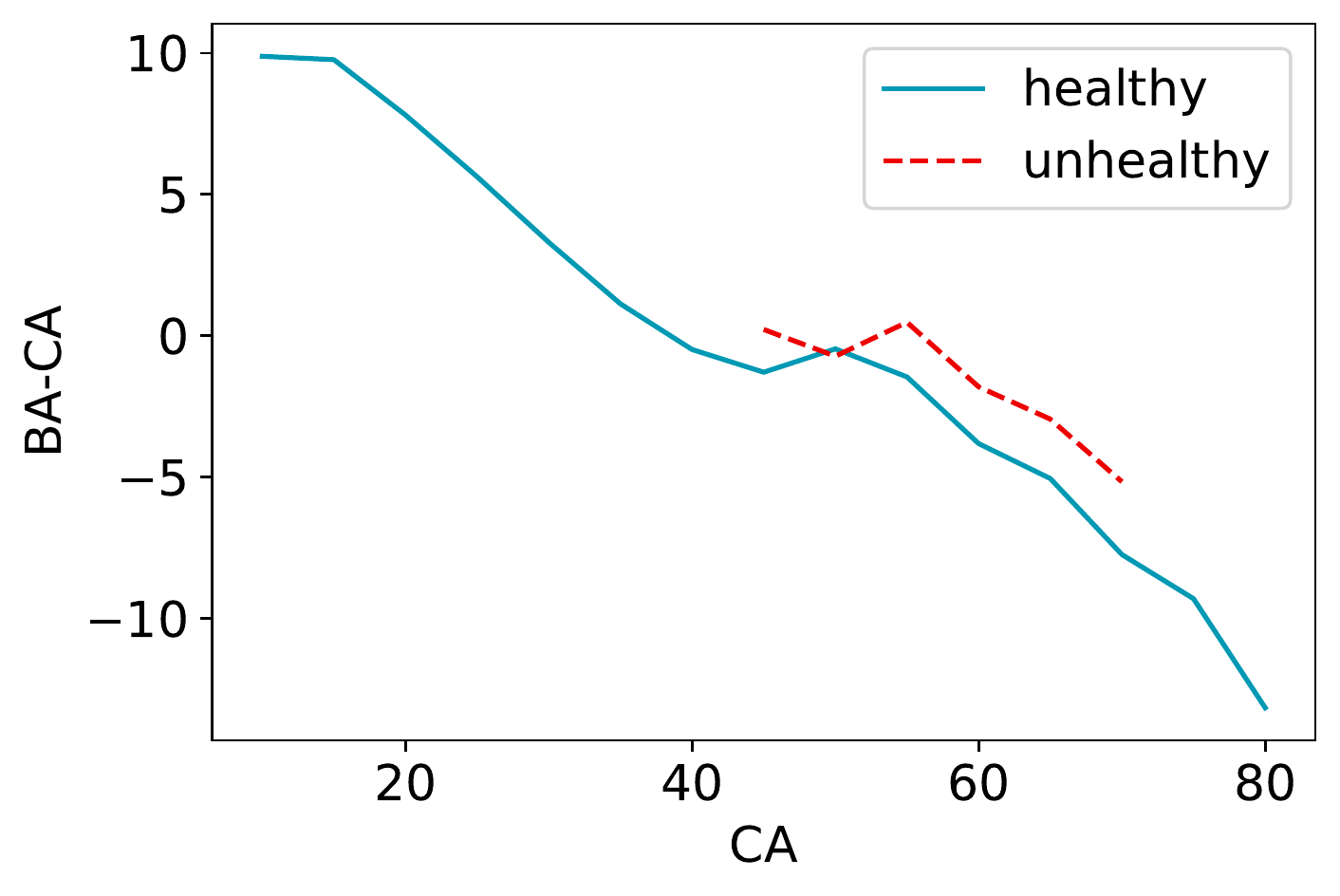}
        \caption{DNN}
    \end{subfigure}
    \begin{subfigure}[t]{.45\textwidth}
        \includegraphics[width=\textwidth]{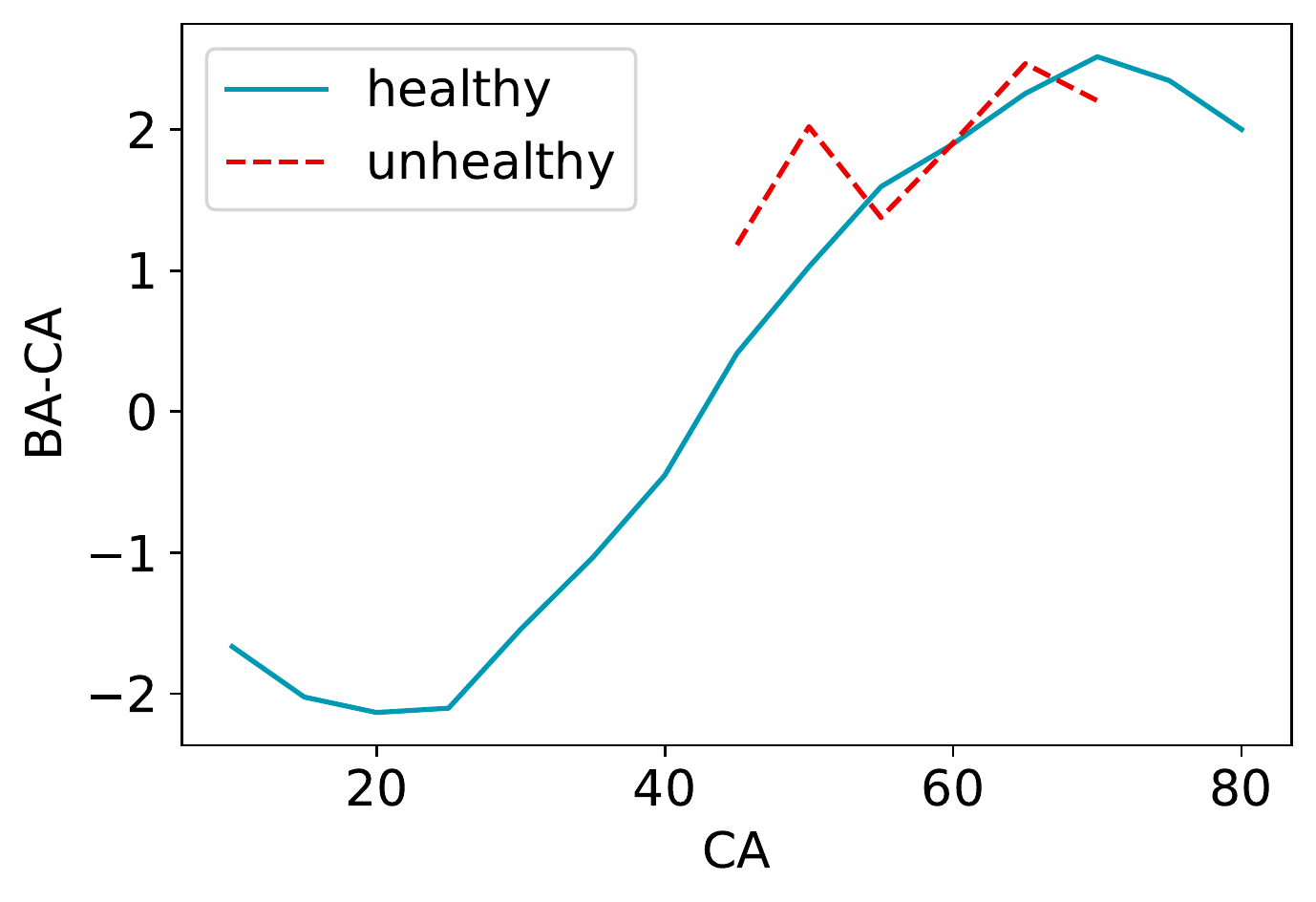}
        \caption{proposed}
    \end{subfigure}
    \caption{The gap between biological and chronological ages depending on the morbidity of \textbf{CVD} in female-\textit{whole}-\lfeature\ case.}
    \label{fig:f-all-l-morbidity-cvd}
\end{figure}


\clearpage
\section{Results of multivariate linear regression for time-to-death}
\label{appsec:gap_models}

\begin{table}[h]
    \centering
    \caption{Linear regression results for time-to-death using chronological and gap values for \textit{whole} population. 
    The slope and PCC are calculated for gap values. }
    \small
    \setlength\tabcolsep{5pt}
    \def\arraystretch{0.8}
    \begin{tabular}{c c c r r r r r r}
        \toprule \midrule
        
        \multirow{2}{*}{Population} & \multirow{2}{*}{Feature set} & \multirow{2}{*}{Model} & \multicolumn{2}{c}{Slope} & \multicolumn{2}{c}{$R^2$} & \multicolumn{2}{c}{PCC} \\ \cmidrule(lr){4-9}
        & & & CA & BA & CA & BA & CA & BA \\ \midrule
        
        \multirow{12}{*}{M} & \multirow{4}{*}{\lfeature} & 
        KDM & \multirow{4}{*}{-4.683}
        & \textbf{-2.737}$^{***}$ & \multirow{4}{*}{0.001} & 
        0.026 & \multirow{4}{*}{-0.032} & -0.144 \\
        & & CAC & & \textbf{-14.891}$^{***}$ & & 0.031 & & -0.168 \\
        & & DNN & & -0.602$^{\quad\;}$ & & 0.001 & & 0.013 \\
        & & proposed & & \textbf{-153.599}$^{***}$ & & \textbf{0.120} & & \textbf{-0.342} \\ \cmidrule(lr){2-9}
        
        & \multirow{4}{*}{\mfeature} & 
        KDM & \multirow{4}{*}{-1.601}
        & \textbf{-6.302}$^{***}$ & \multirow{4}{*}{0.000} & 
        0.018 & \multirow{4}{*}{-0.012} & -0.132 \\
        & & CAC & & -11.034$^{**\;}$ & & 0.010 & & -0.096 \\
        & & DNN & & \textbf{-19.891}$^{***}$ & & 0.012 & & -0.086 \\
        & & proposed & & \textbf{-139.756}$^{***}$ & & \textbf{0.101} & & \textbf{-0.317} \\ \cmidrule(lr){2-9}
        
        & \multirow{4}{*}{\sfeature} &
        KDM & \multirow{4}{*}{6.959$^{\;}$} 
        & \textbf{-8.209}$^{***}$ & \multirow{4}{*}{0.002} & 0.027 &  \multirow{4}{*}{0.048} & -0.158 \\
        & & CAC & & -6.394$^{\quad\; }$ & & 0.006 & & -0.069 \\
        & & DNN & & -14.627$^{**\;}$ & & 0.009 & & -0.096 \\
        & & proposed & & \textbf{-112.617}$^{***}$ & & \textbf{0.080} & & \textbf{-0.273} \\ \cmidrule(lr){1-9}

        \multirow{12}{*}{F} & \multirow{4}{*}{\lfeature} &
        KDM & \multirow{4}{*}{3.794} 
        & 12.141$^{\quad\;}$ & \multirow{4}{*}{0.001} & 0.008 & \multirow{4}{*}{0.029} & 0.011 \\
        & & CAC & & -6.2519$^{\quad\;}$ & & 0.005 & & -0.072 \\
        & & DNN & & -0.998$^{\quad\; }$ & & 0.001 & & -0.021 \\
        & & proposed & & \textbf{-167.651}$^{***}$ & & \textbf{0.086} & & \textbf{-0.274} \\ \cmidrule(lr){2-9}
        
        & \multirow{4}{*}{\mfeature} &
        KDM & \multirow{4}{*}{13.708} 
        & 0.706$^{\quad\;}$ & \multirow{4}{*}{0.011} & 0.014 & \multirow{4}{*}{0.106} & 0.047 \\
        & & CAC & & 2.625$^{\quad\;}$ & & 0.012 & & -0.010 \\
        & & DNN & & 21.222$^{\quad\;}$ & & 0.026 & & 0.037 \\
        & & proposed & & \textbf{-152.662}$^{***}$ & & \textbf{0.074} & & \textbf{-0.221} \\ \cmidrule(lr){2-9}
        
        & \multirow{4}{*}{\sfeature} &
        KDM & \multirow{4}{*}{11.272} 
        & 4.018$^{\quad\;}$ & \multirow{4}{*}{0.007} & 0.032 & \multirow{4}{*}{0.082} & 0.153 \\
        & & CAC & & 9.750$^{\quad\;}$ & & 0.016 & & 0.076 \\
        & & DNN & & 23.433$^{\quad\;}$ & & 0.023 & & 0.064 \\
        & & proposed & & \textbf{-95.941}$^{***}$ & & \textbf{0.049} & & \textbf{-0.210} \\
        \midrule \bottomrule
    \end{tabular}
\end{table}